\documentclass{JQRnew}

\JQRsetup{
 volume   = 47,
 number   = 4,
 year    = 2025,
 month   = 4,
 doi    = 250208,
 funds   = {
 国家自然科学基金资助项目（62325211, 62132021, 62322207, 92148204, 62433010,  62303174）；“兴辽英才计划”高端装备数字孪生总装技术与机器人制造系统（YS2023004）；湘江实验室重大项目（23XJ01009）； 武汉市重点研发计划（2024060702030143）； 
 ‘国家资助博士后研究人员计划’和‘中国博士后科学基金’（BX20250386）；
 国家重点研发计划（2024YFB4708900）；湖南省自然科学基金重点项目（2025JJ30024）；湖南省优秀青年项目（24B0044）；中央高校基本科研业务费专项资金项目（531118010815）
 },
 corrauthor = {徐凯，kevin.kai.xu@gmail.com},
 dates   = {2025-05-13/2025-06-06/2025-06-16},
 cntitle    = {面向柔性制造的具身智能综述},
cnauthor  = {徐\quad 凯\addr{1}, 赵\quad 航\addr{2}, 胡瑞珍\addr{3}, 杨\quad 敏\addr{4}, 刘\quad 浩\addr{5}, 张\quad 辉\addr{4}, 于海斌\addr{5}  },
 runtitle  = {徐凯，等：面向柔性制造的具身智能综述},
 cnaddress = {1.中国人民解放军国防科技大学, 2.武汉大学, 3.深圳大学, 4.湖南大学, 5.中国科学院沈阳自动化研究所},
 cnabstract = {
 在新一代人工智能取得突破性进展的驱动下，具身智能作为人工智能领域的重要分支，正加速向工业制造场景渗透。工业场景因其制造环境半结构化、工况特征相对稳定、工艺流程相对标准，更易实现具身智能技术的快速落地，很有可能成为具身智能的首个规模化应用领域。不过，随着“多品种、小批量”柔性制造模式成为现代制造业主流趋势，产线多品类混流生产、产品迭代更新频繁和制造工艺无序非标已成为制造业新常态，对智能制造系统的复杂工艺处理能力和制造精度保障能力提出了双重要求。因而，柔性制造场景中的具身智能面临三大核心挑战： \ding{172} 受限感知下的工艺精准建模监测难题；\ding{173}柔性适配与高精操控的动态平衡难题；\ding{174}通用技能与专用工艺的协同融合难题。这些挑战也为具身智能本身的发展带来了新机遇。围绕上述机遇和挑战，本文从“工业之眼-工业之手-工业之脑”三个维度对现有工作进行综述：在感知层（工业之眼）重点探讨复杂动态环境下的多模态数据融合与实时建模方法；在控制层（工业之手）深入剖析复杂制造工艺的柔性自适应精准操控方法；在决策层（工业之脑）系统总结工艺规划与产线调度的智能优化方法。通过多层级技术协同、多学科交叉融合的视角，揭示制造系统“感知-决策-执行”闭环优化的具身智能关键技术路径，提出柔性制造场景下具身智能发展的“认知增强-技能跃迁-系统进化”三阶段演进模型， 探讨了未来发展趋势。期望为柔性制造趋势下的工业具身智能跨学科融合发展提供理论框架和实践参考。
 },
 cnkeywords = {工业具身智能；柔性制造； 离散制造； 环境感知； 自主决策；智能调度},
 entitle  = { Embodied Intelligence for Flexible Manufacturing: A Survey},
  enauthor  = {XU Kai\addr{1}, ZHAO Hang\addr{2}, HU Ruizhen\addr{3}, YANG Min\addr{4}, LIU Hao\addr{5}, ZHANG Hui\addr{4}, YU Haibin\addr{5}},
 enaddress = {1. \textit{National University of Defense Technology}, 
        2. \textit{Wuhan University}, 
        3. \textit{Shenzhen University}, 
        4. \textit{Hunan University}, \\
        5. \textit{Shenyang Institute of Automation Chinese Academy of Sciences}
        },
 enabstract = {
Driven by breakthrough advances in modern artificial intelligence, embodied intelligence, an important subfield of AI, is rapidly making inroads into industrial manufacturing. Industrial scenarios, with their semi-structured environments, relatively stable operating conditions, and standardized processes, offer fertile ground for the swift deployment of embodied intelligence technologies and are poised to become the first domain to see their large-scale adoption. However, as the modern manufacturing paradigm shifts toward high product variety and small batch sizes, mixed-model production lines, rapid product iterations, and unstructured, non-standardized processes have become the new normal, placing simultaneous demands on intelligent manufacturing systems to manage complex workflows and guarantee the required level of production precision.
Embodied intelligence in flexible manufacturing faces three core challenges: \ding{172} accurate process modeling and monitoring under limited sensing; \ding{173} maintaining a dynamic balance between flexible adaptation and high-precision manipulation; \ding{174} coordinating general robotic skills with specialized industrial procedures. These challenges create new opportunities for the field.
Accordingly, this survey reviews existing work from three viewpoints: Industrial Eye, Industrial Hand, and Industrial Brain. At the perception level (Industrial Eye), we examine multimodal data fusion and real-time modeling in complex, dynamic settings. At the control level (Industrial Hand), we analyze flexible, adaptive, and precise manipulation for demanding processes. At the decision level (Industrial Brain), we summarize intelligent optimization methods for process planning and line scheduling. By considering multi-level collaboration and interdisciplinary integration, we identify key pathways for closing the perception-decision-action loop. We propose a three-stage evolution model for embodied intelligence—cognition enhancement, skill transition 
and discuss future directions. Our goal is to offer a theoretical foundation and practical guidance for cross-disciplinary progress in industrial embodied intelligence under the trend toward flexible manufacturing.
 },
 enkeywords = {industrial embodied intelligence; flexible manufacturing; discrete manufacturing; environmental perception; autonomous decision-making; intelligent scheduling}
}

\usepackage{algorithm}
\usepackage{algpseudocode}
\usepackage{lipsum}
\usepackage{zhlipsum}
\usepackage{pifont}
\usepackage{balance}
\usepackage{mwe}
\usepackage{textcomp} 
\usepackage{pifont}
\usepackage{array}    
\newcolumntype{M}[1]{>{\centering\arraybackslash}m{#1}}

\floatname{algorithm}{算法}

\begin{document}

\definecolor{green}{rgb}{0, 0.5, 0}
\definecolor{orange}{rgb}{1.0, 0.6, 0.2}
\definecolor{red}{rgb}{1.0, 0.0, 0.0}
\definecolor{blue}{rgb}{0.0, 0.0, 1.0}
\definecolor{teal}{rgb}{0.0, 0.4, 0.4}
\definecolor{purple}{rgb}{0.65,0,0.65}
\definecolor{saffron}{rgb}{0.95,0.75,0.2}
\definecolor{turquoise}{rgb}{0.0,0.5,0.5}
\definecolor{black}{rgb}{0,0,0}

\newcommand{\kx}[1]{{\color{black}#1}}
\newcommand{\rh}[1]{{\color{blue}#1}}
\newcommand{\rl}[1]{{\color{teal}#1}}
\newcommand{\zy}[1]{{\color{turquoise}#1}}
\newcommand{\hx}[1]{{\color{purple}#1}}
\newcommand{\wl}[1]{{\color{saffron}#1}}
\newcommand{\zh}[1]{{\color{black}#1}}
\newcommand{\mm}[1]{{\color{orange}#1}}

\maketitle

\clearpage

在新一代人工智能如 DeepSeek\ucite{guo2025deepseek}、GPT-4\ucite{achiam2023gpt}等取得突破性进展的驱动下，\emph{具身智能}（Embodied Intelligence）作为人工智能领域的重要分支，正加速向工业制造场景渗透。
具身智能强调智能体可以通过与环境的直接互动表现出智能\ucite{brooks1991intelligence}，少依赖或无需显式符号推理\ucite{mccarthy2006proposal}，在家庭服务\ucite{li2024llm, WuCHPGKSRB24}、自主导航\ucite{zheng2019active, ZhangDMFCXW23}、巡检救援\ucite{yuan2022novel, alam2021smart}等开放领域展现出广阔的应用前景。
\emph{工业具身智能}（Industrial Embodied Intelligence）则通过多模态感知融合、动态环境建模与自主决策闭环，使制造系统在特定工业场景中，依据生产目标要求及工艺流程约束，完成生产作业任务。
工业场景因其制造环境半结构化、工况特征相对稳定、工艺流程相对标准，更易实现具身智能技术的快速落地，很有可能成为具身智能的首个规模化应用领域。不过，现代制造业呈现柔性化发展趋势，产线多品类混流生产、产品迭代更新频繁和制造工艺无序非标已成为制造业新常态。
这些特点对智能制造系统的复杂工艺处理能力和制造精度保障能力都提出了更高要求，
\emph{面向柔性制造的工业具身智能}面临三大独特挑战：

\begin{itemize}
    \item \textbf{\emph{受限感知下的工艺精准建模监测：}} 
    工业生产环境的精准建模与监测是实现制造系统自主决策和精确控制的基础，其中包含对制造环境的几何建模和物理动力学辨识。
    前者为制造系统的路径规划与操作控制提供空间信息，后者支持复杂环境中的状态预测、风险规避与策略优化。
    然而，受观测范围、感知精度和遮挡干扰影响，生产过程常常只能监测部分环境信息，导致建模困难。例如，大型复杂工件的几何重建中，单一视角的工业相机难以覆盖全工件表面；螺栓拧紧工艺中，工业相机只能观测螺栓端侧，无法感知内部物理规律。
    工艺建模失准会进一步影响制造系统的决策与控制，降低生产效率与产品质量。\textbf{因此，在受限感知下实现复杂环境的工艺精准建模监测，
    是工业具身智能亟需突破的重要问题。}
    \item \textbf{\emph{柔性适配与高精操控的动态平衡：}} 在应对多品种、小批量生产需求时，制造系统既要保证制造精度（如汽车装配精度往往需要达到丝级 $\pm$ 0.05mm），又要灵活应对因制造品类和工艺动态变化引起的工况变化、产线重构等挑战。
    由于小批量制造难以通过制造规模来摊平成本，柔性制造产线往往无法做到像传统大规模制造产线的精度。
    例如，新能源汽车车型更新迭代快，往往需要一条产线混产多种车型，
    产线精度与传统燃油车制造的专用产线无法同日而语，但并不能因此而降低工艺水平要求。\textbf{因此，如何在低精度产线上智能适应完成高精度工艺，为工业具身智能带来了巨大挑战。}
    \item \textbf{\emph{通用技能与专用工艺的协同融合：}} 
    智能制造系统既要具备跨领域的基础操作能力，如抓取、放置、装配、拧紧、轨迹跟踪、曲面随形等，又要掌握面向特定制造工艺的专家级技能。例如，智能焊接机器人既需要具备焊缝跟踪的基础技能，同时需要具备焊接工艺知识，如不同焊接工艺、母材种类、坡口形状等条件下的工艺参数基础设置\ucite{wanghongguang2010}。知识-数据双驱动的焊接工艺控制技能也需要有效掌握，通过对焊接熔池的实时监控和闭环反馈，对焊接电流、焊枪角度、移动速度等工艺参数进行实时调控，从而确保焊接质量。\textbf{因此，如何基于基础通用操作技能融合实现特定复杂制造工艺，是工业具身智能面临的独特挑战。}
\end{itemize}

\begin{figure*}[t!]
    \centering
    \includegraphics[width=\linewidth]{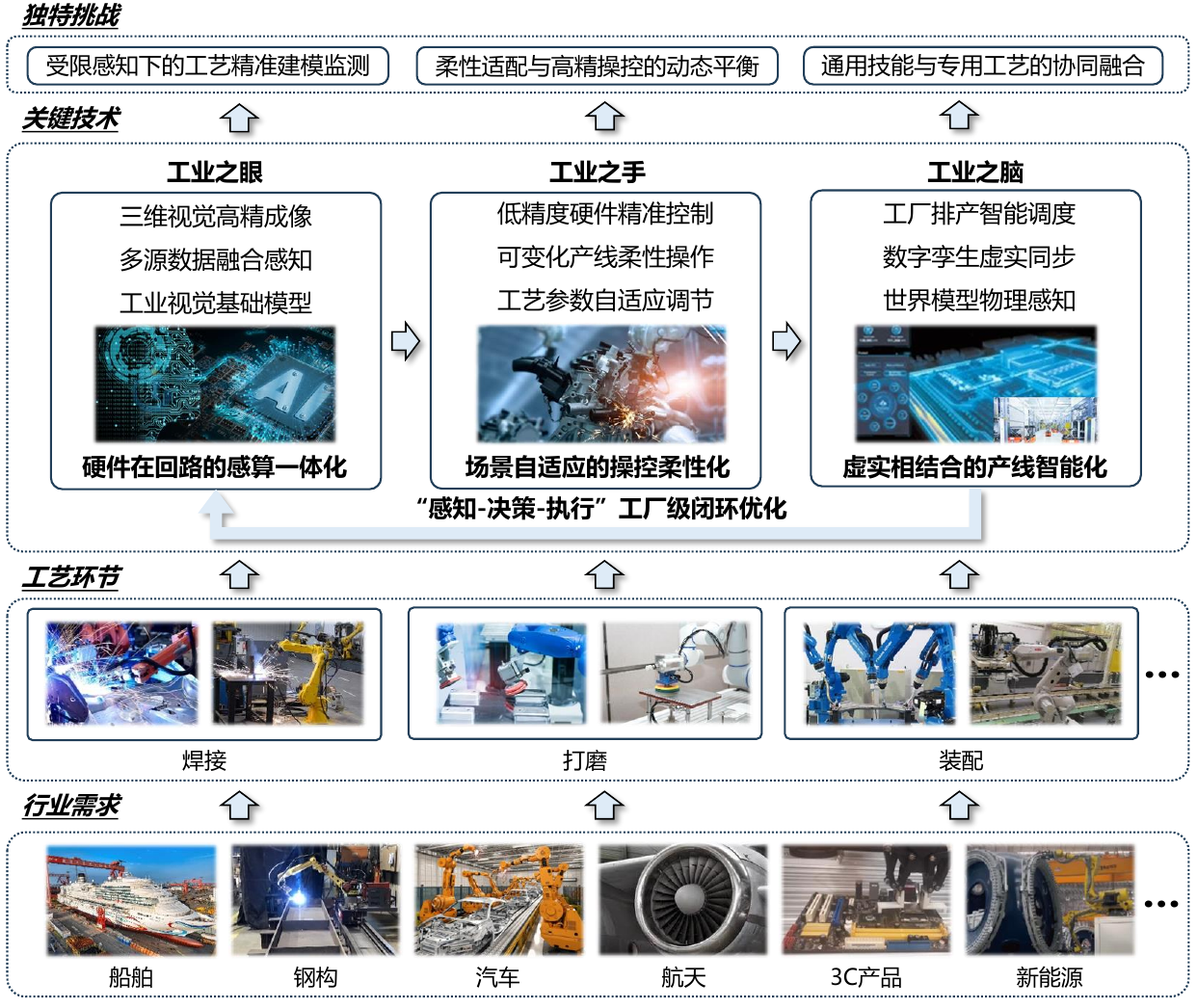}
    \bicaption{综述整体框架}{The overall framework of this survey}
    \label{framework}
\end{figure*}

\begin{figure}[t!]
    \centering
    \includegraphics[width=0.96\linewidth]{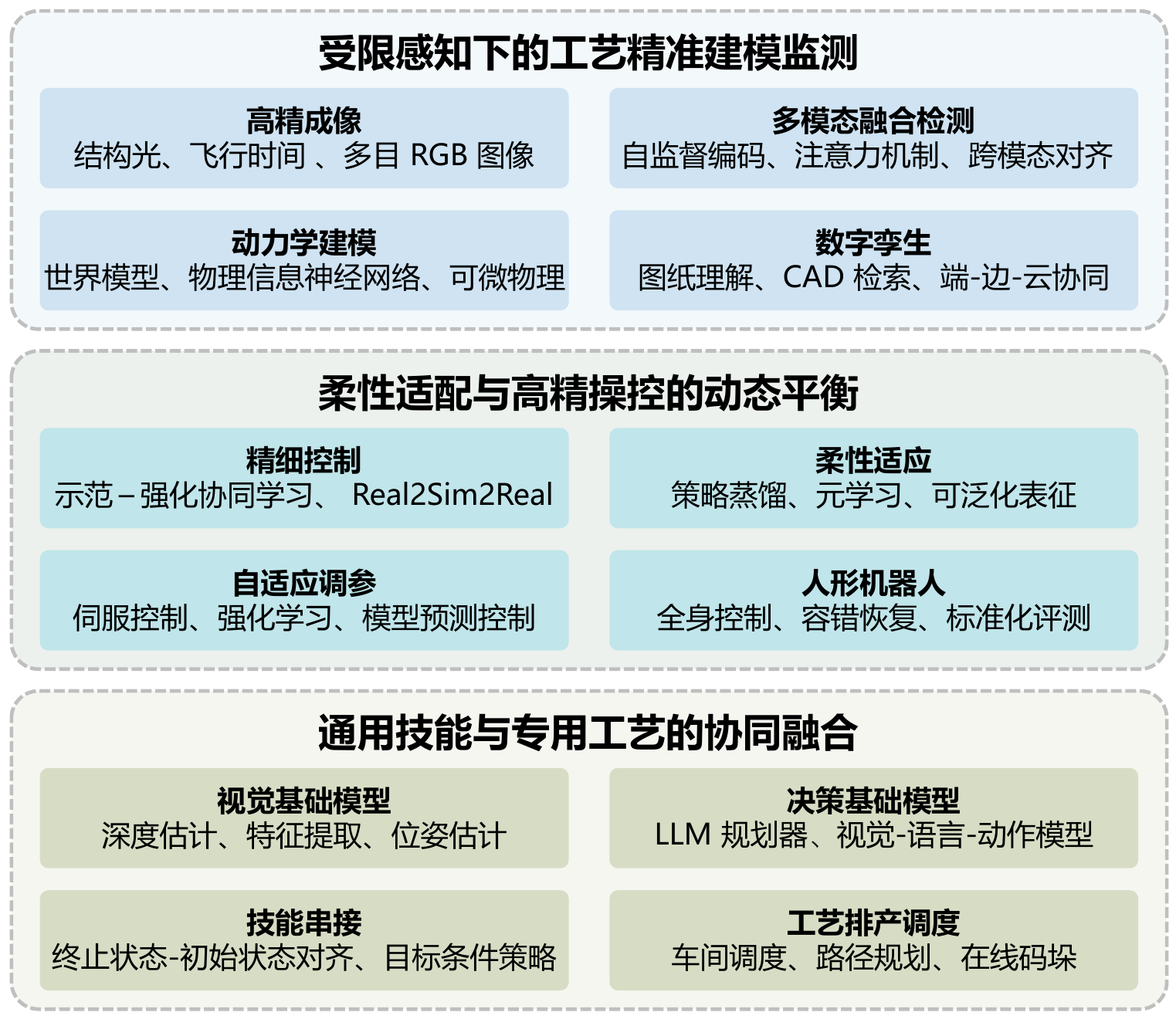}
    \bicaption{\zh{核心挑战与其解决方案} }{\zh{The core challenges and solutions}    }
    \label{technical_pipeline_first}
\end{figure}

这些柔性制造带来的挑战同样为工业具身智能的理论研究以及更加广泛的应用带来了新的机遇。
围绕上述挑战，本文从“工业之眼-工业之手-工业之脑”三个维度对现有工作进行综述。
作为感知层的\emph{工业之眼}
对制造对象和过程精准监测，需突破结构化场景的专用感知，实现面向复杂、可变工况的多模态、通用化感知与理解。需注意的是，工业之眼是一个广义概念，并非只包含视觉感知，也涵盖力触觉、超声波、电信号等多模态感知。
处在控制层的\emph{工业之手}是指制造系统通过实时、精准操控来实现某种制造工艺的执行与调控。需突破面向预设工艺的离线编程限制，实现面向复杂制造工艺的自适应、智能化、精准调控。
而决策层的\emph{工业之脑}负责更为宏观的规划和决策，一般指对整个制造产线的智能调度、工艺规划与最优控制。
需突破固定制造流程的优化，实现面向多任务、多工段、多工位全局排产优化，同时还需要灵活适应生产任务的动态变化（如订单变更、插单等）。当然，整个工业产线的智能、高效运转，是分布在各处的传感器（眼）和执行器（手），在集中或分布式部署的工业大脑的统一监控和调度之下协调运行的结果。本文的主要框架如图~\ref{framework} 所示。

具身智能研究已有半个多世纪积累，在通用具身智能的感知、控制与决策层面已有多篇系统综述\ucite{shentianyu,liu2025aligning,liu2025embodied}，在人形机器人\ucite{SSI-2024-0350}、大模型赋能的具身系统\ucite{SSI-2024-0076,wangwensheng}乃至仿真平台\ucite{duan2022survey}等细分领域也有详尽梳理。
然而，这些工作对工业柔性制造这一垂直领域的独特需求与技术挑战缺乏针对性分析。
Ren 等人\ucite{SSI-2024-0185}探讨了工业大模型在生产过程中的赋能作用，但对具身智能在工业中的具体角色与作用缺乏讨论。随着现代制造业向多品种、小批量、高精度方向柔性化转型，系统评估具身智能技术在工业制造中的应用优势与关键瓶颈成为迫切需要。
本文在继承具身智能经典理论与工业背景定义的基础上，系统总结了面向柔性制造的工业具身智能核心挑战与解决方案，如图~\ref{technical_pipeline_first} 所示。
通过
多层级技术协同，融合机器人学、机器视觉、人工智能等多学科交叉视角，揭示制造系统“感知-决策-执行”闭环优化的具身智能关键技术路径，并结合焊接、打磨、装配等典型案例分析其真实工业应用。
基于这些系统的总结与分析，本研究期望为柔性制造趋势下的工业具身智能跨学科融合发展提供理论框架和实践参考。
需要指出的是，本文聚焦于\emph{离散制造}（Discrete Manufacturing）\ucite{helu2020industry}，即通过加工、装配、检测等一系列明确工序完成产品制造的过程，强调操作序列的建模与执行。与之相对的\emph{过程制造}（Process Manufacturing）\ucite{fisher2018cloud}则侧重于物料与配方的连续处理与参数控制，不在本文讨论范畴内。

内容安排方面，本文依照“工业之眼-工业之手-工业之脑”的顺序依次展开，介绍这些维度如何共同解决面向柔性制造的工业具身智能独特挑战。
面向受限感知下的工艺精准建模监测，工业之脑需要在工业之眼的监测下精准建模与辨识物理运动学，辅助工业之手进行更加精准的制造过程控制。
为实现柔性适配与工艺精度的动态平衡，
工业之脑需要在工况变化时迅速重构产线与调度任务，而工业之眼需要在此变化条件下提供精准感知，反馈给工业之手以确保低精度产线的精准任务执行。
为达到通用技能与专门工艺的有机融合，工业之眼需要结合多模态工艺精准感知环境变化，工业之手在此基础上实现特定工艺的专门适应。
在上述基础上，本文同样对涵盖眼、手、脑三维能力，具备工厂环境适应性与柔性生产能力的人形机器人进行了讨论。
表~\ref{tab:application_comparison}对本文讨论的柔性制造典型应用场景及其技术需求进行了概览，帮助读者快速了解后续内容。
最后，归纳了各研究之间的关联，提出了面向柔性制造场景的“认知增强—技能跃迁—系统进化”三阶段演进模型，并对面向柔性制造的工业具身智能未来挑战与发展方向进行展望。

\begin{table*}[tbp]
  \centering
  \footnotesize
  \bicaption{\zh{柔性制造常见应用场景的技术需求与挑战}}{\zh{Comparison of technical requirements for common flexible manufacturing scenarios
  }}
  \label{tab:application_comparison}
  \setlength{\tabcolsep}{0.6em}
  \resizebox{\linewidth}{!}{
  \begin{tabular}{c|c|c}
    \toprule
    \textbf{应用场景} & \textbf{核心技术需求} & \textbf{主要挑战}\\
    \midrule
    高精度成像 & 亚毫米级成像、多视角快速配准、在线点云压缩与拼接 & 遮挡与高反射导致点云缺失、单一传感器视域受限、大数据量实时计算\\
    表面缺陷检测 & 自动化缺陷定位、尺寸量化、多尺度几何与纹理特征联合 & 微小裂纹与凹坑检测困难、表面纹理多样性、标注数据稀缺、伪影干扰\\
    多模态异常监控 & 视觉＋深度＋力／声／红外同步融合、毫秒级异常检测与告警 & 异构信号采样率不一致、噪声与遮挡干扰、跨模态对齐与语义一致性\\
    焊接工艺 & 熔池实时监控、电流/电压/送丝速度闭环调参、多物理仿真校准 & 高温高亮场景成像困难、熔池动态非线性、质量反馈稀疏与延迟\\
    打磨工艺 & 精确力控与表面粗糙度在线估计、磨具磨损补偿、路径自适应 & 不规则曲面接触不稳定、振动/噪声干扰、磨具性能随时间衰减\\
    装配工艺 & 亚毫米级对位、六轴力位混合控制、多零件协同装配 & 零件几何误差不确定、遮挡下状态不可见、夹紧/扭矩安全限值\\
    工厂排产调度 & 多目标（周期、能耗、延迟）动态优化、实时插单与故障弹性响应 & NP-hard 导致搜索空间爆炸、订单与设备状态不确定、跨设备调度\\
    移动单元路径规划 & 多 移动单元 实时避障与重规划、优先级及拥堵控制 & 狭窄通道死锁、动态障碍混行、通信延迟与同步可靠性\\
    物料仓储装箱 & 空间利用率最大化、在线决策、装箱连续稳定 & 到货顺序未知、物体形状多样、放置物理稳定性与安全碰撞约束\\
    \bottomrule
  \end{tabular}}
\end{table*}

\section{工业之眼（Industrial Eye）}

工业之眼旨在实现对制造环境与操作对象的精确感知与实时监控。面对柔性制造中多品种、小批量的挑战，其感知能力需在精度、鲁棒性和跨场景迁移方面持续提升。近年来，计算机视觉与图形学的进展为工业之眼提供了坚实支撑。
三维视觉技术具备亚毫米级成像、检测和测量能力，克服了二维视觉在复杂装配环境中精度不足、空间信息缺失、对环境变化敏感等问题，有效保障制造过程的一致性与精度；多模态感知系统融合视觉、触觉、声学等多源信息，增强了在动态、遮挡与噪声干扰下的感知稳定性与可靠性；此外，大规模预训练视觉模型具备强大的特征提取与泛化能力，使工业之眼能够快速适应新产品、新工艺和新场景，实现跨任务的零样本或少样本迁移。本节将围绕以上关键技术路径，系统阐述工业之眼如何全面感知生产环境，并通过典型案例展示其实际应用。

\subsection{三维视觉高精成像}

在柔性制造车间中，频繁切换的生产任务对缺陷检测与尺寸测量提出了更高要求，这些能力已成为实现闭环制造与保障产品质量的关键。然而，传统方法自动化程度低、效率不高、结果一致性差，且过度依赖人工经验，难以满足高节拍、高精度和全覆盖检测的需求。
以常用的三坐标测量法\ucite{hocken2012coordinate}为例，其在复杂结构件上适应性差，且测量程序与人工操作的依赖导致品种切换时效率低下。相比之下，三维视觉依托高分辨率传感器，可重建工件表面三维几何模型，结合几何特征分析实现自动化缺陷识别，并输出毫米级甚至亚毫米级的尺寸与形态参数，提升测量的效率、精度与一致性。
本节介绍三维视觉在高精成像、缺陷检测和尺寸测量中的关键技术。

\begin{figure}[h!t]
    \centering
    \includegraphics[width=\linewidth]{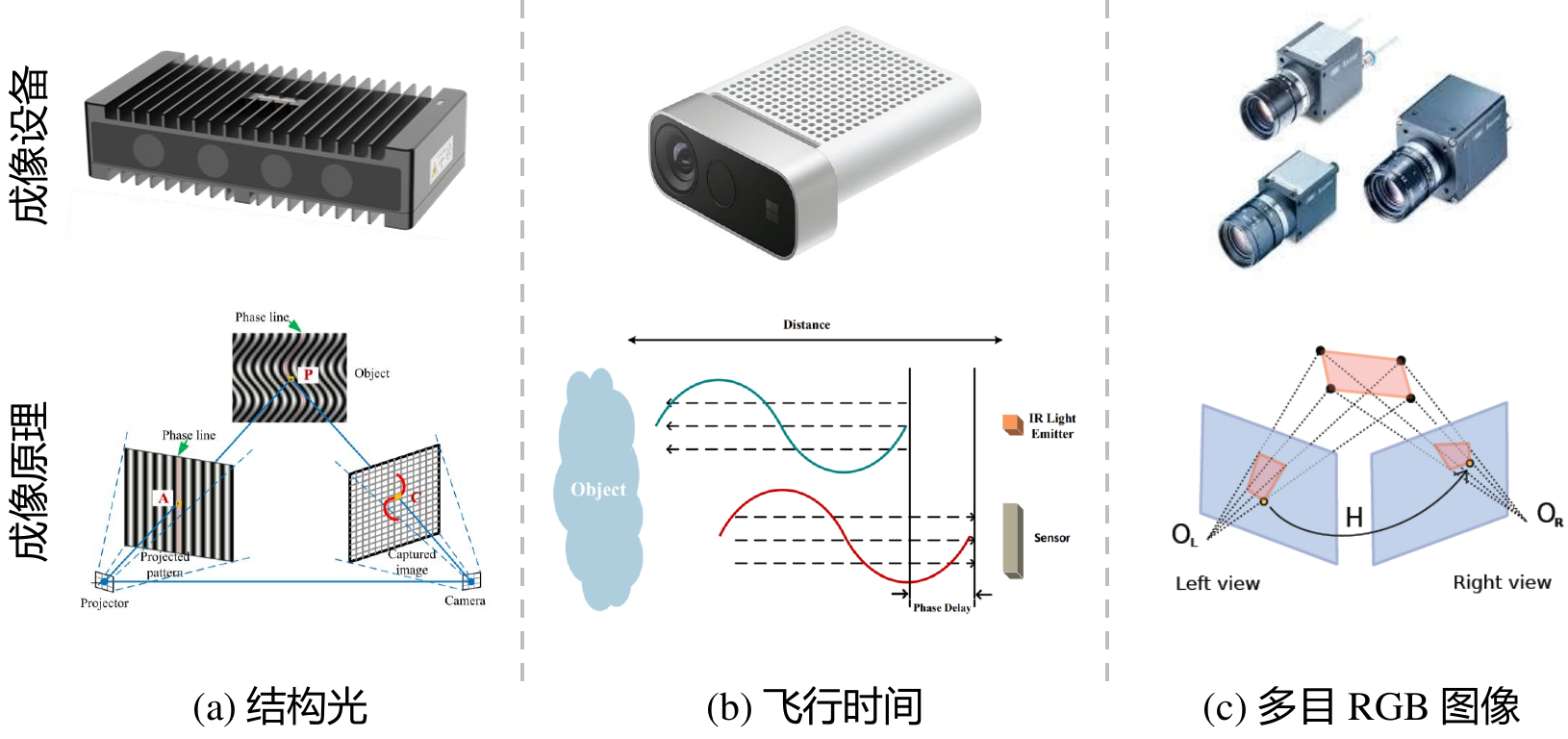}
    \bicaption{主流三维成像方法的工作原理\ucite{zhang2018high,horaud2016overview, moulon2017openmvg}}{Working principles of mainstream 3D imaging methods}
    \label{3D_vision_reconstruction}
\end{figure}

\paragraph{基于三维视觉的实时精准成像}
三维视觉通过传感器采集环境或物体的视觉数据，并将其转化为三维几何信息，实现对场景或物体的精确建模。相比二维图像，三维成像提供更丰富的空间信息，可更准确地识别复杂物体的形状、结构和位置关系，以实现精细缺陷检测和高精度尺寸测量，提升产品质量与制造精度。
近年来，三维成像技术发展迅速，方法多样、各有侧重。本文按成像原理将其分为结构光、飞行时间（Time of flight, TOF）和多目RGB图像三类，如图~\ref{3D_vision_reconstruction} 所示，各方法的性能对比见表~\ref{tab:reconstruction}。

\begin{table}[ht!]
   \bicaption{\zh{主流三维成像技术对比}}{\zh{Comparison of mainstream 3D imaging technologies}}
   \label{tab:reconstruction}
   \centering
   \footnotesize
   \setlength{\tabcolsep}{0.6em}
   \resizebox{0.5\textwidth}{!}{
   \begin{tabular}{c|c|c|c}
       \toprule
       \textbf{成像原理} & \textbf{结构光} & \textbf{飞行时间} & \textbf{多目RGB图像} \\
       \midrule
       精度               & 亚毫米级        & 厘米级            & 厘米级               \\
       测量范围           & 0.1m–5m         & 0.1m–10m         & 0.5m–50m            \\
       分辨率             & 高              & 中等              & 低                  \\
       实时性             & 中等            & 高                & 低                  \\
       硬件成本           & 中等            & 低至中等         & 极低                \\
       环境光干扰         & 敏感            & 中等              & 依赖环境光          \\
       适用材质           & 反射/漫反射表面 & 漫反射表面    & 依赖纹理            \\
       动态适应性     & 差              & 较强              & 差                  \\
       \bottomrule
   \end{tabular}
   }
\end{table}

基于结构光的三维成像通过向工件表面投射编码光（如条纹、点阵），结合光学三角测量计算因形变引起的光信号偏移，从而恢复表面三维坐标\ucite{sansoni2009state}。
该方法不依赖表面纹理，具备高精度与快速成像优势，但在远距测量或强光干扰下性能下降。
该技术可追溯至 Takeda 等人提出的傅里叶变换轮廓术（Fourier Transform Profilometry, FTP）\ucite{takeda1983fourier}，通过投射正弦条纹并进行相位傅里叶分析实现高密度重建，但存在相位包裹和光照敏感等问题。相移法（Phase-Shifting Profilometry, PSP）\ucite{zhang2006high}采用多帧相移恢复绝对相位，提高了深度精度。
多频相移算法\ucite{zhang2018high}则结合低频相位解包裹和高频解码，在少帧下实现高精度重建。
为适应动态场景，De Bruijn 序列编码\ucite{li20243d}利用单帧投射多种不重复图案，为每个像素提供唯一标识，支持运动物体快速扫描。
针对高反光或光滑材质，传统结构光效果受限，相位偏折术（Phase Measuring Deflectometry, PMD）\ucite{huang2018review}
通过投射正弦条纹并计算相位偏移以重建表面梯度，积分重建三维形貌，在高反射表面下更具精度与稳定性。
此外，线扫相机（Line-Scan Camera）\cite{shi2024methodology}也常用于三维成像。其原理是在工件表面投射一条激光或结构光线，线阵相机沿移动方向连续获取扫描线深度，拼接成完整三维模型，适用于PCB、卷材、金属管材等工件检测\ucite{ferreira2012low, van2015embedded, sun2016sensor}。

\begin{figure}[h!]
    \centering
    \includegraphics[width=\linewidth]{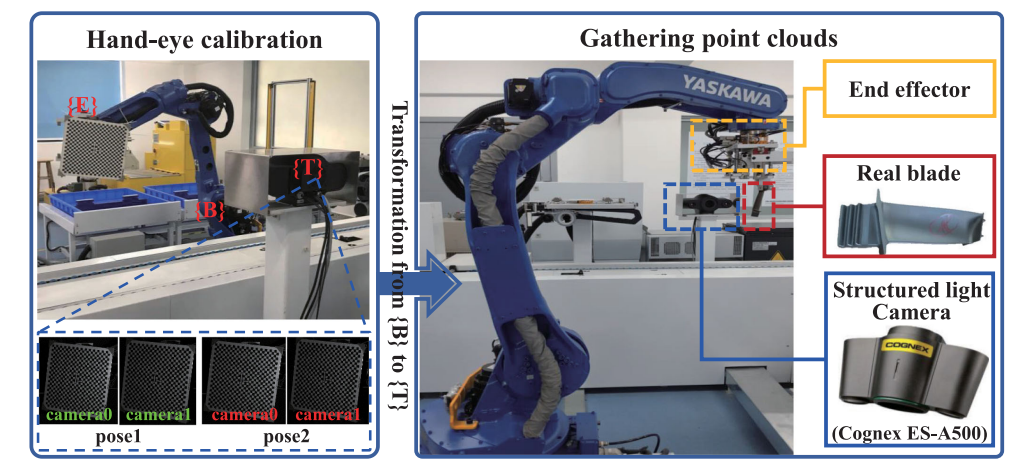 }
    \bicaption{多视图点云联合配准重建发动机叶片\ucite{peng2021stochastic}}{Multi-view point cloud joint registration for aero-engine blade reconstruction
    }
    \label{robot_reconstruction}
\end{figure}

基于飞行时间的三维成像通过发射调制光（通常为红外光），测量光从发射到被物体反射再返回的时间差或相位偏移，实时获取每个像素的深度信息。比如，激光雷达发射激光束并接收反射光，计算传播时间差，旋转扫描记录稀疏点云\ucite{raj2020survey}，具备强抗干扰能力，适用于远距和强光环境，但体积大、成本高、重建精度受限。
相比较之下，RGB-D 相机采用面阵传感器设计，无需机械扫描即可捕获全场景稠密深度图，硬件紧凑、适合实时应用。
KinectFusion\ucite{Newcombe2011KinectFusion}利用 GPU 加速的截断有符号距离函数（Truncated Signed Distance Function, TSDF）体素融合，实现实时稠密表面重建。ROSEFusion\ucite{Zhang2021ROSEFusion}通过粒子滤波优化提升相机高速运动下的重建稳定性。MIPS-Fusion\ucite{tang2023mips}结合多子图与梯度跟踪，实现大规模场景的在线重建。
多视角重建需对不同视角点云进行配准（Registration）\ucite{zhao2023registration}，如图~\ref{robot_reconstruction} 给出了多视图点云联合配准重建发动机叶片的实例。
Ge 等人\ucite{ge2021online}构建双相机系统，结合工件线性运动改进 ICP （Iterative Closest Point）算法，用于机器人喷涂场景快速在线建模。Peng 等人\ucite{peng2021stochastic}通过变参数图优化与无迹卡尔曼滤波提升航空叶片重建精度。
GeoTransformer\ucite{qin2022geometric}针对低重叠点云设计几何不变性编码，提出端到端配准方法，无需 RANSAC，实现百倍加速。

基于多目 RGB 图像的三维成像通过多个视角同步拍摄彩色图像，结合特征提取与匹配算法（如 SIFT\ucite{lowe2004distinctive}、ORB\ucite{rublee2011orb}等），利用相机参数与三角测量恢复三维坐标，并融合为完整点云模型。
其无需主动光源，可同时获取几何与纹理信息，但重建精度依赖于视角数量、相机标定和纹理质量。
多视图重建起源于运动恢复结构（Structure from Motion, SfM）\ucite{snavely2006photo}，先检测图像中的局部特征并进行跨视角匹配，再通过相机位姿估计与稀疏点云的三角化重建，恢复场景的三维结构。
在 SfM 提供初始相机姿态和稀疏点云基础上，多视图立体匹配（Multi-View Stereo，MVS）\ucite{furukawa2009accurate}用于生成稠密点云。
神经网络的引入提升了弱纹理和遮挡区域的重建能力。
典型如 MVSNet\ucite{yao2018mvsnet}使用多视角代价体（Cost Volume）进行深度预测，SuperGlue\ucite{sarlin2020superglue}通过图神经网络实现高质量特征匹配，提升了测量可靠性。
成熟工具如 COLMAP\ucite{schoenberger2016sfm, schoenberger2016mvs}、
OpenMVS\ucite{stathopoulou2019open}、AliceVision\ucite{griwodz2021alicevision}等，提供完整的图像建模流程。
近年来，神经辐射场（Neural Radiance Fields, NeRF）\ucite{mildenhall2021nerf}与高斯泼溅（Gaussian Splatting，GS）\ucite{kerbl20233d}利用隐式场景建模与体渲染实现高质量新视角合成，适用于离线渲染但几何精度与效率受限，在工业领域的应用较少。新兴基础模型 DUSt3R\ucite{wang2024dust3r}和VGGT\ucite{wang2025vggt}则跳过 SfM 等传统流程，直接从多图像快速重建场景，并可反推图像匹配与相机参数，为工业视觉任务提供高质量输入。

\paragraph{缺陷检测与尺寸测量}

基于三维成像结果，可提取表面缺陷与几何特征等关键信息，在柔性工业生产中搭建“感知”与“执行”之间的桥梁。精确检测有助于实时发现质量问题，几何特征提取支持定位与测量，提升效率与一致性。面对多品种、小批量需求，快速、准确反馈这些关键信息是实现工艺切换和高精制造的关键。

工业视觉系统中的缺陷检测针对制造过程中的物理异常（如裂纹、凹坑、装配错位等），通过定位表面瑕疵来防范经济损失与安全风险。
与传统二维图像方法\ucite{zhang2021mrsdi, zhou2019automated}相比，三维视觉引入深度信息，可准确反映表面几何变化，避免纹理与光照干扰，并量化缺陷的尺寸、深度与体积，支持后续修复与优化。
Auerswald等人\ucite{auerswald2019laser}基于激光线三角测量实现了微米精度的大尺寸齿轮全齿面三维重建，并支持微米级崩缺与划痕的精准识别。
Li等人\ucite{li2024novel}基于点云配准检测金属厚度波动，实现0.1mm级检测精度。
Yan等人\ucite{yan2025inner}通过密度聚类和区域生长算法实现精细分割缺陷点云，可以有效支撑管道内壁缺陷检测。
Huang 等人\ucite{huang2025entropy}设计熵驱动的邻域拟合算法，实现对磁性瓦等复杂曲面上亚毫米级裂纹的准确定位和拟合误差评估。
Vokhmintcev 等人\ucite{vokhmintcev2024fusion}提出 Fusion-ICP 算法，通过正交变换优化点云配准，适用于装配间隙与弯曲变形建模。

三维视觉测量通过提取物体表面的几何特征，实现高精度、非接触式尺寸检测，特别适用于重型机械、航空零部件等大尺寸工件。相比传统的三坐标测量机、激光跟踪器或超声波测厚等接触式或点式测量方式，三维视觉具备更高的灵活性与效率，能有效避免接触带来的刮擦与形变风险，并通过多视角融合实现复杂曲面的全覆盖测量。依托高分辨率传感器和精准标定，系统可在流水线或在线场景下实现微米级数据采集。
Yin等人\ucite{yin2014development}开发了集成结构光、立体成像与误差补偿模块的大型自由曲面扫描系统，实现 $\pm$0.2 mm 精度的非接触式测量。
Wang等人\ucite{wang2021mobile}利用机器人搭载立体视觉系统，结合手眼标定与位姿跟踪，完成风机叶片等部件的自动扫描。
Huang 等人\ucite{huang2022high}采用多视角相移结构光与特征约束配准，有效解决金属高光干扰，实现涡轮叶片的完整重建。
Ma等人\ucite{ma2023flexible}基于双线扫相机的高分辨率系统也被应用于发动机壳体等复杂结构的连续扫描，重建误差小于0.05mm，具备工业级稳定性，适用于高速检测与离线质控。

\paragraph{案例研究1：汽车漆面重建与缺陷检测}

在汽车制造过程中，漆面质量直接影响整车外观与市场竞争力。传统检测依赖人工，效率低、精度差，误检率可高达 15\%-20\%，高疲劳条件下漏检率甚至超过 30\%。
典型缺陷如尘粒、缩孔、橘皮、流挂等，尺寸多在 0.05-0.3mm 范围，难以稳定识别。
三维视觉的发展推动了漆面缺陷自动识别的实现，能自动识别微小缺陷，降低人为干扰，为漆面质量控制提供保障。

\begin{figure}[h!t]
    \centering
    \includegraphics[width=1\linewidth]{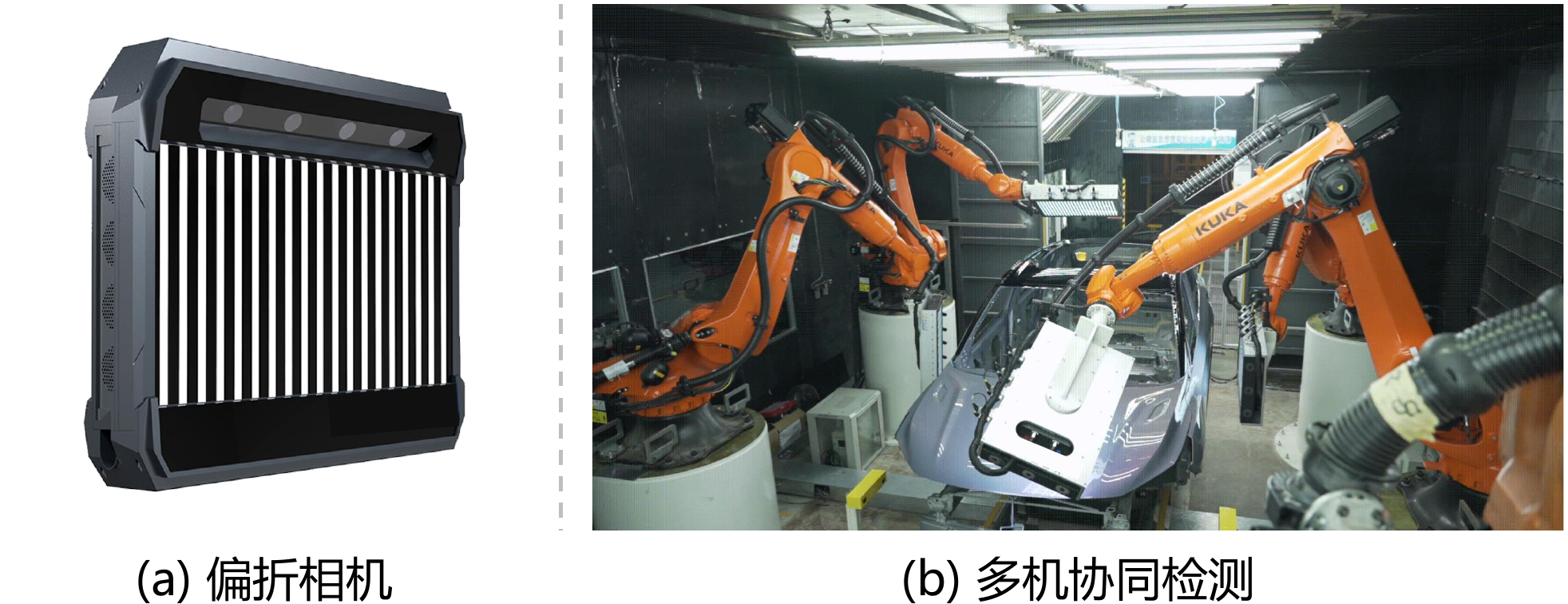}
    \bicaption{基于相位偏折术的汽车漆面重建}{Automotive paint surface reconstruction via phase measuring deflectometry 
    }
    \label{measurement_car}
\end{figure}

在漆面三维重建中，传统结构光或激光系统受高反射表面影响，易产生噪声与扫描空洞，难以满足微小缺陷定位需求。相比之下，基于相位偏折术 PMD 的重建技术通过测量反射光的相位变化，能在高反光条件下准确捕捉表面形貌。偏折相机如图~\ref{measurement_car}（a）所示。
由于车身远大于单个相机视野，采用多相机与多机器人协作可显著提升测量效率，通过路径规划实现整车漆面无盲区覆盖，如图~\ref{measurement_car}（b）所示。
面对高光或阴影遮蔽下的微小瑕疵，传统图像处理难以识别，可结合三维缺陷检测与分类方法，通过与车身模型配准，实现高精度缺陷定位与识别\ucite{ChinaVision2023PaintPro}。经视觉AI系统处理，漏检率可降至 <1\%，误检率 <3\%，为后续打磨、喷涂等修复工序提供精准支持。

\zh{\paragraph{案例研究2：船舶制造非结构化焊接场景成像}
在船舶制造中，焊接工序约占船体建造总成本的40\%\ucite{liu2024prediction}，当前仍以人工操作为主，存在效率低、质量不稳、成本高等问题，亟需引入智能化手段提升焊接效率与一致性。
在柔性制造环境中，焊接任务的尺寸精度要求通常在 $\pm 0.05$ mm 到 $\pm 0.5$ mm 之间。
然而，大型船舶如油轮、客轮或舰艇的零部件往往超过百万件，涵盖甲板板块、管路接头、支撑梁等形状各异构件，如图~\ref{fig:boat_welding}（a）所示。同时，焊缝结构复杂、尺寸巨大且缺乏标准化。
准确获取工件几何、焊缝位置及坡口角度与宽度等关键参数，才能为焊枪轨迹规划提供可靠的几何依据。

\begin{figure}[h!t]
    \begin{minipage}{0.46\textwidth}
    \centering
    \subfigure[在船舶制造业中，零部件种类繁多，作业场景结构复杂
    ]{
        \includegraphics[width=\linewidth]{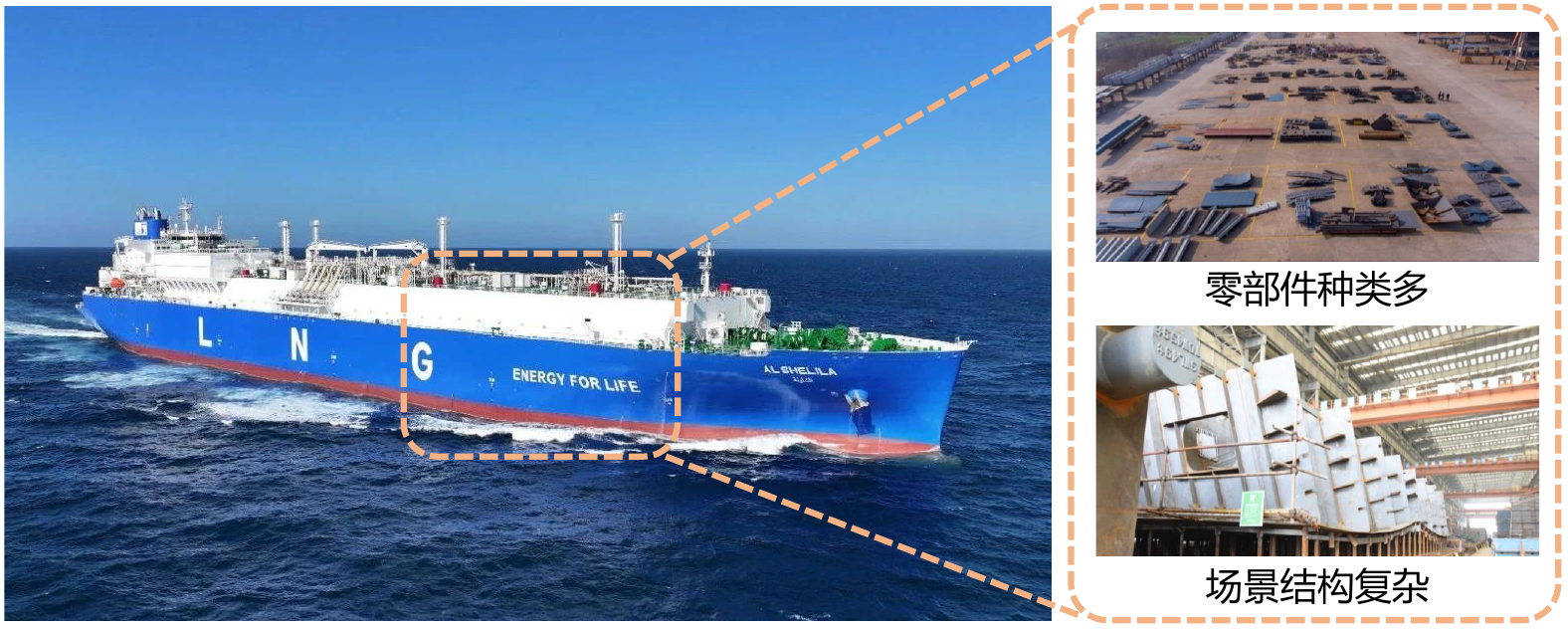}}
    \end{minipage}
    \begin{minipage}{0.46\textwidth}
    \centering
    \subfigure[\centering FoundationStereo 在多变场景无需微调精确估计深度]{
        \includegraphics[width=\linewidth]{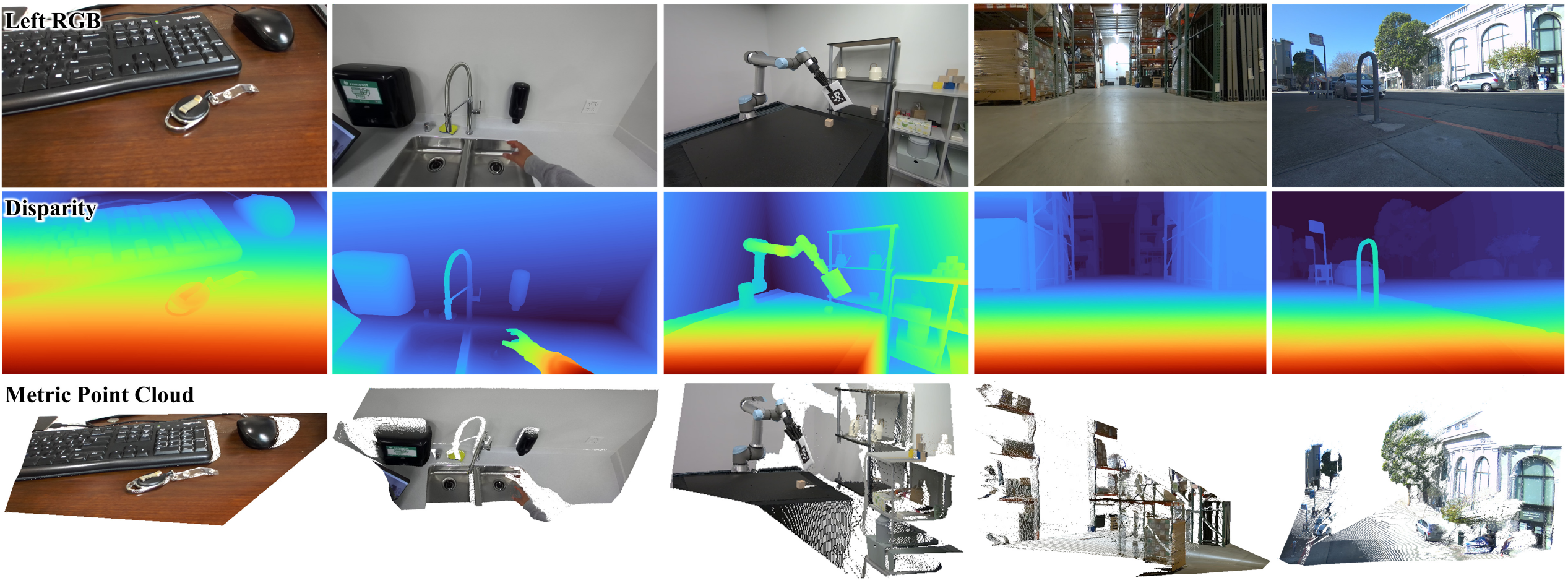}}
        \vspace{-2pt}
    \end{minipage}
    \bicaption{船舶焊接的快速精准成像需求\ucite{wen2025foundationstereo}
    }{Rapid and precise imaging requirements in ship welding
    }
    \label{fig:boat_welding}
\end{figure}
深度立体匹配（Deep Stereo Matching）通过分析多视角图像对，可快速生成像素级深度图并重建三维场景，具备轻量化采集和亚像素级精度，
在非结构化焊接场景中展现了良好的适配性。
然而，深度立体匹配依赖目标场景微调，难以适应焊接生产线的频繁切换。FoundationStereo\ucite{wen2025foundationstereo}提出了一种无需微调即可实现高精度稠密深度估计的方法（图~\ref{fig:boat_welding}（b）），通过百万级高保真合成图像自监督预训练，结合侧调优结构引入单目先验，并融合空间-视差一体的注意力机制，有效抑制遮挡与噪声干扰，实现 $\pm 0.2$ mm 级深度重建。该方法具备良好的泛化能力，可为柔性焊接中的工件识别、焊缝检测与路径规划提供稳定的三维感知基础。
}

\subsection{多源数据融合感知}

\begin{figure*}[h!t]
    \centering
    \includegraphics[width=\linewidth]{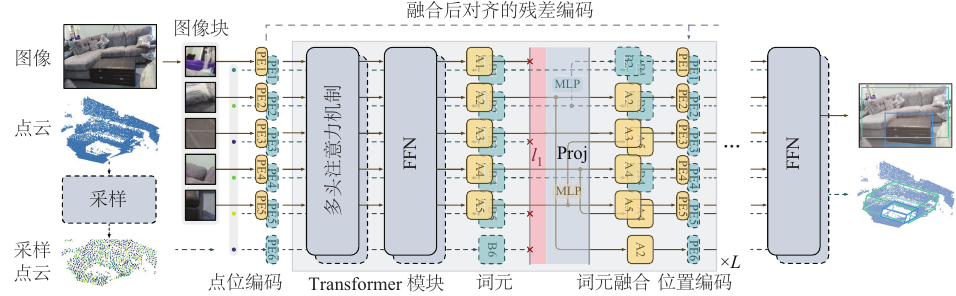}
    \bicaption{图像-点云多模态融合管线\ucite{wang2022multimodal}}{Pipeline for fusing image and point cloud modalities     }
    \label{mixt}
\end{figure*}

柔性制造中任务多样、工况复杂、切换频繁，对感知系统的鲁棒性与适应性提出更高要求。单一模态传感（如仅依赖视觉、触觉或声学）在遮挡、反光、表面变化或环境噪声下易出现信息缺失或误判，难以支撑复杂任务中的稳定感知与环境理解。
多模态感知融合技术\ucite{zhang2024multi, yuan2025survey}通过整合视觉、深度、力觉、声学与红外等信息，实现对象状态、环境约束与交互过程的综合感知，具备信息互补、抗干扰强、表达能力丰富等优势。图~\ref{mixt} 展示了图像与点云融合用于下游分割与检测的示例，表~\ref{tab:modality_comparison} 总结了常见模态的特性与适用场景。相比单模态系统，多模态融合能够提升系统在复杂工况下的稳定性与智能水平，为高精操作、异常识别与智能控制提供有力支撑。接下来将围绕特征提取、对齐与融合三方面展开介绍。

\paragraph{多模态特征提取}
工业多源传感数据具有明显的结构异构性：RGB / RGB-D 相机输出多通道图像，深度信息可转为点云或体素网格；力-扭矩、惯性测量单元（Inertial Measurement Unit, IMU）、音频、电压电流为高频一维时序数据；红外热像以灰度图表示温度分布。不同模态的数据特征需要被有效提取和表示，为后续的对齐、融合提供基础。
\begin{table}[ht!]
  \bicaption
    {\zh{常见工业传感模态特性总结}}
    {\zh{Summary of common industrial sensing modalities
}}
  \label{tab:modality_comparison}
  \centering
  \footnotesize
  \setlength{\tabcolsep}{0.7em}
  \resizebox{0.95\linewidth}{!}{
  \begin{tabular}{c|c|c|c}
      \toprule
      \textbf{模态} & \textbf{主要优势} & \textbf{主要劣势} & \textbf{典型应用场景}\\
      \midrule
      RGB 图像  &
      \makecell[l]{纹理与颜色信息丰富;\\公开数据与预训练模型充足;\\迁移与部署成本低} &
      \makecell[l]{对光照、遮挡敏感; \\缺乏深度、几何信息;\\纹理缺失材质鲁棒性差} &
       \makecell[l]{缺陷检测;\\装配定位;\\表面质量评估} \\
      \midrule
      深度图及点云 &
      \makecell[l]{直接表达三维几何，\\外观变化鲁棒;\\支持零样本泛化} &
      \makecell[l]{视角遮挡导致局部稀疏;\\设备与标定成本高;\\粉尘振动带来噪声} &
       \makecell[l]{机器人抓取;\\工件配准;\\空间占据分析} \\
      \midrule
      声学  &
      \makecell[l]{可穿透遮挡，诊断内部故障;\\传感成本低，布置灵活;\\预训练声学嵌入可复用} &
      \makecell[l]{车间噪声需降噪;\\频域特征可解释性弱;\\声源定位精度低} &
       \makecell[l]{齿轮故障诊断;\\焊接电弧音监测;\\轴承健康评估} \\
      \midrule
      红外热像 &
      \makecell[l]{不依赖可见光，\\适用高温或低照度;\\对热分布、裂纹空洞敏感} &
      \makecell[l]{分辨率低、噪声大;\\受环境温漂影响;\\热像与几何配准成本高} &
       \makecell[l]{焊缝熔池监控;\\热缺陷检测;\\温升分析} \\
      \bottomrule
  \end{tabular}
  }
\end{table}

在特征提取阶段，各模态依托不同的深度学习架构，各自侧重不同的特征优势。
对于 RGB / RGB-D 图像，深层卷积神经网络（如 ResNet\ucite{he2016deep}）擅长捕捉局部纹理与边缘特征，而基于自注意力机制的 Vision Transformer (ViT)\ucite{dosovitskiy2020image}则能够建模全局上下文信息。
深度图或点云常借助 PointNet\ucite{qi2017pointnet}及其层次化扩展 PointNet++\ucite{qi2017pointnet++}在无序点集中学习全局-局部几何特征，而 Point Transformer\ucite{zhao2021point}则进一步利用自注意力捕获长程依赖和方向信息。
对于力-扭矩、惯性测量单元、电流电压等高频一维信号，通常采用一维卷积神经网络（1D CNN）\ucite{kiranyaz2015real}、时序卷积网络（Temporal Convolutional Network, TCN）\ucite{bai2018empirical}或门控循环单元（Gate Recurrent Unit, GRU）\ucite{ChoMGBBSB14}，通过空洞卷积或门控机制高效建模时序依赖。
声学与振动数据先经傅里叶变换映射成 二维 频谱，再由卷积网络提取频域特征。
红外热像等灰度图像常采用 U-Net\ucite{ronneberger2015u}或双分支网络\ucite{simonyan2014two}，以保持细粒度特征，从而提升缺陷检出率和模型泛化能力。

尽管各模态已有成熟的特征提取方法，但其数据采集和训练时间的成本开销对工业应用仍然较高。在柔性生产环境中，模型需快速迭代以适应频繁切换的工艺与设备，
应用预训练模型可以有效降低训练成本。在 RGB 领域，已发布的自监督 ViT 模型 MAE\ucite{he2022masked}和 DINO\ucite{oquab2024dinov}提供了大规模预训练权重，少量 / 无需微调即可适配缺陷检测任务。
在三维点云方面，PointMAE\ucite{pang2022masked}与 PTv3\ucite{wu2024point}在 ShapeNet 等数据集上预训练后，对工业点云配准与抓取位姿估计展现出优异的零样本与少样本泛化能力。
声音与振动信号可直接利用在 AudioSet\ucite{gemmeke2017audio}上预训练的 PANNs\ucite{kong2020panns}和 YAMNet\ucite{howard2017mobilenets}，将频谱映射为通用声学嵌入并应用于工业故障诊断。
若需端到端时序特征，可引入 PatchTST\ucite{nie2023a}和 TimesNet\ucite{WuHLZ0L23}等预训练骨干网络，对工业时序数据进行快速微调。在热红外领域，InfMAE\ucite{liu2024infmae}基于百万级热像帧的自监督预训练模型，为热缺陷检测与工件温升监测提供了良好的初始化。
这些预训练库为工业场景的多源数据融合提供了可靠的通用表示，降低了标注和训练成本，满足了柔性生产系统对模型快速迭代与部署的需求。

\paragraph{异构特征对齐}
完成单模态特征提取只是多源感知的“第一公里”，深度特征的对齐与融合是进一步实现多模态感知的关键。
多模态对齐指将不同模态的特征映射到共享的语义空间，实现跨模态的语义一致性。而多模态特征融合则强调在完成对齐后，按照任务需求将各模态的互补信息进行层次化整合，生成单一模态无法提供的更完整、更可信的综合表征。

多模态特征对齐方法涵盖多种策略，包括对比学习、联合嵌入和注意力机制等，以适应不同模态间的分布差异和结构差异。
对比学习通过拉近同源正样本、推远异源负样本，将不同模态映射到共享语义空间。代表性的 CLIP\ucite{radford2021learning}将图像-文本特征紧密对齐，并在零样本缺陷识别中展现强大迁移能力\ucite{ma2025aa, cao2024adaclip}。
联合嵌入方法则侧重于设计投影头或显式潜变量（如深度典型相关分析\ucite{andrew2013deep}、变分联合分布\ucite{wu2018multimodal}）来学习共同的潜在表示，从而消除模态间维度和尺度差异。此类方法已在多源传感融合与跨模态检索中验证了高效性和可解释性\ucite{mithun2019joint}。

注意力机制\ucite{vaswani2017attention}能够在词元（Token）级特征中显式捕获模态间的对应关系和互补依赖，既保留局部细节又兼顾全局语义，已被广泛应用。
MulT\ucite{tsai2019multimodal}是最早系统引入跨模态注意力机制的代表性工作，该方法设计了 Crossmodal Attention 模块，通过 Transformer 将音频、视觉与文本等不同模态序列直接对齐，无需显式同步时间步。
EMT\ucite{sun2023efficient}通过双层恢复模块实现了对缺失模态的重建与对齐，特别适用于实际中数据不完整的多模态场景。
AnyGPT\ucite{ZhanDYZZLZYZL0F24}引入“任意到任意”多模态对话能力，将大语言模型相关的所有模态（音频、文本、图像）统一映射为离散词元序列，无需改动现有模型结构或训练范式。
TEAL\ucite{yang2023teal}提出“Tokenize and Embed All”策略，将任意模态离散为词元序列并映射到共享嵌入空间，继而以自回归方式生成输出，使冻结的大语言模型既保留文本能力，又可高效处理多种非文本模态。
M2PT\ucite{zhang2024multimodal}提出一种跨模态路径增强方法，构建模态无关的跨模型参数共享机制，实现异构模态知识迁移。
Meta-Transformer\ucite{zhang2023meta}通过统一的框架展示了其在多模态学习中的潜力，支持高光谱图像、音频、视频、时间序列、点云等多种模态的输入。

\paragraph{多模态特征融合}
在完成异构特征对齐后，多模态融合旨在有机整合各模态信息，以构建更全面、鲁棒的表征。融合过程中需确保时空与语义对齐，发挥不同模态的互补优势，避免冗余与信息丢失。同时兼顾实时性与计算复杂度，并在单一模态失效或噪声干扰时保持整体性能。
根据融合的时机和方式，可将常见的多模态特征融合方法分为早期融合（Early Fusion）、晚期融合（Late Fusion）和中间融合（Intermediate Fusion）\ucite{zhao2024deep}。不同的特征融合方法在工业应用中各有优势与局限，如表~\ref{tab:fusion_strategy_comparison} 所示。

\begin{table*}[ht!]
  \bicaption
    {\zh{多模态特征融合方法优缺点对比}}
    {\zh{Advantages and disadvantages of multimodal feature fusion methods}}
  \label{tab:fusion_strategy_comparison}
  \centering
  \footnotesize
  \setlength{\tabcolsep}{0.7em}
  \resizebox{0.95\linewidth}{!}{
  \begin{tabular}{c|c|c|c}
      \toprule
      \textbf{融合方法} & \textbf{主要优势} & \textbf{主要劣势} & \textbf{典型适用场景}\\
      \midrule
      早期融合 &
      \makecell[l]{  信息完整，易捕捉跨模态互补关系;\\  端到端实现简单，单次前向即可融合} &
      \makecell[l]{  特征维度膨胀，显存与计算开销大;\\  对时空同步要求高，同步误差易放大} &
      \makecell[l]{高精装配等视觉-力觉强耦合场景，\\模态数量有限且同步精度高} \\
      \midrule
      中间融合 &
      \makecell[l]{  多层交互捕捉深层依赖，协同充分;\\  交互深度与算子可灵活定制} &
      \makecell[l]{  网络设计复杂，超参调优成本高;\\  多轮交互增加参数量与推理时延} &
      \makecell[l]{工业异常检测、动态决策控制等需\\细粒度协同和高鲁棒性的任务} \\
      \midrule
      晚期融合 &
      \makecell[l]{  模块化强，子系统可独立训练替换;\\  单模态失效时系统容错性好
      } &
      \makecell[l]{  缺乏细粒度交互，互补信息利用不足;\\  融合权重或门限依赖经验，不易全局最优} &
      \makecell[l]{多源检测报警、容错决策等模态\\耦合度低且需易扩展维护场景} \\
      \bottomrule
  \end{tabular}
  }
\end{table*}

在早期融合中，不同模态的数据被直接拼接组合，形成一个统一的特征表示后输入模型进行处理。这种方法简单直观，能够在感知阶段就捕捉模态间的全局关系。
在工业机器人装配任务中，视觉与力觉信息的早期融合可以帮助机器人在环境中准确定位物体并进行高精度操作。Lee等人\ucite{LeeZZTSSFGB20}通过跨模态对比学习提取力觉-视觉数据的紧凑联合表征，将预训练表征迁移至策略网络，在钉孔装配任务中实现孔洞形状、装配间隙及外部扰动的强鲁棒性。
然而，早期融合存在着特征维度过高、计算复杂度增加的问题，特别是在处理高分辨率视觉与低采样频率力觉数据时，可能导致效率低下。

晚期融合先独立处理每个模态的数据，得到各自的预测结果，再通过加权平均、投票等方式将这些结果融合成最终决策。
在多模态机器人打磨系统中，图像和声学传感器分别收集打磨过程中的音频信号，并将两种模态的处理结果进行融合\ucite{nakajima2023robotic}。当音频反馈的RMS功率小于阈值时，结合视觉反馈测量的粉末区域半径来做决策，即如果当前半径小于或等于之前测量的半径，就决定收集粉末；否则继续研磨。
晚期融合模块化设计，易于实现和调试，但可能忽略模态间的交互信息，尤其是在需要高度协同的信息融合任务中，可能无法充分发挥各模态的互补优势。

中间融合在模型的中间层对不同模态的特征进行融合，通常采用注意力机制、门控机制等方式进行交互。在多模态工业异常检测中，视觉与力觉信息的中间融合能够有效捕捉到细粒度的交互信息，使机器人能够在动态环境中做出精准的决策。M3DM\ucite{wang2023multimodal}提出融合RGB图像与三维点云的多模态特征，采用块级对比学习促进模态交互、减少干扰，并通过多记忆库存储不同模态特征以避免信息丢失，最终基于多记忆库决策，在MVTec-3D AD\ucite{BergmannJSS22}工业异常检测数据集上性能领先。
然而，中间融合的设计复杂度较高，需要仔细调整融合策略，增加了系统的开发难度和计算开销。

\paragraph{案例研究 1：多模态融合零件装配}
零件装配等接触密集型（Contact-Rich）任务因操作过程不确定性大，在工业场景中极具挑战。零件几何形状的多样性（如异形插销）、装配间隙的微小差异（通常仅为 0.1 mm $-$ 0.5 mm），以及接触状态的瞬时变化（如碰撞、滑动）都会增加操控难度。
以汽车门铰链装配为例，其配合公差需控制在 $\pm 0.05$ mm 否则易造成卡滞或松动。单一模态感知（如仅依赖视觉或触觉）难以稳定应对遮挡、光照变化或空间感知缺失问题。多模态融合可弥补各模态局限，更全面理解任务环境，提升机器人在复杂装配任务中的鲁棒性。

\begin{figure}[h!t]
    \centering
    \includegraphics[width=\linewidth]{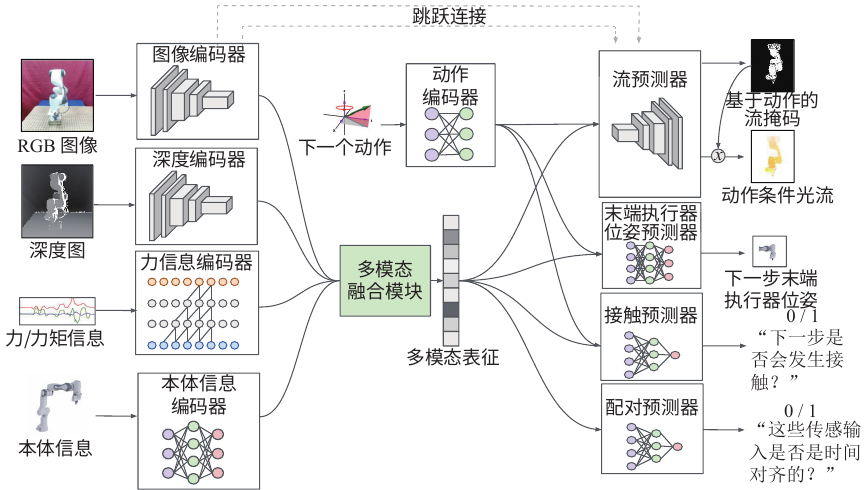}
    \bicaption{自监督多模态表征学习架构\ucite{LeeZZTSSFGB20}}{Architecture of self-supervised multimodal representation learning
    }
    \label{multimodal_observation}
\end{figure}

斯坦福大学（Stanford University）研究团队\ucite{LeeZZTSSFGB20}在装配任务中融合视觉（RGB-D 图像）、触觉（六轴力-扭矩）与本体感知（末端位置与速度），提升状态表示的完整性，如图~\ref{multimodal_observation} 所示。为实现高效融合，采用基于变分自编码器的架构，通过专家乘积（Product of Experts）\ucite{wu2018multimodal}将不同模态的潜在表示联合建模。
模型还引入光流估计与接触事件判断等自监督任务，以捕捉模态间的动态关联，为策略学习提供紧凑且语义丰富的输入。该方法提升了装配性能，实现了跨形状迁移与强抗干扰能力，验证了其在真实机器人系统中的鲁棒性。这表明，多模态感知结合自监督学习可降低对人工标注的依赖，为工业具身智能在复杂动态环境中的落地提供支持。

\paragraph{案例研究 2：多模态融合焊接质量监测}
焊接过程中需实时监测状态并提前预测熔透不足、烧穿、错位等缺陷，以避免成品瑕疵。传统单模态传感器存在感知盲区：视觉易受弧光干扰误差超 1 mm；声学受噪声影响信噪比常低于 10 dB；电流/电压传感器仅能反映热输入信息，难以准确预测复杂的熔池与焊缝几何。
多模态融合\ucite{yu2024monitoring, ji2024multi}成为解决这些挑战的关键，通过整合视觉、声学、电信号等信息，可弥补单源局限，提升预测的鲁棒性与泛化能力，实现焊接过程的高效感知与预警。

\begin{figure}[h!t]
    \centering
    \includegraphics[width=\linewidth]{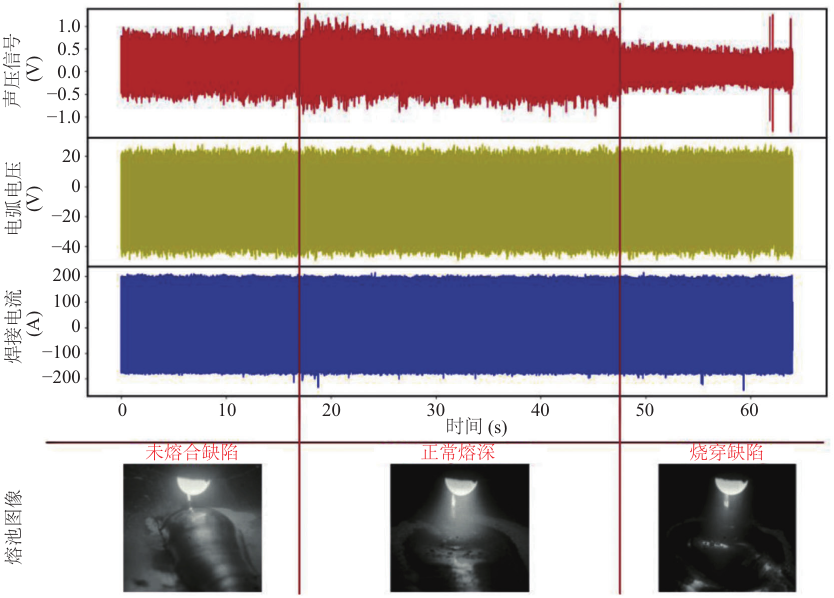}
    \bicaption{多源异构信息融合预测焊接质量\ucite{chen2021prediction}}{Fusion of multi-source heterogeneous information for welding quality prediction
    }
    \label{weld_multimodal_observation}
\end{figure}

上海交通大学研究团队\ucite{chen2021prediction}设计了面向弧焊过程的多模态特征提取与融合架构，以提前预测焊接质量缺陷并指导工艺调整。
该架构涵盖视觉、声学、电信号三类模态，如图~\ref{weld_multimodal_observation} 所示。视觉模态通过CNN处理焊池图像，提取熔池面积、对称性等特征。声学模态采用时域与频域分析相结合的方法，提取平均能量、幅值、标准差及时频段统计特征。电信号模态则提取电流、电压的时域统计特征。这些异构特征通过归一化处理后输入LSTM网络进行融合，利用其时间序列建模能力捕捉动态关联。在混合模态输入下，模型可提前0-2秒预测熔透不足、烧穿等缺陷，为工艺调整争取关键时间窗口。

\subsection{工业视觉基础模型}

在柔性制造场景中，产品迭代快、工艺流程多变、环境条件复杂，对视觉系统提出了高度的泛化性与适应性要求。传统的视觉模型多以任务为中心进行定制训练，难以在跨场景、跨工艺之间灵活迁移，导致模型复用性低、维护成本高。为此，工业视觉基础模型被广泛关注。通过大规模预训练，基础模型具备少样本适应、跨任务泛化和多模态扩展等能力，能够有效支撑多种视觉任务在柔性生产中的快速部署与稳定运行。借助基础模型构建视觉通用能力中枢，有望成为推动工业感知系统向智能化与平台化演进的重要方向。

\paragraph{视觉基础模型的学习与泛化}
视觉基础模型（Vision Foundation Models, VFM）是在大规模图像或视频数据上预训练，具备广泛适应性与强泛化能力的视觉模型。它们通过一次预训练即可支持图像分类\ucite{radford2021learning, caron2021emerging, oquab2024dinov}、目标检测\ucite{liu2023grounding, wang2023detecting}、图像分割\ucite{kirillov2023segment, ravi2024sam}、三维理解\ucite{wang2024dust3r, zhu2023ponderv2, 3D-VLA}等任务，降低任务特定模型的开发门槛，实现“即插即用”的视觉智能。

自监督学习已成为视觉基础模型预训练的主流方式。它通过设计无需人工标注的预训练任务，让模型从大规模无标签数据中学习特征。主流方法包括掩码图像建模（Masked Image Modeling, MIM）、对比学习、以及教师-学生框架。
MIM 受 BERT\ucite{devlin2019bert}启发，通过随机遮挡图像区域并重建，提升了模型语义建模能力，尤其适用于视觉Transformer架构。
对比学习方法如 SimCLR\ucite{chen2020simple}、MoCo\ucite{he2020momentum}、BYOL\ucite{grill2020bootstrap}通过构造正负样本对，学习具有判别力的全局语义特征，强调对图像整体结构与语义的建模，提升跨任务迁移能力。
教师-学生方法（如 DINO\ucite{caron2021emerging}）借助知识蒸馏\ucite{hinton2015distilling}机制，将教师模型知识迁移至轻量学生模型，在避免使用负样本的同时，仍可学得高质量表征，具备良好的稳定性与泛化性。

视觉基础模型的泛化能力则确保模型在面对多样化任务时保持强适应性。该机制主要体现在三个方面：
（1）统一任务建模：通过构建通用输入-输出对，模型可一体处理分类、分割、检测等任务。例如 LaVin-DiT\ucite{wang2024lavin}使用联合扩散Transformer和空间-时间变分自编码器，实现对20+任务的支持；
（2）上下文学习：模型通过引入任务描述或示例，实现零样本或少样本的任务迁移，MetaVL\ucite{monajatipoor2023metavl}首次将语言模型中的上下文学习迁移至视觉语言模型，实现紧凑模型的高效适配；
（3）多任务学习与共享表示：如 Uni-Perceiver v2\ucite{li2023uni}使用统一的最大似然策略处理视觉与视觉语言任务，在未经微调的情况下也能表现跨模态泛化的基础能力。

\paragraph{从二维到三维的视觉基础模型}
在二维视觉任务中，SAM（Segment Anything Model）\ucite{kirillov2023segment}分割模型是视觉基础模型的重要突破，
实现了基于点、框、文本等多种提示的零样本分割，具备良好的模块化与任务扩展性。
其升级版本 SAM 2\ucite{ravi2024sam}支持视频输入，借助流式内存 Transformer 和交互式数据引擎，实现了 4K 视频的实时分割（30 fps）。
Depth Anything\ucite{yang2024depth}深度估计模型通过教师-学生架构，在无标签图像上生成伪标签，结合 Vision Transformer 实现鲁棒的单目深度估计，其 V2 版本\ucite{yang2024depthv2}引入合成数据训练教师模型，提升了深度预测精度。
在二维图像位姿估计方面，NVIDIA 的 FoundationPose\ucite{wen2024foundationpose}提供统一的 6D 位姿估计与跟踪框架，支持 CAD 或图像输入，通过神经隐式表示与语言模型策略，在对比学习中实现跨物体泛化。
在视觉特征统一表征方面，Meta AI 的 DINOv2\ucite{oquab2024dinov}通过改进的多尺度 ViT 利用全局与局部对比、视角抖动以及知识蒸馏协同训练，产出在分割、目标检测、关键点估计等多种下游任务上的强泛化特征。
在多视图预训练方面，斯坦福大学与 Google Robotics 提出的 3D-MVP\ucite{qian20243d}构建大规模多视角 RGB-D 语料库，联合跨视图对比与几何一致性重建，捕捉语义与几何特征，在姿态估计与装配任务中带来显著的性能提升。

随着二维基础模型的成熟，其思想正扩展至三维视觉领域。这类模型不再依赖传统的点云卷积或体素处理，而是通过多层注意力机制建立空间与语义间的对齐。3D-VisTA\ucite{zhu20233d}、RangeViT\ucite{ando2023rangevit}与 UniT3D\ucite{chen2023unit3d}展示了 Transformer 在三维视觉-语言任务中的强大表达能力，体现出多任务统一、模态融合与语义泛化的趋势。
在三维视觉建模方面，DUSt3R\ucite{wang2024dust3r}引入点图（Pointmap）表示，无需相机参数即可从图像中预测三维结构与相对位姿，简化了多视角重建流程。VGGT\ucite{wang2025vggt}在此基础上构建端到端多任务框架，可从单张或多张图像同时预测相机参数、深度、点图与三维跟踪特征，并通过视觉词元与相机词元的显式交互建模图像间几何一致性，进一步提升了三维任务的性能与效率。

\paragraph{模型微调机制}
尽管基础模型在通用任务中具有良好泛化能力，但在工业特定任务中仍需微调以更好满足精度和实际需求。目前主流微调方法包括全量微调（Full Fine-tuning）\ucite{achiam2023gpt}、线性探测（Linear Probing）\ucite{radford2021learning}与参数高效微调（Parameter-Efficient Fine-Tuning，PEFT）\ucite{guo2025deepseek}。
全量微调适用于资源充足、追求极致性能的场景；线性探测仅训练任务头，适合快速适配或资源受限环境。PEFT 则在性能与效率之间取得平衡，逐渐成为主流，尤其适用于多任务或边缘设备部署。代表方法包括 	低秩适配 （Low-Rank Adaptation，LoRA）\ucite{hu2022lora}、适配器微调（Adapter Tuning）\ucite{houlsby2019parameter}和提示词微调（Prompt Tuning）\ucite{lester2021power, jia2022visual}。LoRA 参数开销小，适合大模型低资源场景；Adapter Tuning 具备良好的跨任务迁移能力；Prompt Tuning 通过输入引导实现最小代价适配。三者提供了灵活的微调策略选择，正逐步成为工业多任务系统的标准工具。不同 PEFT 方法的性能对比总结在表~\ref{tab:finetuning} 中。

\begin{table}[!ht]
\centering
\bicaption{常用参数高效微调方法对比}{Comparison of common PEFT methods}
\label{tab:finetuning}
\resizebox{\linewidth}{!}{ 
\begin{tabular}{ccccc}
\hline
\textbf{微调方法}& \textbf{训练参数比例}& \textbf{训练成本}& \textbf{适用场景}\\ \hline
全量微调 & 高 & 高 & 高性能需求、数据充足的任务 \\
线性探测 & 极低 & 低 & 特征评估、快速原型开发 \\
参数高效微调 & 低 & 低 & 多任务学习、资源受限的环境 \\ \hline
\end{tabular}}

\end{table}

\paragraph{案例研究1：基于视觉基础模型的少样本缺陷检测}
缺陷检测是工业制造保障产品质量的核心环节。\zh{然而，实际工业场景中的缺陷数据往往呈现长尾分布，常见良品样本则占到 95\% $-$ 99\%，且缺陷类型复杂多样（如划痕、污渍、形变等），标注代价高昂。}
传统的深度学习等方法依赖大量精标注数据\ucite{yang2025survey}，难以满足工业场景对算法可泛化及快速部署的需求。视觉基础模型具备强大迁移能力，能够提取高质量视觉表示、迅速适配目标任务，提升缺陷检测性能。
\begin{figure}[!h]
    \centering
    \includegraphics[width=1\linewidth]{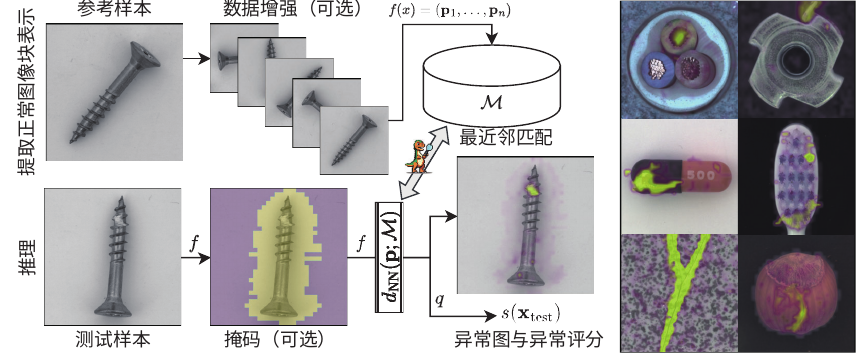}
    \bicaption{基于AnomalyDINO的工业缺陷检测\ucite{damm2025anomalydino}}{Industrial defect detection using AnomalyDINO}
    \label{fig:AnomalyDINO}
\end{figure}

AnomalyDINO\ucite{damm2025anomalydino}通过引入视觉基础模型DINOv2\ucite{oquab2024dinov}，实现了高效少样本异常检测，如图~\ref{fig:AnomalyDINO} 所示。该算法基于DINOv2提取的二维视觉特征构建记忆库，同时设计随机旋转以增强少样本记忆库的泛化性。方法利用DINOv2的零样本分割能力生成对象掩码以剔除背景噪声，以便定位工业图像的异常区域，在多个工业检测数据集上均实现性能突破。视觉基础模型具备几何感知与语义理解能力，所提取的预训练视觉特征能够有效区分微小缺陷与正常纹理变化。同时，视觉基础模型兼具特征通用性与适配灵活性，能够快速部署于工业场景，构建高效的缺陷检测系统。

\paragraph{案例研究2：工业弱纹理图像匹配 }
图像匹配是工业视觉检测的关键前置环节，支撑目标定位、缺陷识别和三维重建等任务。其核心在于跨视角、时间或条件下，准确对齐图像中的对应区域或特征点，提供精确几何基础。
近年来基于深度学习的匹配方法得到发展。 SuperPoint\ucite{detone2018superpoint}通过联合学习特征与匹配策略提升鲁棒性，LoFTR\ucite{sun2021loftr}和 COTR\ucite{jiang2021cotr} 引入 Transformer 提升全局建模能力。
然而，在弱纹理、重复图案或结构高度相似的工业场景下，特征点难以区分，局部匹配信息不足，传统方法易失效，匹配准确性仍然面临严峻挑战。
\begin{figure}[h!]
    \centering
    \includegraphics[width=1\linewidth]{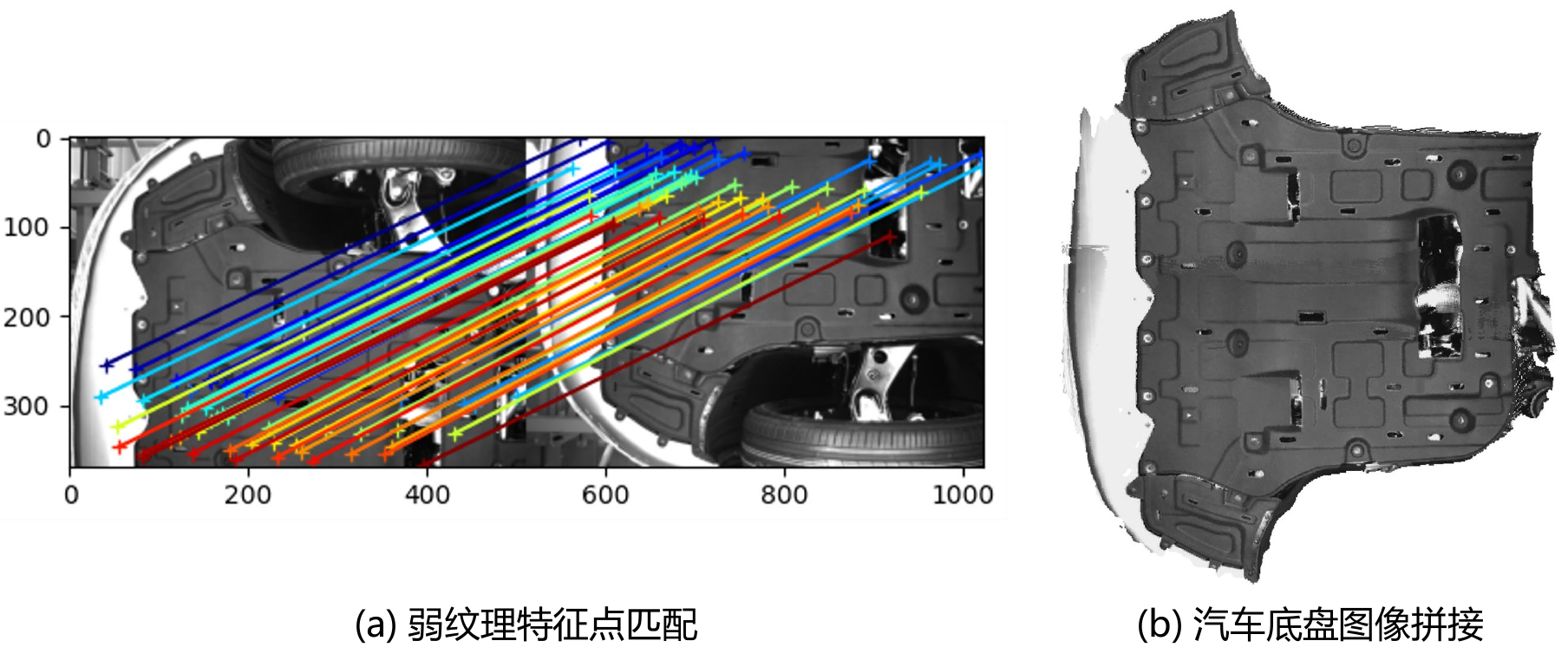 }
    \bicaption{DUSt3R 在弱纹理汽车底盘图像上的拼接示例}{Example of DUSt3R stitching on low-texture automotive-chassis images}
    \label{fig:matching_case_study}
\end{figure}

DUSt3R\ucite{wang2024dust3r}以三维重建为目标，也在图像匹配任务中展现出优异表现。其核心创新在于引入 Pointmap 表示，将像素映射至三维空间，实现无需相机参数的跨图像配准。不同于传统二维特征方法，DUSt3R 从三维视角理解图像，嵌入全局一致的空间框架，增强了特征的几何稳定性。
Pointmap 可编码法线、深度一致性与空间邻接等几何约束，即使在弱纹理场景下也能保持匹配的结构连续性与位置稳定性。相比仅依赖图像平面局部特征的传统方法，DUSt3R 具备更强的跨视角不变性和结构对齐能力，支持高质量匹配与点云生成，简化检测流程并提升效率。如图~\ref{fig:matching_case_study} 所示，在汽车底盘检测中，DUSt3R 实现了弱纹理图像的精确拼接，展现出其在工业场景中的应用潜力。

\section{工业之手（Industrial Hand）}

\begin{figure}[h!t]
    \centering
    \includegraphics[width=1.0\linewidth]{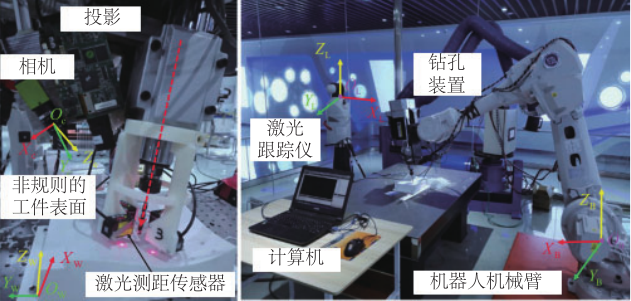}
    \bicaption{机器人在不确定环境中高精度钻孔\ucite{rao2019fringe}}{High-quality robotic drilling in uncertain environments
    }
    \label{fig:drilling}
\end{figure}

工业之手通过机器人等自动化设备完成精密操作，是制造任务的核心执行体。随着制造向柔性化、小批量、多品种转型，产线布局、产品类型与工艺流程日益动态，传统依赖高精度、固定流程的控制模式已难以适应，工业系统正逐步采用低成本、模块化硬件。
这些趋势对工业之手提出更高要求：在硬件精度受限下仍需实现高精度操作；在产线快速重构中具备策略的柔性适应与迁移能力（图~\ref{fig:drilling}）；在复杂工艺约束下可基于感知反馈动态调节参数，保障过程稳定与产品一致。
从提升工业之手的柔性生产能力角度出发，本节围绕低精度硬件精准控制、可变环境下柔性操作与自适应工艺参数调控展开讨论。

\subsection{低精度硬件精准控制}
柔性制造要求快速重构产线，往往以低成本、低精度硬件为代价。为保障系统稳定性与加工精度，需依赖高精度感知与控制技术进行补偿，本节围绕操作对象位姿精确识别、控制策略主动适应、虚拟到现实迁移三个方面讨论对控制精度的补偿方法。

\paragraph{操作对象位姿精确识别}
位姿识别\ucite{haralick1989pose}是提升控制精度的基础。通过从观测数据中估计目标的空间位置与姿态，系统可动态调整操作路径、夹持方式与交互策略，实现高精度、强鲁棒性的智能操作\ucite{bauer2024challenges}。
在多品种、小批量场景下，准确位姿识别尤为关键，能为不同工件提供实时、可泛化的几何支持。例如，Guo 等人\ucite{guo2024vision}基于点云与位姿优化生成稳定的打磨轨迹。
工业场景中位姿识别方法主要可分为基于图像的二维方法与融合几何结构的三维方法，下文将分别进行介绍。 

基于二维图像的位姿识别通常依赖 RGB 图像，通过深度神经网络直接回归物体的六自由度（6DoF）姿态，常采用有监督训练，结合合成数据（由 CAD 模型生成）\ucite{tremblay2018deep, thalhammer2019sydpose}或真实标注数据 \ucite{hodan2017t, xiang2017posecnn}学习旋转与平移参数。
DOPE\ucite{zhang2021practical}结合深度网络与 PnP 算法实现快速解算，并通过双重平移校正提升在光照变化与背景干扰下的鲁棒性。
针对类别内物体形状差异较大时导致姿态估计失准的问题，SOCS\ucite{wan2023socs}提出语义关键点引导的形变对齐方法，结合坐标注意力与扩散生成结果筛选，提升复杂工件匹配的一致性与精度。
Wan等人\ucite{wan2025equivariant}进一步引入扩散模型与 A5 群等变结构，联合估计物体姿态与几何形状，具备处理遮挡、歧义等不确定性场景的能力，即便输入不完整也能稳定输出多组姿态解。

\begin{figure}[h!t]
    \centering
    \includegraphics[width=1.0\linewidth]{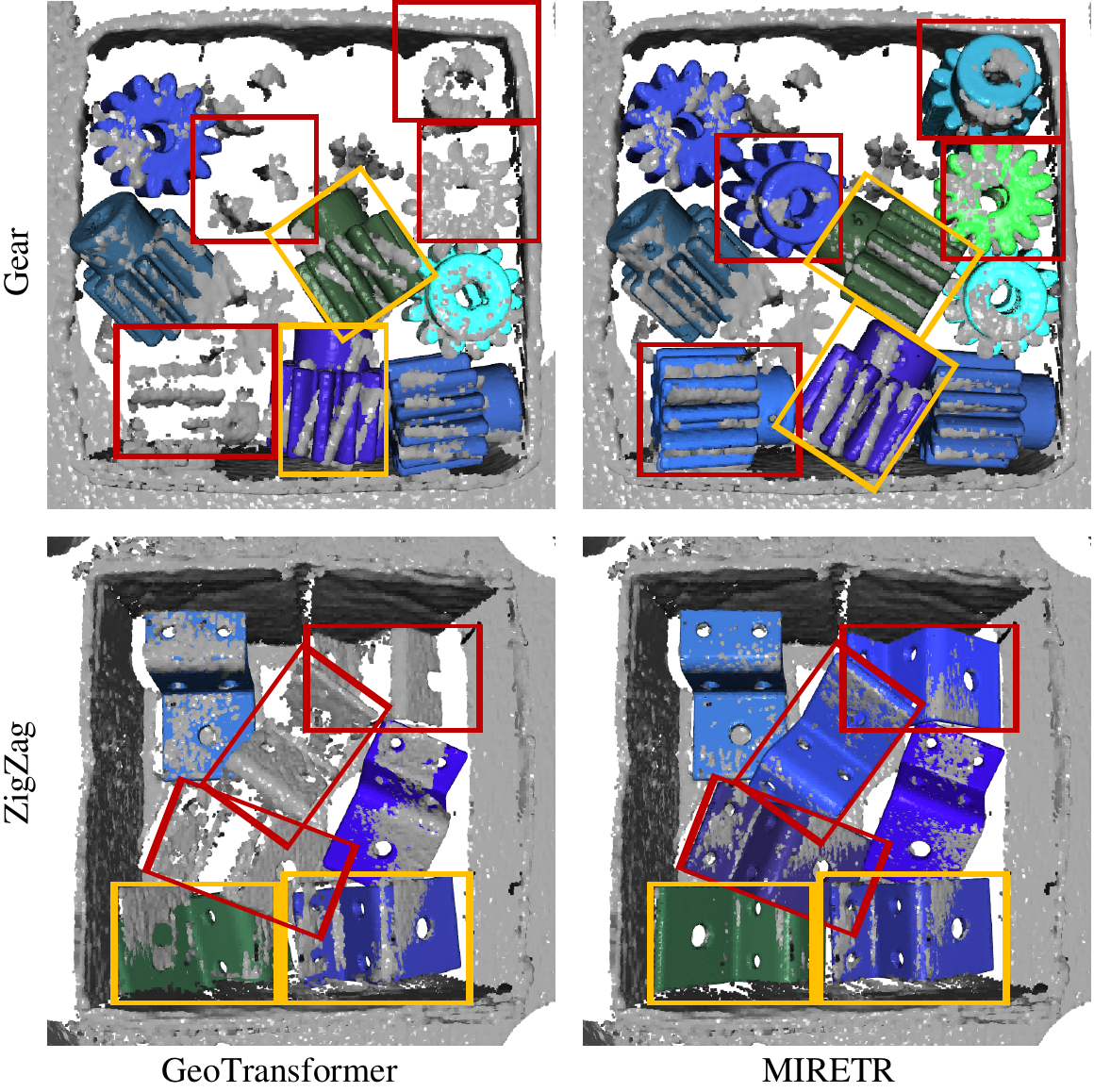}
    \bicaption{杂乱多实例场景下零件位姿识别效果\ucite{yu2024learning}}{Part pose recognition in cluttered multi-instance scenes}
    \label{fig:vision_servo}
\end{figure}

三维位姿识别方案融合深度图和点云，显式建模物体的空间结构，特别适用于无纹理、遮挡、堆叠和多实例干扰等工业场景。
Liu等人\ucite{liu2022robust}提出基于局部特征的像素级预测方法，通过编码器-解码器结构生成密集预测，适应物体几何复杂与遮挡情况。研究\ucite{liu20246dof}则利用边缘区域和姿态验证机制提升遮挡与堆叠下的识别鲁棒性。
面向多实例且背景干扰较大的工业场景，传统的几何特征提取方法易受到不相关点和实例的干扰，MIRETR\ucite{yu2024learning}通过实例掩码与超点特征提取，有效隔离目标信息，提升操作精度，如图~\ref{fig:vision_servo} 所示。Gao 等人\ucite{gao2024learning}提出可微模板匹配方法，针对掩码与灰度图像的跨模态差异，引入边缘感知模块并通过粗到精的点对应优化，实现亚像素级配准，进一步提升工业零件定位的精度与稳定性。

\begin{table*}[ht!]
  \bicaption
    {\zh{工业控制方法特性对比}}
    {\zh{Comparative characteristics of industrial control methods}}
  \label{tab:control_strategy_comparison}
  \centering
  \footnotesize
  \setlength{\tabcolsep}{0.6em}
  \resizebox{\linewidth}{!}{
  \begin{tabular}{c|c|c|c|c|c|c}
      \toprule
      \textbf{类别} & \textbf{主要依赖} & \textbf{适应能力} & \textbf{数据 / 样本需求} & \textbf{优势} & \textbf{局限} & \textbf{典型适用场景}\\
      \midrule
      传统模型驱动 &
      精确或线性化模型、规则库 &
      低—中 &
      少 &
      \makecell[l]{结构简单、实时性好；\\工程经验成熟} &
      \makecell[l]{对模型误差敏感；\\高维耦合或非线性性能下降} &
      \makecell[l]{单变量或弱耦合、环境\\可预估的过程控制} \\
      \midrule
      模仿学习 &
      专家示范数据 &
      中 &
      中—高 &
      \makecell[l]{收敛快， 初始性能优；\\可处理高维动作} &
      \makecell[l]{分布偏移；\\示范覆盖不足时泛化差} &
      \makecell[l]{具备熟练工轨迹、需快\\速部署的仿人操作} \\
      \midrule
      强化学习 &
      环境交互与奖励函数 &
      高 &
      高 &
      \makecell[l]{不依赖精确模型；\\适应非线性和多变量耦合} &
      \makecell[l]{采样效率低，早期不稳定；\\真实试错成本高} &
      \makecell[l]{可仿真或沙箱验证、环\\境动态复杂的场景} \\
      \midrule
      IL+RL混合 &
      示范数据 + 在线交互 &
      高 &
      中 &
      \makecell[l]{安全启动后持续优化；\\效率优于纯 RL} &
      \makecell[l]{权重设计复杂；\\实现成本高} &
      \makecell[l]{需继承人类经验并超越\\示范水平的高难度任务} \\
      \bottomrule
  \end{tabular}
  }
\end{table*}

\paragraph{控制策略主动适应}
低精度传感器与执行器的广泛应用使工业环境中的精确控制面临挑战。传统控制方法依赖固定参数与预设策略，难以适应动态负载与工艺变化，常导致精度和稳定性下降。相比之下，基于学习的方法可通过离线数据或在线交互优化控制策略，适应动态工况，具备处理非线性、多变量和高维任务的能力。本节将对这些控制方法进行讨论，不同方法的特性以及适用场景总结在表~\ref{tab:control_strategy_comparison} 中。

传统控制方法通常基于精确或线性化模型，通过反馈/前馈设计实现闭环控制，适用于结构简单、环境可控的场景。
比例-积分-微分（PID）控制\ucite{li2006pid}简单实用，是最常见的工业手段，适用于动态缓慢、模型不完整的系统，但在非线性、高耦合、多变量系统中性能有限。
线性二次调节器（Linear Quadratic Regulator, LQR）\ucite{anderson2007optimal}在已知状态空间下最小化控制代价，适用于线性系统但对建模误差敏感。
模糊逻辑控制（Fuzzy Logic Control）\ucite{ross2005fuzzy}不依赖精确模型，基于规则库运行，适用于经验充分、建模困难的系统。
进化与群体智能优化方法\ucite{yang2023optimal}，适用于模型不可微或目标函数复杂的场景。
然而，传统方法通常需离线调参，易陷入局部最优，且实时性不足，难以应对工业现场的快速变化需求。

基于学习的控制方法通过历史数据或与环境的交互调整策略参数，以适应生产环境中动态负载。
模仿学习（Imitation Learning, IL）\ucite{hussein2017imitation,bergamini2020deep}模仿专家数据学习控制策略，可加速训练并提升高维任务适应性。
例如，Ng等人\ucite{ng2016programming}捕捉熟练工人轨迹提取工艺策略，Zhang 等人\ucite{zhang2021learning}融合轨迹、力控与阻抗调节建模，实现高质量仿人操作。这些方法强化了机器人在非结构环境下的适应性。
然而，模仿学习存在分布偏移问题，当智能体遇到与训练数据差异较大的新状态时，可能产生不可预见的动作，导致泛化性能差。
强化学习（Reinforcement Learning, RL）\ucite{sutton2018reinforcement}则通过与环境交互优化策略，在复杂工业任务中展现出更强适应力。
Xu 等人\ucite{xu2021deep}基于 MADDPG 实现多机器人协同路径规划与避障，
Zhong 等人\ucite{Zhong2022Collision}引入逆运动学先验提升策略学习效率，Hu 等人\ucite{Hu2022Collision}结合最大熵强化学习与主次动作结构提升焊接路径规划的稳定性与采样效率。尽管强化学习方法更具灵活性，但通常训练开销大、学习过程慢，初期策略不稳定，限制了其在工业场景中的快速部署。

将已有控制器与强化学习的主动探索相结合成为一种更为高效的解决方案\ucite{rana2023bayesian}，机器人可以基于已掌握的技能进一步提升，更加精准高效地完成任务。
常见做法是先利用模仿学习初始化策略，再通过强化学习微调优化。
Deepmimic\ucite{peng2018deepmimic}借助模仿奖励引导策略逼近参考动作，并结合环境奖励实现多技能组合与复杂场景适应。
Luo等人\ucite{luo2024precise}提出动态权重机制，融合离线示范与在线交互数据，使策略既能继承人类经验，又不断优化
DDT\ucite{chen2024deep}以单次示范为动态参考，结合强化学习自适应追踪示范并应对环境扰动。
早期模仿与强化结合多依赖手工设计模仿奖励，易引发“奖励黑客”（Reward Hacking）\ucite{murphy2025would}等问题，即智能体学到的策略能够获得高分，但不是完成实际任务。
AMP\ucite{peng2021amp}借助判别器自动生成对抗式模仿奖励\ucite{ho2016generative}，并提高了鲁棒性，因而成为当前机器人操控学习的主流范式。
残差策略（Residual Policy）\ucite{silver2018residual}则在模仿策略基础上学习残差，通过强化学习修正不足，实现高效迁移与快速适应。

\paragraph{虚拟到现实迁移}
工业设备昂贵且易受损，任何故障都可能带来高额损失。因此，控制策略多在仿真环境中进行训练，以规避实际风险，常用的仿真环境包括 IsaacGym\ucite{makoviychuk2021isaac}、Pybullet\ucite{coumans2016pybullet}、MuJoCo\ucite{todorov2012mujoco}等。但仿真与现实存在差异，训练策略难以直接部署到真实场景，这一问题被称为虚拟到现实迁移（Sim2real）。

域随机化（Domain Randomization）在仿真的渲染或物理参数上引入广泛随机性，提升策略对现实环境变化的鲁棒性。
Peng等人\ucite{dynamicsDM}随机化物理参数增强对动态变化的适应，
Miki等人\ucite{miki2022learning}在多样化的模拟物理环境中训练四足机器人。在测试过程中，机器人首先通过物理接触探测地形，然后预先规划并适应步态，从而获得较高的鲁棒性和速度。
Tobin等人\ucite{visionDM}通过随机化渲染（如纹理、光照和背景）增强了仿真环境中的视觉检测能力，以提高实际场景中的表现。
Yue等人\ucite{yue2019domain}通过使用辅助数据集中的真实图像风格随机化合成图像，学习领域不变的表示。
Dai等人\ucite{DaiW0WGZ0024}提出了一种自动化流水线，将现实世界的场景转化为多样化的互动数字环境“数字表亲”，相比数字孪生\ucite{grieves2017digital}进一步提高了真实世界的迁移成功率。

尽管域随机化提供了简便的 Sim2Real 途径，过度随机化缺乏现实先验，易扩大仿真空间、增加策略学习负担\ucite{mehta2020active}，导致策略性能保守甚至下降\ucite{evans2022context}。
为此，学者们提出将真实世界映射到仿真环境（Real2Sim）进行策略学习，再将仿真策略迁移到现实世界进行 Sim2Real 测试，形成 Real2Sim2Real 闭环。有效的机器人仿真交互环境主要考虑建模几何（Geometry）、视觉外观（Appearance）、和物理动力学（Physics）\ucite{abouphysically}。
高斯泼溅\ucite{kerbl20233d, wu2024recent, xie2024physgaussian}将物体的几何和外观属性整合到高斯粒子中，能够实现外观和几何的同时建模优化，被广泛应用于Real2Sim2Real管线。
SplatSim\ucite{qureshi2024splatsim}利用高斯泼溅生成高度逼真的视觉渲染，缩小 Sim2Real 视觉差距。RobogSim\ucite{li2024robogsim}结合高斯泼溅优化场景外观和几何，使用IsaacGym 物理引擎作为底层动力学，训练仿真环境中的控制策略，并将其迁移到现实世界测试。ManiGaussian\ucite{lu2024manigaussian}在学习高斯模型后，采集真实交互数据拟合动力学，辅助策略训练，但数据驱动拟合动力学往往面临分布外泛化（Out of Distribution, OOD）挑战，需要大量的真实世界交互数据。
PIN-WM\ucite{li2025pinwm}将高斯泼溅与可微物理\ucite{3DLCP}结合，能够通过单条任务无关（Task-Agnostic）交互轨迹识别动力学模型，并在其邻域内进行物理感知扰动构建数字表亲（Physics-Aware Digital Cousins），增强策略 Sim2Real 迁移能力。

\paragraph{案例研究1：残差策略学习3C产品装配}
在计算机、通信和消费电子（3C）产品的装配任务中，零部件结构复杂、公差极小（$\pm 0.05$ mm  $- \pm 0.2$ mm），接触状态变化频繁，稍有误差即可能导致装配失败，影响产品可靠性。
提高机器人在低精度硬件上的执行精度和效率，是当前研究的一个关键问题。结合示例数据与主动探索的残差策略模型为工业机器人提供了灵活且高效的解决方案。示例数据
（通常需要 $\geq$ 50 条示范轨迹\ucite{lin2024data}）
能够使机器人在较短时间内获得高质量的训练素材，探索学习使得机器人能够主动在动态环境中进行策略优化。

\begin{figure}[h!t]
    \centering
    \includegraphics[width=1.0\linewidth]{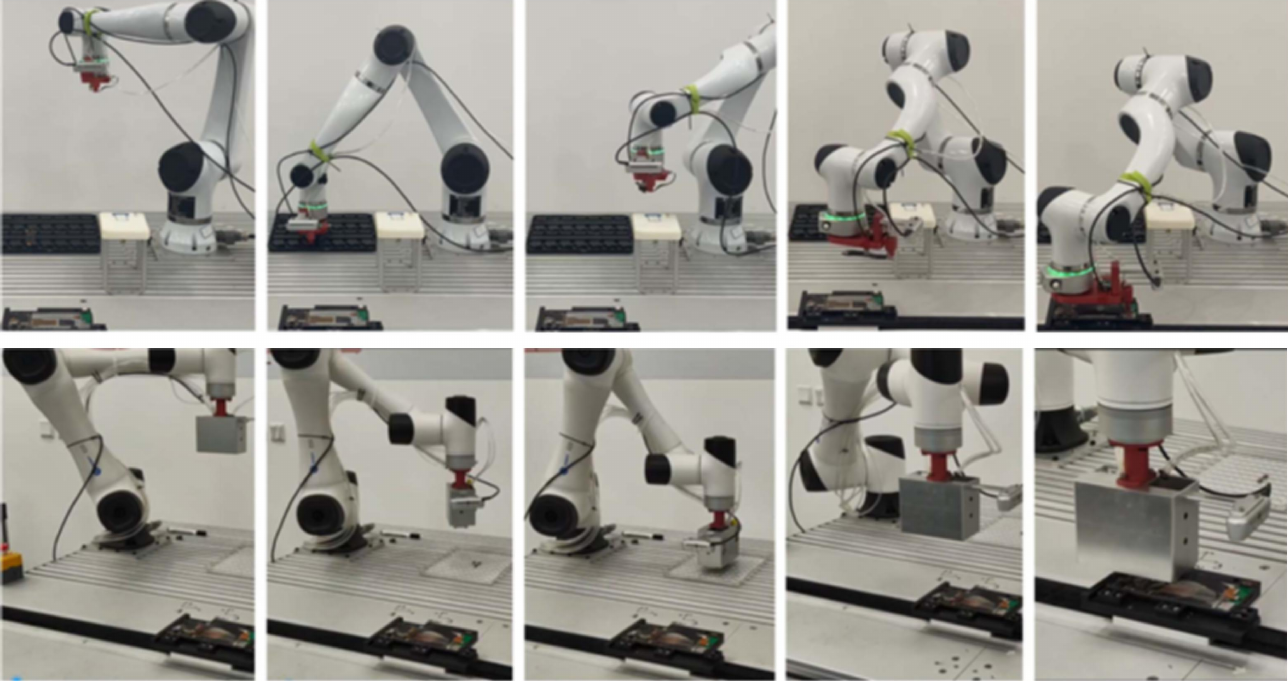}
    \bicaption{ 柔性印刷电路组装与手机摄像头装配\ucite{sun2024digital}}{Flexible circuit  and smartphone camera assembly 
    }
    \label{fig:3c_assembly}
\end{figure}

清华大学\ucite{sun2024digital}
提出结合数字孪生与残差策略的装配方案：利用虚拟现实（Virtual Reality，VR）设备采集人类多模态示范数据，包括触觉、视觉与语音信息，涵盖抓取、放置及应对变化的操作策略；在数字孪生环境中重建真实车间进行大规模仿真训练，并引入课程学习机制，逐步从理想条件适应到光照变化、组件偏移等干扰因素。该方法能够减少对大量专家数据的依赖，实现机器人在动态环境中的稳定学习与策略优化（见图\ref{fig:3c_assembly}），为高精度工业装配提供了可行路径。

\paragraph{案例研究2：低成本机器人虚拟到现实精细操作  
}
精细操作任务，如插入电池、安装钻头等，往往要求机器人具备高精度协调能力。传统工业级机器人系统虽然能够完成这种精细任务，但其高昂的成本和复杂的校准流程限制了其在实际的柔性生产应用场景扩展。在低精度硬件的背景下，如何通过智能算法提升工业机器人在现实任务中的精确控制能力，是自动化和智能制造领域的一项重要挑战。

\begin{figure}[h!t]
    \centering
    \includegraphics[width=1.0\linewidth]{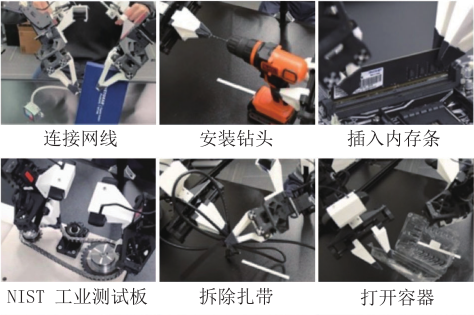}
    \bicaption{低成本硬件实现精细操作\ucite{ZhaoKLF23}}{Fine manipulation enabled by low-cost hardware
    }
    \label{fig:low_cost_robot}
\end{figure}

斯坦福大学的 ALOHA 系统\ucite{ZhaoKLF23, fu2024mobile}通过模仿学习算法 ACT，在不足 2 万美元的低成本硬件上实现了毫米级精度的双手协调操作，如图~\ref{fig:low_cost_robot} 所示。
为弥补硬件精度和动力学耦合问题，ALOHA 构建了“感知—策略—执行”的闭环系统，融合多摄像头视觉输入和 Transformer 编码器，预测多步动作序列，有效缓解误差积累。
系统采用三阶段 Sim2Real 迁移框架：首先构建与真实设备对齐的数字孪生环境；其次通过渐进式域随机化提升任务复杂度；最后结合混合现实系统与人类反馈优化训练数据。该方法在电池插入、钻头安装等精密任务中将成功率由 60\% 提升至 96\%，操作周期缩短 30–40\%，为低成本设备在柔性制造中的规模化应用提供了有效方案。

\subsection{可变化产线柔性操作}

传统工业生产依赖固定产线与预设流程，适合大规模单一产品制造，但难以应对多样化和定制化需求。现代工业要求产线具备更高灵活性，依托工业之手实现快速调整与柔性操作，适配不同产品与工艺变化。
本节将从通用控制策略学习、交互表征设计与决策基础模型三方面，探讨工业之手在可变产线下的柔性操作方案。

\paragraph{通用控制策略学习}

通用控制策略的核心目标是学习一个策略框架，使机器人在多任务场景下快速迁移。
本文主要从策略蒸馏和元学习两个角度阐述。

策略蒸馏\ucite{RusuCGDKPMKH15, czarnecki2019distilling}将多个深度网络中的策略知识迁移至一个网络，使最终的合并策略在多种环境中都能表现出色。
AutoMate\ucite{TangA0WHWFS0N24}将多个专家装配策略蒸馏到一个通用策略中，实现了在 20 种装配任务中的通用性。
Wu 等人\ucite{wu2024dexterous}提出基于师生框架的混合策略，将专家行为融合至扩散策略（Diffusion Policy）\ucite{chi2023diffusion}，支持超30类物体的动态抓取。
Mosbach等人\ucite{mosbach2024grasp}针对机器人从杂乱环境中灵巧抓取物体难题，结合强化学习与策略蒸馏两阶段学习，能够有效抓取多种物体，并展现了针对新物体的零样本迁移。

元学习\ucite{finn2017model}通过学习多个任务的经验，使模型能在新任务中有效利用已有知识进行快速学习和泛化。
元强化学习\ucite{rakelly2019efficient}则通过多任务学习使智能体在不同环境中迅速调整策略，并通过少量经验获得良好表现。
工业插入任务中的接触力学和摩擦效应难以通过传统反馈控制解决，Schoettler等人\ucite{schoettler2020meta}使用元强化学习捕获仿真任务的潜在结构。在仿真中预训练策略后，通过少量现实世界试验和误差校正，快速适应工业插入任务。
针对插入任务中的探索成本高、成功率低和适应新任务难的问题，ODA 算法\ucite{zhao2022offline}融合离线数据、上下文元学习与在线微调，仅用少量演示数据和约30分钟在线训练，便可高效完成插入任务适应，降低探索成本并提升成功率。

\paragraph{交互表征设计}

交互表征旨在提取跨任务、跨环境的通用特征，使机器人能快速迁移并高效应对复杂变化。通过建模感知与操作之间的关联，提升系统的泛化与适应能力。本文从二维与三维两个方向展开介绍。

二维表征方面，可供性（Affordance）\ucite{HuangWZL0023}用于定位图像中可交互区域，作为感知与动作之间的桥梁。
例如，在机器人抓取场景中，可供性锚定帮助模型寻找物体上最佳的抓取位置。 
VIMA\ucite{jiang2023vima}通过预训练检测器识别目标物体并提取分割区域，引导抓取策略聚焦关键区域。
Mu等人\ucite{mu2018robotic}提出结合图像简化与深度网络的方法，优化抓取位姿，显著提升在传感器模糊和结构复杂情况下的抓取成功率。
Instruct2Act\ucite{huang2023instruct2act}则结合 SAM 等\ucite{kirillov2023segment}开放物体检测模型，结合任务指令与三维坐标，生成抓取动作驱动机械臂操作。

三维交互表征方法方面，现有工作通过设计足够强泛化能力的三维交互表征，提升机器人在复杂环境中的操作能力。
She等人\ucite{SheHXLXH22}提出基于交互平分曲面（Interaction Bisector Surface, IBS）的状态建模方法，细粒度表达手爪与物体关系，实现对复杂形状的灵巧抓取。
Xiao等人\ucite{xiao2025designing}设计末端与物体接触区域的动态表征，驱动按钉式夹爪实现自适应抓取与手内重定位。
研究\ucite{she2024learning}提出跨抓手策略迁移框架，利用通用策略预测关键点位移，再由适配模块转换为具体控制信号，适配不同手爪结构。
DP3\ucite{ZeZZHWX24}关注利用稀疏点云的紧凑三维表示，通过高效的点编码器提取三维视觉特征，展现出良好的泛化能力。

\paragraph*{决策基础模型}
决策基础模型
决策基础模型 通过在多任务、多环境中训练，学习共享特征与通用策略，具备跨任务泛化能力，突破了传统方法对单一场景的依赖。
决策基础模型大致形成两条范式，分别侧重 “先规划后执行” 与 “端到端决策”。
本文将从这两条路线的代表性工作进行阐述。

“先规划后执行” 范式以大型语言模型（Large Language Model, LLM）充当高层规划器，将自然语言任务指令解析为低阶技能序列，交由底层控制器执行。
该方法依托 LLM 的世界知识与逻辑推理能力进行任务分解\ucite{hu2024human, vijayaraghavan2025development}，具备可解释性强、模块化好、便于融合符号规则等优点，但对技能库覆盖和感知—动作接口的一致性要求较高。
Palm-E\ucite{PaLM}将图像与状态等感知输入编码为与文本统一的潜变量，通过 LLM 自注意力机制联合处理，输出语言形式的任务计划。
SayCan\ucite{ahn2022can}则结合 LLM 生成的技能候选与可行性评分网络，筛选出最可执行、最有效的动作序列。
Ha等人\ucite{ha2023scaling}提出 LLM 引导下的扩散策略框架，通过高层任务规划与失败检测生成多样轨迹，提升多任务策略泛化能力。
思维链 (Chain of thought, CoT) 模拟人类的思考过程，将复杂问题分解为一系列更小、更易于处理的步骤\ucite{wei2022chain}。
Embodied-GPT\ucite{mu2023embodiedgpt}通过 CoT 生成更详细和可执行的计划， 从而提高机器人执行任务的成功率。

在“先规划后执行”范式中，大语言模型负责生成任务规划与技能序列，而底层技能库通常预定义。此时，技能衔接（Skill Chaining） 成为关键问题，即需确保前一技能的终止状态与后一技能的初始状态紧密匹配，以实现无缝切换\ucite{chen2024scar}。
T-STAR\ucite{lee2021adversarial}通过奖励正则化优化各子任务策略，使其终止与起始状态尽可能重合。
Huang等人\ucite{huang2023value}则训练高层调度器，为每阶段选择最可能完成衔接的目标条件策略（ Goal-conditioned Policy），调度器同样通过强化学习获得。然而，多数方法依赖固定顺序的子策略重训练，难以应对生产工序的灵活调整。
DeCo\ucite{chen2025deco}提出了一种更具实践意义的方案：为每个技能定义一个起始关键帧，每当完成当前工序后，用运动规划（Motion Planning）\ucite{riviere2024monte}算法自动转移至下一个技能的关键帧，以实现技能自由组合。

“端到端决策”通过联合编码多模态观测与语言指令，直接输出低层控制信号，实现感知—理解—控制的一体化闭环\ucite{liu2025fusion, wang2025goal}。该方法在大规模交互轨迹上训练，可零样本迁移到新物体与场景，但长时序推理与安全验证仍待突破。
LaMo\ucite{shi2024unleashing}首次探索将小规模 GPT-2 用作离线强化学习策略，以条件模仿学习框架训练。
Google 的 Robot Transformer 系列\ucite{BrohanBCCDFGHHH23, zitkovich2023rt}借助更大模型和数据集，在多种具身任务中取得优异表现。
OpenVLA\ucite{kim2024openvla}基于 LLaMA 2\ucite{touvron2023llama}、DINOv2\ucite{oquab2024dinov}与 SigLIP\ucite{zhai2023sigmoid}，在 97 万条真实演示数据上训练，支持消费级 GPU 微调，具备多任务与语言指令泛化能力。
ManipLLM\ucite{li2024manipllm}通过微调多模态大模型，实现对末端执行器位姿的直接预测。
$\pi_{0.5}$\ucite{intelligence2025pi05visionlanguageactionmodelopenworld}构建在 $\pi_{0}$\ucite{black2024pi0visionlanguageactionflowmodel}基础之上，整合异构任务数据，支持开放世界下的长时序执行。
3D-VLA\ucite{3D-VLA}融合三维场景理解与动作规划，实现了对真实物理世界的多步操作生成能力。

尽管端到端方法可借助大规模数据学习通用策略，但在复杂多样的工业任务中训练出“一劳永逸”的模型仍具挑战。更可行的路径是：以预训练策略为基础，结合少量任务数据进行高效微调，从而提升适应性并降低数据需求。
RLDG\ucite{xu2024rldg}提出以强化学习生成高质量示范，再用于精细操控任务的微调流程，显著提升模型成功率约 30-50\%。
ConRFT\ucite{chen2025conrft}针对 VLA 模型在示范稀缺、分布偏差和适配困难等问题，设计了统一的离线-在线强化微调框架。离线阶段通过少量示范与一致性训练初始化模型（如 OCTO\ucite{team2024octo}），在线阶段结合人类干预与交互数据实现快速安全适应。
在 8 项高接触任务中，ConRFT 仅用 45–90 分钟微调便将平均成功率提升至 96.3\%。这些方法为工业具身智能中通用策略的快速适配与部署提供了可行路径。

\paragraph{案例研究1：多零部件柔性装配}

\begin{figure}[h]
    \centering
    \includegraphics[width=\linewidth, trim=0 40 0 180, clip]{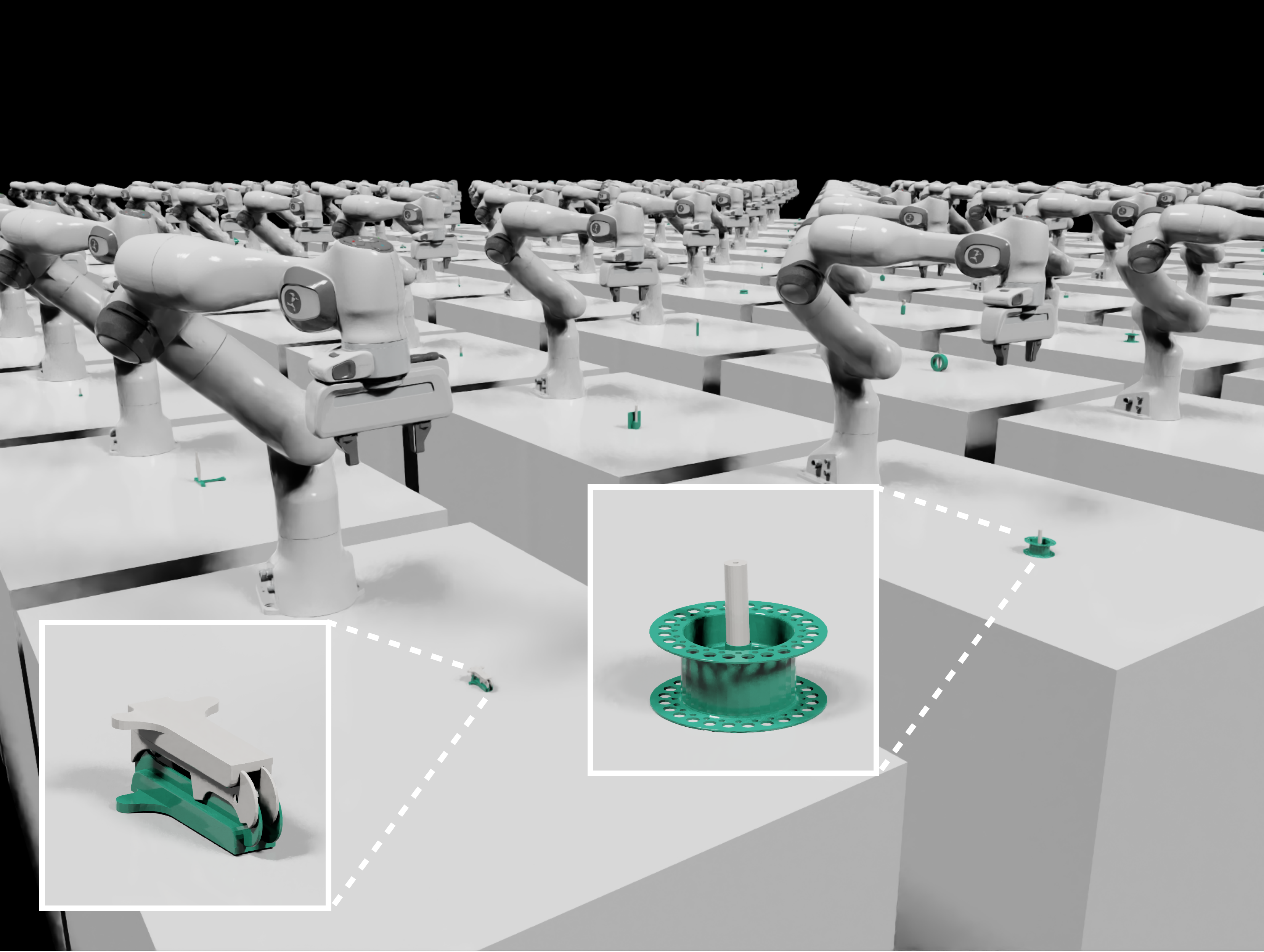}
    \bicaption{柔性装配场景中零部件多形态示例\ucite{TangA0WHWFS0N24}}{Examples of diverse part geometries and poses in flexible assembly scenarios  
    }
    \label{multiple_assembly}
\end{figure}

面对柔性制造多类型、多配置和个性化定制的需求，传统固定生产线通常依赖于高度定制的工程设计和固定的运动路径，难以灵活应对多样化零件的装配需求。装配中零部件结合方式的变化直接影响接合强度，机器人需要具备高度的自适应调整能力以适应各种几何形状和姿态的零部件，如图~\ref{multiple_assembly} 所示，这对传统机器人控制技术提出了挑战。

NVIDIA 提出通过策略蒸馏学习通才装配策略\ucite{TangA0WHWFS0N24}，以适应不同几何形状与姿态的零部件装配。
方法首先采用行为克隆（Behavior Cloning, BC）\ucite{rajeswaran2017learning}初始化策略，随后结合 DAgger\ucite{ross2011reduction}和课程式强化学习进行微调，并将多个专家策略蒸馏至单一通用策略网络。该策略在 80 种装配任务中平均成功率超 80\%，对未见过零件在 $\pm$0.5$-$1mm 公差下装配成功率达 88\%。在实机验证中，通用策略在 20 类不同零件上实现 86.5\% 成功率，展现出毫米级装配精度。

\paragraph{案例研究2：大模型指导的通用打磨控制}
打磨是风电、航空、造船等行业中影响产品性能的关键工序，但面临缺陷类型多样、工件几何复杂等挑战。
航空发动机叶片翼型表面打磨后的形状偏差要求控制在 $\pm 0.03$mm 以内；风电叶片打磨后表面粗糙度需达到 Ra $\leq 0.8 \mu m$，形状误差不超过 $\pm 0.2$mm；而船用钢板打磨则对平整度的允许偏差一般在 $\pm $0.1 mm $- \pm  0.3$ mm 之间。
传统基于固定流程的方法难以适应多变需求，亟需通用且自适应的智能控制技术。

\begin{figure}[h]
    \centering
    \includegraphics[width=\linewidth]{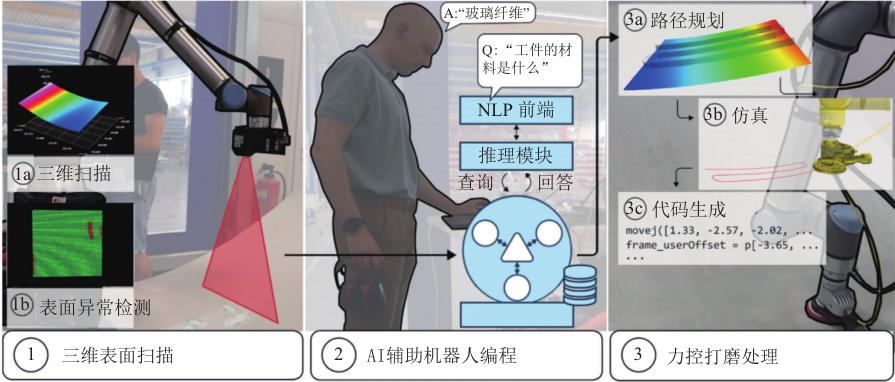}
    \bicaption{RoboGrind：智能通用打磨系统\ucite{alt2024robogrind}}{RoboGrind: a generalized robotic  grinding system }
    \label{fig:RoboGrind}
\end{figure}

RoboGrind\ucite{alt2024robogrind}基于 LLM 构建了集成三维感知、本体推理、自然语言交互与力控执行于一体的通用打磨控制系统，如图~\ref{fig:RoboGrind} 所示。
系统利用工件点云与自然语言输入生成结构化任务指令，自适应补全信息并完成上下文推理。底层通过基于模型的强化学习融合点云反馈，动态优化轨迹与力控参数。即便在任务临时变更或语言描述不精的情况下，系统仍能稳定完成任务规划与控制生成。该工作展示了 LLM 在任务理解、策略推理与跨任务迁移方面的潜力，具备向其它工业具身任务扩展的广泛价值。

\subsection{工艺参数自适应调节}
相较于强调轨迹精度的机器人操作任务，工艺参数调节聚焦于生产过程中对影响性能与质量的关键参数进行动态控制，以确保过程稳定性与产品一致性。
焊接、打磨和装配是工业制造的代表性工艺，均涉及复杂物理过程和多种工艺参数的调节。
焊接环节需综合调节电弧参数、送丝速度、焊枪速度和角度，以实现熔池形态和热输入的最优匹配。打磨过程需动态控制接触力、切削速度、磨头位姿与路径规划，以确保表面光洁度和形状精度。装配工艺则需精确管理位置与姿态、装配力与扭矩、运动速度及环境条件，以保证零件准确对位和稳定连接。

传统工艺参数调节主要面向单一产品，依赖人工经验和反复试验，在高度结构化产线上获取稳定参数。如 Wang 等人\ucite{wang2024effect}探讨了转速、磨头数量和磨削方向等参数对磨削碳纤维增强复合材料的影响，并建议了能够降低表面粗糙度的参数组合。
随着制造向多品种、小批量、高定制化转型，预设参数难以应对频繁换线与快速调整。
柔性制造亟需数据驱动的自适应调参技术，通过实时感知系统状态与环境变化，动态优化控制参数，实现稳定高效的工艺执行。
数据驱动控制采用状态空间建模，适用于多输入输出系统，强调从历史数据中学习参数-性能映射，具备强泛化性与在线优化能力。下文将以焊接、打磨和装配为例，探讨相关研究进展。

在焊接工艺中，电流、电压、送丝速度和轨迹等参数直接影响焊缝的结构强度\ucite{huang2020real}和成型精度\ucite{ma2021effect}。
Kershaw等人\ucite{kershaw2021hybrid}提出结合 CNN 和 MLP 的自适应控制方法，根据熔池图像预测焊缝宽度并调整速度。
Wang等人\ucite{wang2022data}进一步提出基于先进梯度下降的自适应控制方法，通过标准化历史梯度与当前梯度的均值和方差，提高控制稳定性，降低误差带和初始焊池分裂率。
Jin等人\ucite{jin2019intelligent}提出基于强化学习的熔池宽度控制策略，验证了其在 GTAW 与 GMAW 焊接中的有效性。
Masinelli等人\ucite{Masinelli2020Adaptive}将强化学习应用于激光焊接中激光功率的闭环自适应控制，通过智能体与反馈系统实现无需先验知识的焊接质量优化。

打磨工艺中，切入量与进给速度关系到表面粗糙度与稳定性。切入量过大将增加磨削力和热输入，易引发工件烧伤、磨粒堵塞，恶化表面粗糙度和几何精度。过高进给速度会引发振动，对加工质量造成影响\ucite{guo2024vision}。
Cheng 等人构建数据驱动的工艺参数匹配模型，实现流程制造中的参数优化\ucite{程进2017数据驱动的流程制造工艺参数匹配方法}。
Liu 等人\ucite{liu2022digital}提出的IPSO-GRNN模型通过融合实时加工数据与加工机理预测表面粗糙度，并动态优化切削参数。
Li等人\ucite{li2022contact,li2024process}提出融合多参数的去除模型与力-位混合控制方法，有效提高打磨精度和质量。
Zhang 等人\ucite{zhang2021learning}通过建模人工打磨过程中的力控参数与轨迹规划，实现复杂动态下的精准响应。
整体上，融合物理建模与人类经验的智能控制能够增强系统的灵活性与鲁棒性。

装配工艺需精准控制末端执行器的位置、姿态、力/扭矩与速度，以确保零件准确对位和稳定连接，避免损坏或松动\ucite{wu2020torque}。
Zhang等人\ucite{zhang2022semi}提出用于航空发动机转子装配的强化学习紧固方法，通过建模螺栓间弹性作用并结合 GRU 网络预测同轴度变化，提升装配精度。
Zhou等人\ucite{zhou2023research}提出了一种结合视觉和力信息的螺栓紧固方法。通过椭圆弧拟合和三点法估计螺纹孔位置，利用被动柔顺性监测控制径向力，并设计了自适应控制器以减少力冲击和提高跟踪精度。
Shtabel等人\ucite{shtabel2024automated}针对小型航天器装配设计了基于视觉与无线工具的自动控制系统，简化硬件配置并验证其实用性。
You 等人\ucite{you2024disturbance}提出融合扰动观测器与有限时间控制的视觉伺服算法，在稳态精度上优于传统控制器如 PID、LQR 和 MPC。

\paragraph{案例研究1：弧焊工艺参数实时控制}
焊接通过加热、加压或两者结合使接触面材料熔化或塑性变形，冷却后形成连接。在柔性制造中，焊接的智能化水平直接影响产线适应性与效率。
常规焊缝跟踪精度要求为 $\pm$0.2$-\pm$0.5 mm，高精度场景（如航天、精密管道）则需控制在$\pm$0.1–$\pm$0.2 mm。若焊缝宽度或深度误差超出 $\pm$0.5 mm，易导致未熔合、应力集中等缺陷，削弱结构强度并增加疲劳风险。然而焊接过程受材料、装配误差、热变形和环境扰动等多因素影响，传统固定参数控制难以适应复杂动态工况，易造成质量波动与缺陷。

\begin{figure}[h!]
    \centering
    \includegraphics[width=1.0\linewidth]{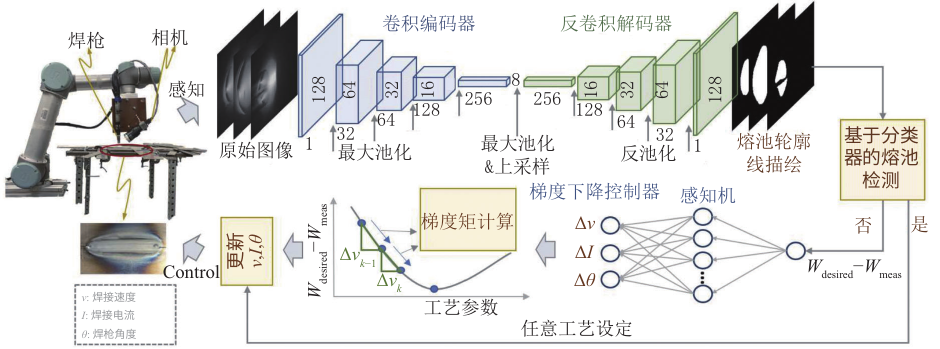}
    \bicaption{基于熔池观测的工艺参数实时控制流程\ucite{wang2022data}}{Real-time parameter control based on molten pool observations}
    \label{fig:Kentucky_weld_pipeline}
\end{figure}

肯塔基大学研究团队提出了一种基于数据驱动的弧焊实时控制方法\ucite{wang2022data}，算法框架如图~\ref{fig:Kentucky_weld_pipeline} 所示。
他们首先通过光学成像捕捉焊池宽度，并利用像素级图像分割网络精确测量。然后，基于焊池宽度变化，采用梯度下降驱动的控制器在线优化焊接参数，实现快速且连续的反馈调节。该方法提高了调整效率，减少了稳态误差，能够在仅七个控制周期内将焊池宽度收敛至目标范围。

\paragraph{案例研究2：基于元强化学习的打磨工艺参数自适应调节}

\begin{figure}[h]
    \centering
    \includegraphics[width=1.0\linewidth]{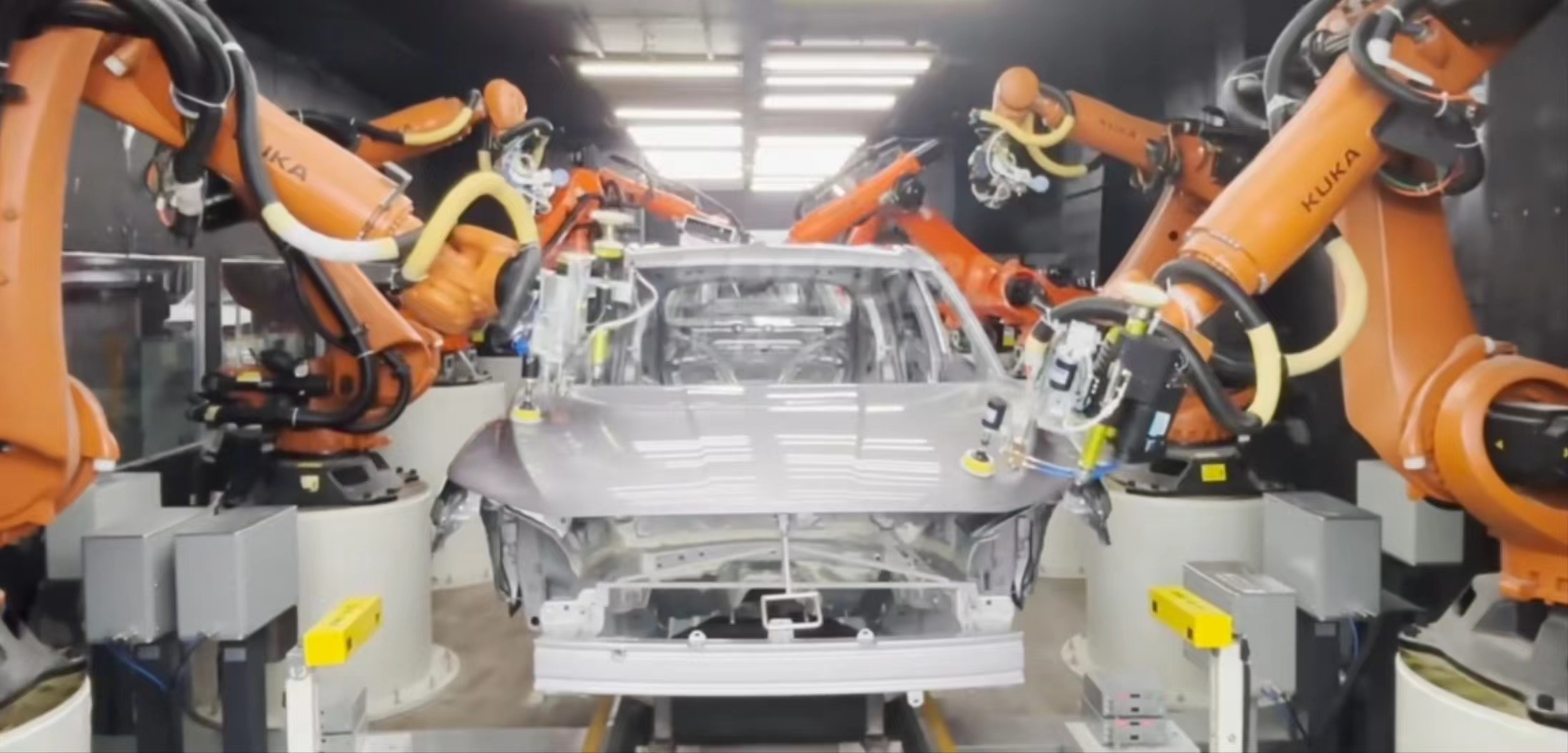}
    \bicaption{实时调整打磨工艺参数以提升材料去除精度}{Real-time adjustment of grinding parameters to improve material removal accuracy}
    \label{robot_grinding}
\end{figure}

打磨通过磨料去除表面不规则，提升平整度与光滑度。在航空、汽车等领域，关键零件常要求表面粗糙度 Ra $\leq$ 0.8 $\mu$m、平整度误差$\leq$ 0.05 mm。传统人工或刚性自动化打磨在面对曲率剧变、不规则接触和工件差异时，适应性差、效率低，难以稳定达标。
在线打磨控制可根据工件状态与磨具磨损动态调整参数，提升去除精度与加工效率，如图~\ref{robot_grinding} 所示。

华中科技大学提出一种基于元强化学习的打磨参数自适应调节方法\ucite{pan2025adaptive}。
该方法维护“优质经验池”以优先利用高回报轨迹，提升材料去除精度，并结合模型无关元学习（Model-Agnostic Meta-Learning，MAML）\ucite{finn2017model}与 PPO （Proximal Policy Optimization）算法，通过多轮梯度更新获取通用初始策略。面对新任务时，仅需少量样本即可快速适应不同工件特性和磨具条件。实验表明，MAML-PPOBE 在去除误差和收敛速度上优于 MAML、PPOBE、SAC 和模糊控制，展现出更强的鲁棒性与一致性，为实现高精度、实时自适应打磨控制提供了有效路径。
\section{工业之脑（Industrial Brain）}

在工业具身智能体系中，工业之脑负责多工序、多工位、多任务的全局调度与决策。不同于依赖经验与静态规则的传统排产方式，工业之脑以数据驱动与模型推理为核心，构建具备动态适应与实时优化能力的智能中枢。其核心能力包括：多工位任务的智能排产与资源调度，应对订单变动与产线重构；生产全流程的数字建模与虚实同步，实现制造状态的精准感知与快速响应；以及对复杂工艺的物理建模与推理，支撑更高精度与自主性的生产控制。下文将工厂排产智能调度、数字孪生虚实同步、以及世界模型的物理感知等核心技术展开探讨。

\subsection{工厂智能排产调度
}

柔性生产要求系统能够随订单变化快速调整计划、动态协调资源，应对复杂工艺流程。在此背景下，工厂级智能调度与决策成为实现柔性响应的关键。一方面，工序并行、产线重构与插单等动态约束使全局最优调度方案对产能与交付效率至关重要；另一方面，设备、人员与物料的高效协同也对数据驱动、模型感知与智能优化提出更高要求。
为应对多任务、多资源、高动态的调度挑战，本节将从智慧工厂中的任务排产、路径规划与物料仓储等角度介绍相关工作。

\paragraph{车间作业调度}

 车间作业调度（Job Shop Scheduling Problem, JSSP）\ucite{manne1960job}旨在多工序、多设备约束下，为各作业分配合理顺序与起止时间，以最小化生产周期、资源占用或延迟成本。
作为生产计划核心，JSSP不仅关系到产能利用与交付效率，也决定系统对插单与故障的响应能力。
因其约束复杂且为 NP-hard 问题，长期是组合优化研究重点。传统方法如分支定界\ucite{brucker1994branch}适用于小规模精确解，遗传算法\ucite{zhang2021surrogate, sun2024improved}、模拟退火\ucite{fontes2023hybrid}、禁忌搜索\ucite{xie2023hybrid}等元启发式方法则可在中等规模问题中获得高质量近似解。然而，这些方法多依赖离线优化，缺乏对动态信息的实时感知与响应能力，难以应对柔性制造中频繁变化的任务与资源调度需求，限制了其在实际复杂场景中的适用性。

 随着传统调度方法在响应速度与泛化能力方面日益受限，基于学习的调度策略逐渐成为研究热点。
 Zhang等人\ucite{zhang2020learning} 利用图同构网络（Graph Isomorphism Network, GIN）\ucite{XuHLJ19}编码调度状态，结合策略梯度从数据中直接学习策略。
研究\ucite{zhang2022dynamic}针对插单与设备故障引入多智能体强化学习框架，视设备为边缘智能体，通过改进 PPO 与合同网协议实现协同调度。
Destouet等人\ucite{destouet2023flexible}提出面向可持续柔性车间的多目标强化学习调度方法，兼顾效率、能耗与延迟。
Li 等人\ucite{li2024synchronized}针对同步双臂重排，结合注意力机制与成本预测提升高维规划效率。
Zhang 等人\ucite{zhang2024learning}提出双臂机器人在线分层调度算法，高层使用 RL 进行任务分配，低层通过启发式规划运动，避免物体增多时的搜索爆炸。
Yao等人\ucite{yao2024knowledge}融合关键路径邻域搜索与知识引导的混合优化，提升求解效率与质量。
SeEvo\ucite{huang2024automatic}通过将大语言模型 LLM 融入自动算法设计，
生成启发式提示与个体程序，然后通过个体与集体的自我进化反射机制，实现了车间调度策略的动态生成与优化。

\paragraph{自主移动单元路径规划}

在柔性生产中，自主移动单元（如 AGV）的路径规划对提升物流效率与产线吞吐量至关重要。
面对共享通道、交叉区域等复杂环境，系统需动态规划路径以避免碰撞、拥堵与死锁，并确保关键物料准时送达。
该问题可归结为旅行商问题（Traveling Salesperson Problem, TSP），即在所有预定取放点之间设计一条最短闭环路径\ucite{gavish1978travelling}。其多车辆扩展形式车辆路径问题（Vehicle Routing Problem, VRP）\ucite{toth2002vehicle}将多辆车视为多名“旅行商”，求解从仓库出发访问各客户的最优路线集合，与工业场景中多运输单元的路径规划高度契合。传统上，TSP 和 VRP 同样依赖精确算法\ucite{concorde, gurobi2018}或元启发式方法\ucite{cook2011traveling, LKH, 2_opt}求解。但在大规模、动态、多目标场景下，常面临解稳定性差与实时性不足等挑战\ucite{mahmudy2024challenges}。

为突破上述瓶颈，研究者开始引入强化学习、监督学习和图神经网络等技术，探索智能化、高效且可扩展的路径规划新范式。
Xue等人\ucite{xue2018reinforcement}提出通过全车间信息共享和强化学习，使多 AGV 在流水车间中实时协同决策，降低整体完工时间。AM\ucite{KoolHW19}结合 Transformer 注意力机制与 REINFORCE 算法\ucite{sutton2018reinforcement}，实现了对 VRP 的端到端学习求解。ELG\ucite{gao2023towards}通过融合可转移的局部策略与全局策略，提升了模型在不同实例间的泛化能力。DIFUSCO\ucite{DIFUSCO}将 TSP 建模为离散 0-1 向量优化问题，利用图去噪扩散模型生成高质量路径。DISCO\ucite{yu2024disco}进一步通过残差引导与解析式加速，构建高效的扩散求解器，并结合分治策略，实现了对超大规模 NP-hard 问题的高效推理。

\paragraph{物料仓储优化}

\begin{figure}[h!]
    \centering
    \includegraphics[width=1.0\linewidth]{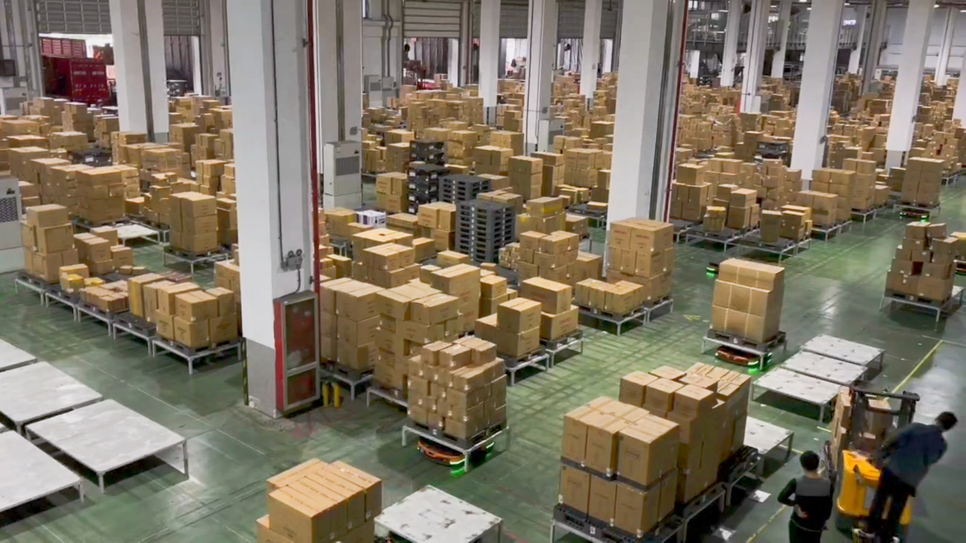  }
    \bicaption{工业生产中装箱需求普遍存在}{Packing are ubiquitous in industrial scenarios}
    \label{intelligent_transport}
\end{figure}

在有限空间内高效存储多品种物料，提升系统灵活性、降低库存成本，同样是提升生产系统灵活性与降低库存成本的关键问题。该问题可形式化为装箱问题（Bin Packing Problem, BPP）\ucite{MartelloPV00}，即在有限容器中最优放置物品以最大化空间利用率，如图~\ref{intelligent_transport} 所示。
但工业环境下装箱任务面临物品顺序未知的在线决策挑战\ucite{Seiden02}、需实时响应与确保操作安全等动态约束，显著复杂于理想的离线场景。
传统的 BPP 解法\ucite{MartelloPV00, faroe2003guided,de2003greedy}多依赖完整序列信息，难以适应上述动态需求。

研究者将工业中的在线装箱问题建模为受物理约束的马尔可夫决策过程，并采用深度强化学习优化策略。
CDRL\ucite{zhao2021online}引入可行性掩码约束无效动作，结合 Actor-Critic 框架提升装箱效率，表现优于人类专家，已实现工业部署\ucite{zhao2022learning}。
PCT\ucite{zhao2021learning}则将装箱策略学习转化为树结构的
层级动作扩展问题，仅关注树结构中有限的叶
子节点作为装箱动作，限制决策的解空间大小，首次实现连续解域下的三维在线装箱，并兼容多种工业约束。
Zhao 等人\ucite{zhao2025deliberate}进一步扩展 PCT 至考虑动态运输稳定性的场景（如 AGV 运输）。
针对不规则物体装箱，研究\ucite{zhao2023learning}通过候选动作生成与异步强化学习缩小动作空间并加速训练。上述方法在实时性、鲁棒性和多约束兼容性方面为工业在线装箱提供了参考。

除了基本的装箱位置优化，实际工业应用还关注装箱顺序与位置的协同优化。TAP-Net\ucite{hu2020tap, xu2023neural}结合几何特征和优先级图，通过 RNN 解码器在实时高度图上迭代决策，实现了装箱顺序与放置位姿的共同优化。
Zhao 等人\ucite{zhao2025deliberate}将 PCT 结合至规划框架中，将装箱问题的顺序优化和位置优化建模为一个统一的搜索框架，
仅依赖一个预训练的 PCT 模型，即可实现对于多种装箱问题形式即插即用的问题求解，如前瞻装箱\ucite{grove1995online}、缓冲装箱\ucite{puche2022online}、离线装箱等\ucite{MartelloPV00}。

\paragraph*{案例研究1：制造系统多智能体动态车间调度}
在智能制造转型中，车间控制正由集中式向分布式自主协同演进，调度系统亟需具备更强的动态响应与适应能力，以应对订单波动、资源异常与设备故障等扰动。多智能体调度架构通过将关键资源建模为具备感知与决策能力的自治体，实现任务重构、资源重分配与局部自主—全局协同统一，为柔性制造提供更鲁棒、可扩展的调度方案。
\begin{figure}[h!t]
    \centering
    \includegraphics[width=1.0\linewidth, trim=0 0 0 0, clip]{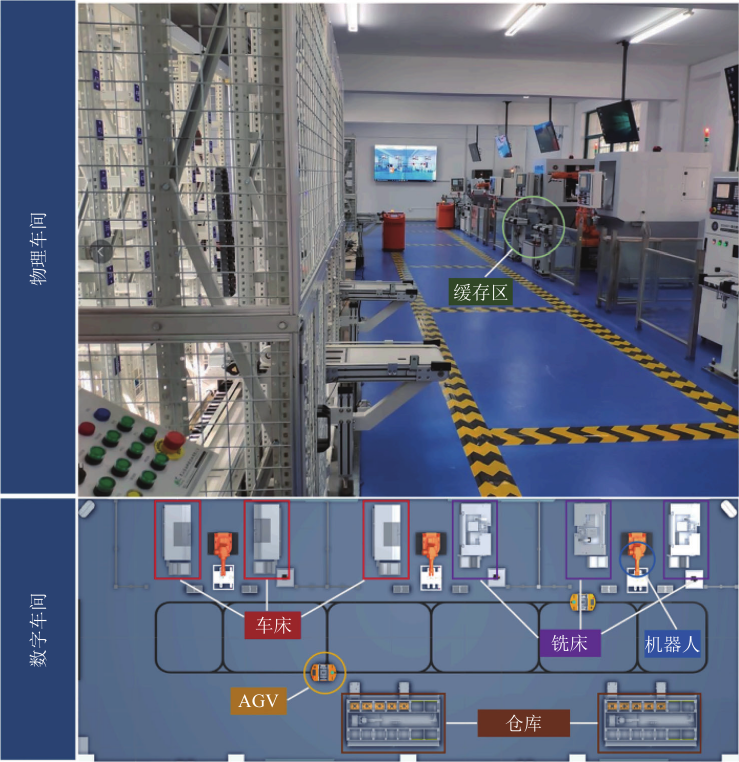  }
    \bicaption{智能车间总体布局\ucite{zhang2022dynamic}}{The overall layout of a smart workshop
    }
    \label{fig:dynamic_smart_workshop}
\end{figure}

南京航空航天大学的研究团队\ucite{zhang2022dynamic}设计了一种基于深度强化学习的分布式多智能体车间调度系统。
将车间每个设备建模为具备边缘计算能力的“智能体”，并嵌入多层感知网络构成的“AI调度器”，依据车间感知状态生成生产决策，同时采用PPO算法进行周期性训练优化。该系统在紧急订单插入与设备故障等动态场景下展现出动态适应性、多目标协同性及抗干扰鲁棒性等优势。
多智能体系统通过将每个制造资源（如机床、物流单元、传感器）建模为自治智能体（图~\ref{fig:dynamic_smart_workshop}），使其具备本地感知、自主决策与相互协作的能力。
这为实现“云边协同”与“局部最优—全局协调”的调度架构奠定基础，为工业场景提供高鲁棒性的解决方案。

\paragraph*{案例研究2：多尺寸物料在线混合码垛}

工业制造生产链中往往需要对多种类物料混合码垛。由于来料在尺寸、形状等几何特征上具有多样性，工业码垛往往涉及复杂的几何空间组合决策，这部分工作仍主要依赖人工完成。搭建工业在线码垛的具身智能系统，机器人自主决策最优装
箱位置，最大化空间利用效率并实现不间断自动化生产，具有重要经济价值。

\begin{figure}[h]
    \begin{minipage}{0.46\textwidth}
    \centering
    \subfigure[工业码垛机器人]{
        \includegraphics[width=\linewidth,trim=60 20 0 80,clip]{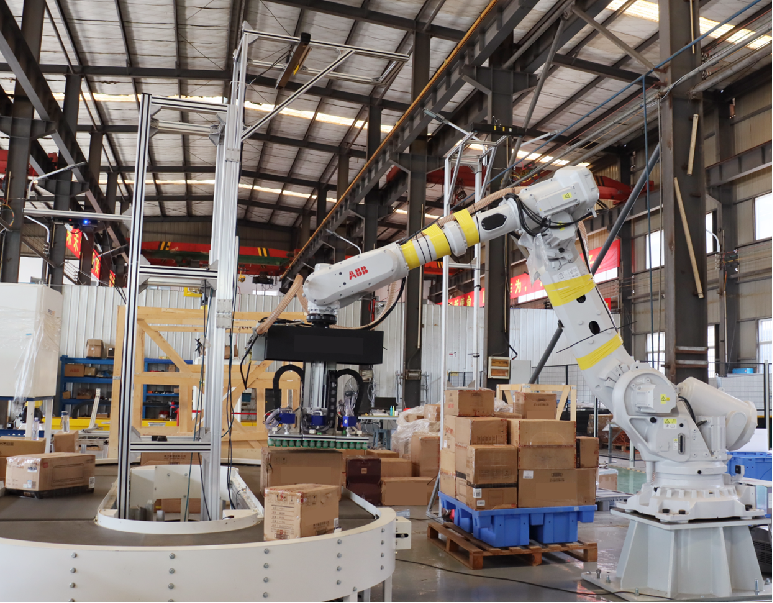  }}
    \end{minipage}
    \begin{minipage}{0.46\textwidth}
    \centering
    \subfigure[在线装箱结果]{
        \includegraphics[width=\linewidth]{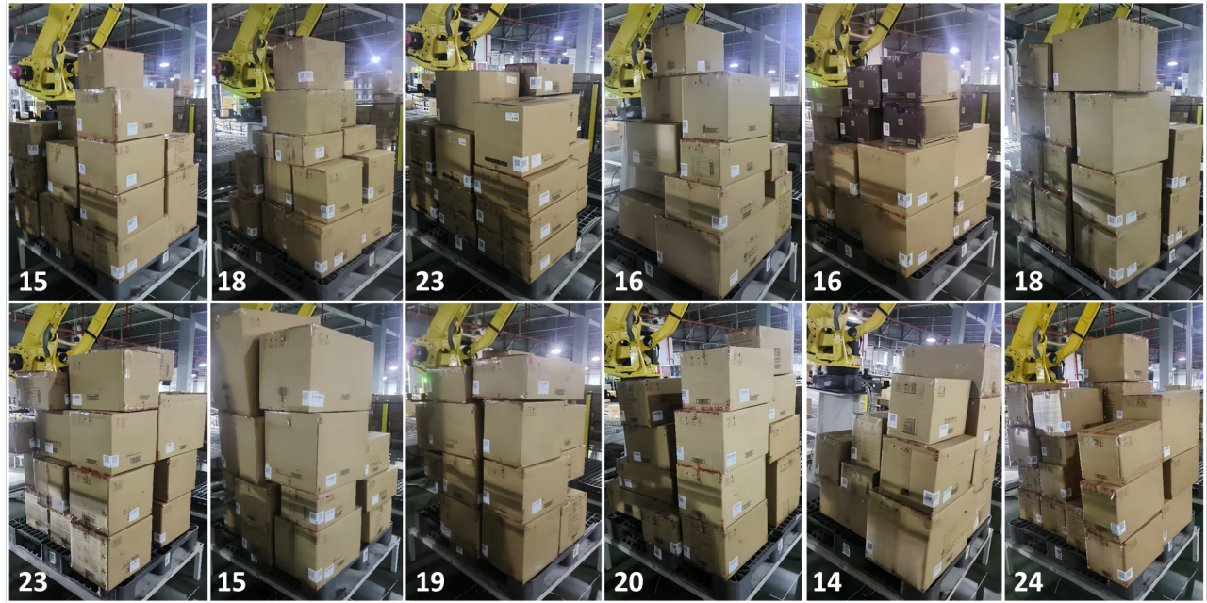  }}
    \end{minipage}
    \bicaption{在线混合码垛系统示意\ucite{zhao2025deliberate}}{Schematic of an online hybrid packing system}
    \label{industry_packing}
\end{figure}

Zhao 等人\ucite{zhao2025deliberate}在工业仓库搭建了可实际部署的在线码垛系统，专为满足受限放置\ucite{choset2005principles}和运输稳定性\ucite{hof2005condition}等工业需求而设计。不同于依赖箱壁保护的实验室原型 \ucite{yang2021packerbot,xu2023neural}，该系统直接在无保护托盘上码垛，更贴近实际工况。为降低轻微碰撞导致的堆叠失稳风险，系统采用模块化末端执行器，根据箱型自适应调整自身形态，
以在最大化抓取力的同时降低碰撞风险。为应对运输过程中的不确定性，在测试时每次放置均在多组干扰条件下仿真评估，且仿真借助 GPU 批量并行加速保障生产节拍\ucite{makoviychuk2021isaac}。
该系统在工业标准的无保护托盘上实现了高效、可靠的码垛操作，每个箱子仅需约10秒。对于相对较大的箱子，平均每个托盘可装19个箱子，空间利用率达到57.4\%，且所码放跺型可直接由 AGV 等自主移动单元运输，
码垛结果如图~\ref{industry_packing} 所示。

\subsection{数字孪生虚实同步}

\begin{figure}[h!t]
    \centering
    \includegraphics[width=1.0\linewidth]{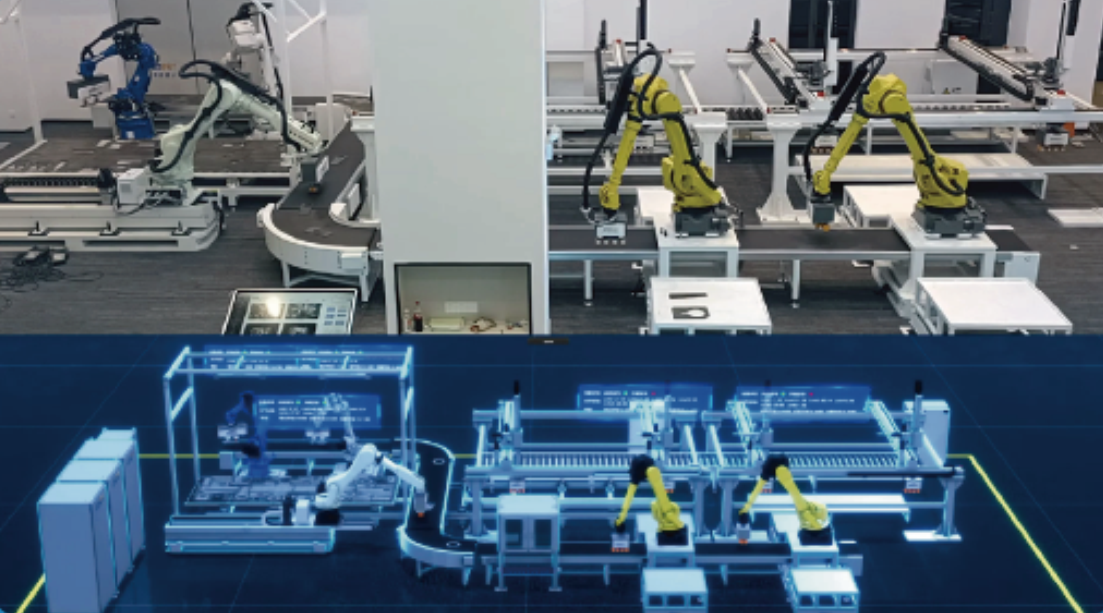  }
    \bicaption{工业级数字孪生}{Industrial digital twin}
    \label{digital_twin}
\end{figure}

在柔性制造场景下，生产模式多样、流程频繁变更、设备配置复杂，对系统的实时感知与智能决策提出更高要求。传统依赖经验和静态建模的方法难以应对故障、磨损和工况扰动等动态变化，常导致响应滞后与调度僵化。
数字孪生（Digital Twin）\ucite{dong2024digital1,liu2019digital,leng2019digital}技术通过将物理系统实时映射至虚拟空间，实现制造过程的全面感知与智能优化，提升系统的柔性适应性与调度效率，如图~\ref{digital_twin} 所示。
国际标准化组织 ISO 将其定义为“对可观察的制造元素的数字映射，并与实际的制造元素之间保持同步”\ucite{shao2024manufacturing}。
本节将从虚拟模型创建、状态感知同步与基于孪生体的智能调度三方面探讨其在柔性制造中的应用。

\begin{figure}[h!t]
    \begin{minipage}{0.46\textwidth}
    \centering
    \subfigure[场景级孪生]{
        \includegraphics[width=\linewidth]{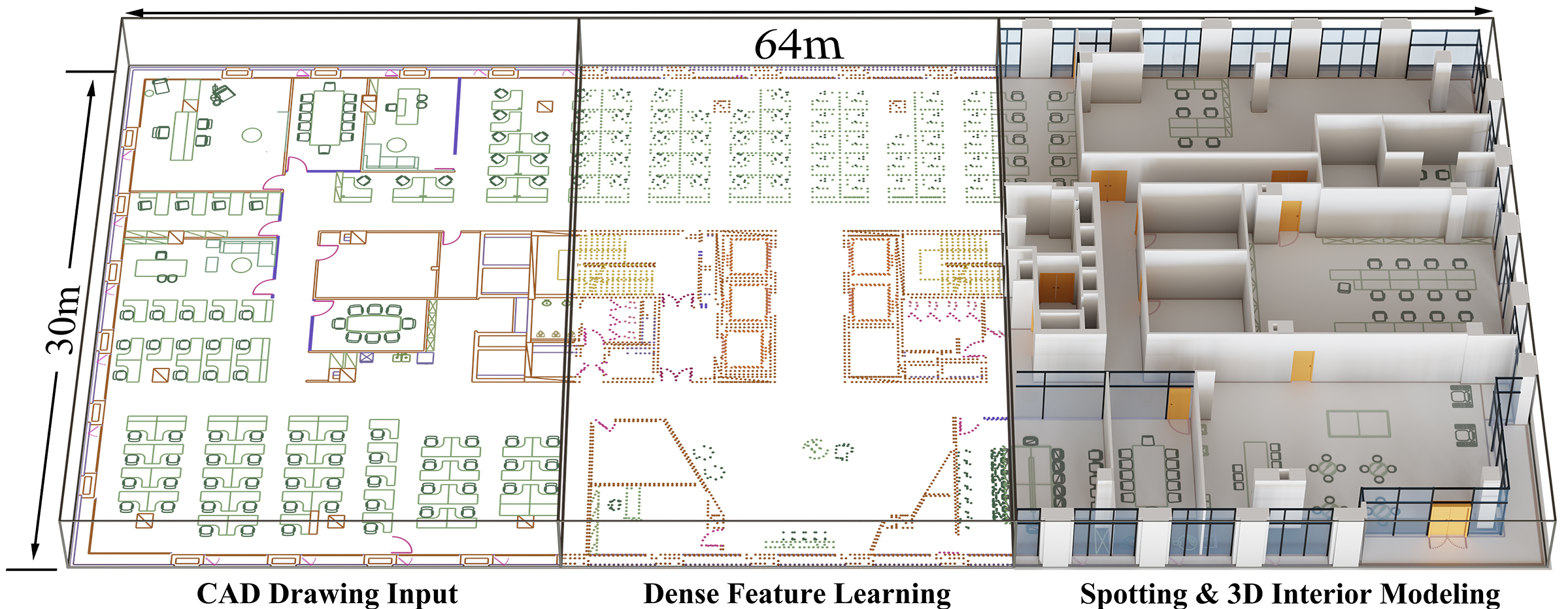}}
    \end{minipage}
    \begin{minipage}{0.46\textwidth}
    \centering
    \subfigure[物体级孪生]{
        \includegraphics[width=\linewidth]{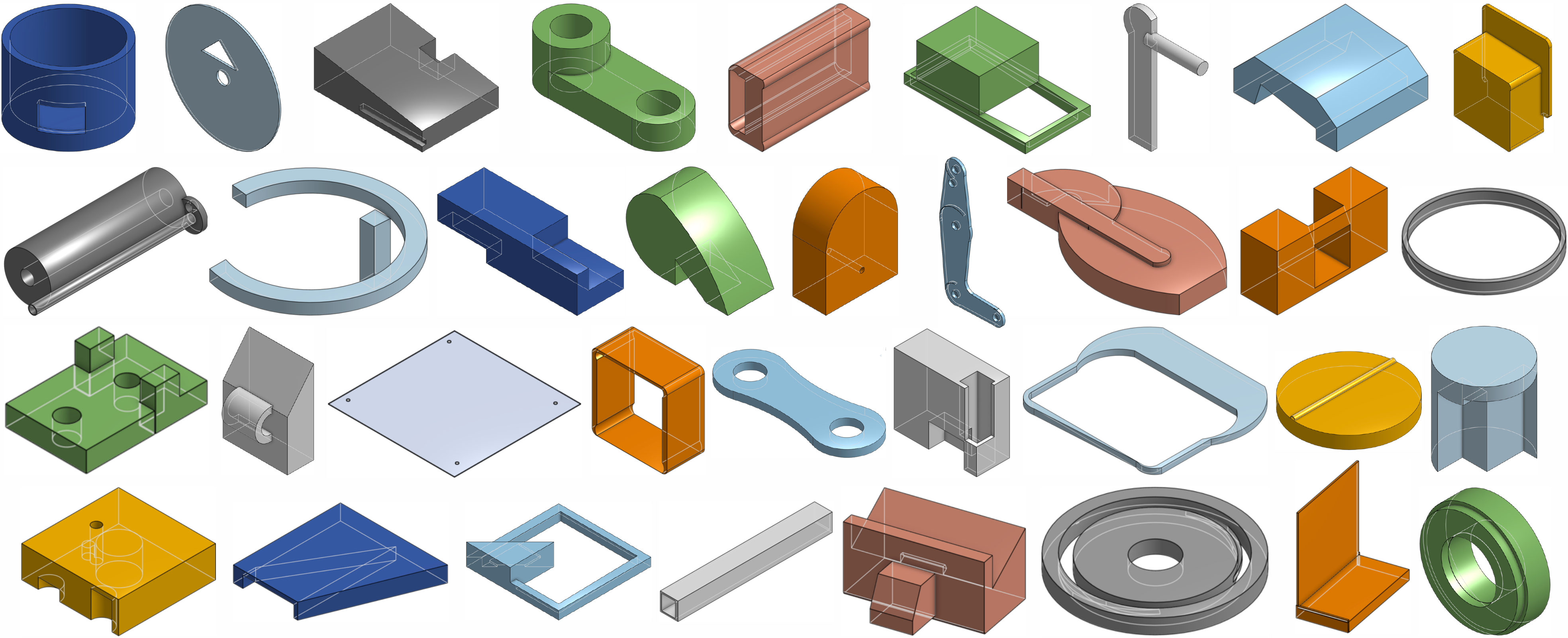}}
    \end{minipage}
    \bicaption{虚拟模型创建\ucite{mu2024cadspotting,wu2021deepcad}
    }{ Virtual model creation}
    \label{fig:digital_modeling}
\end{figure}

\paragraph{虚拟模型创建}
旨在将物理资产的几何结构、物理属性与行为逻辑数字化重构，为数字孪生的感知与决策提供基础。在柔性生产中，因订单变化、产品切换与设备升级，车间布局与工艺流程需频繁调整，传统重建方法难以兼顾时效与精度。工业场景具备 CAD 图纸与模型库等结构化资源，可通过自动解析、参数化与语义检索实现高效建模与更新。
本节将从场景级和物体级虚拟模型创建（图~\ref{fig:digital_modeling}）进行介绍。

在工业数字孪生中，场景级创建通常会先对 CAD 图纸进行全景符号检测（Panoptic Symbol Spotting），识别可数的设备符号与不可数的语义区域，实现图纸的全局解析。
传统基于手工特征与滑窗搜索的方法效率低，难以适应大规模图纸。Nguyen 等人\ucite{nguyen2009symbol}提出基于向量模型与倒排索引的文本化索引方法，通过构建几何不变的视觉词汇，实现图形文档的快速匹配。深度学习方法则借助神经网络自动学习复杂符号特征，如 Fan 等人\ucite{fan2022cadtransformer}提出的 CADTransformer 框架，融合邻域感知注意力、分层特征聚合与图层重组增强，提升异形符号检测鲁棒性。完成检测后，可基于几何原语与空间参数从模型库中生成三维构件，借助参数化重建管线快速构建结构完整的场景模型\ucite{mu2024cadspotting}。
Dong 等人\ucite{dong2024digital1}提出基于大语言模型的自动化代码生成框架，将自然语言需求映射为建模代码；Wang 等人\ucite{wang2024parametric} 提出的两阶段网络可将手绘草图转化为高质量 CAD 图纸，进一步降低模型创建门槛。

数字孪生不仅需重建整体场景结构，还要求精准还原物体几何形态。然而，工业现场零部件种类繁多、结构复杂，尤其在船舶与航空等行业，常涉及万级异构构件，包含曲面、倒角、连接孔等微特征。在小批量柔性生产中，频繁扫描重建代价高、效率低，可借助现有 CAD 模型库进行替代。通过点云或图像提取几何特征，在模型库中检索相似 CAD 模型替换原始物体，实现高效、可扩展的物体级建模。
Sun 等人\ucite{孙志强2024}基于神经渲染获取点云信息，并构建语义映射网络实现点云到 CAD 模型的匹配，显著降低建模成本。Agapaki等人\ucite{agapaki2022geometric}提出一种基于特征融合的匹配方法，将图像特征与点云特征进行融合以实现CAD模型的匹配。Long 等人\ucite{automation3040028}则提出“双层包围盒 + 多视角比对”方法，先筛选尺寸匹配模型，再精细评估相似度，实现快速精准的 CAD 检索。

\begin{figure*}[h!t]
    \centering
    \includegraphics[width=1.0\linewidth]{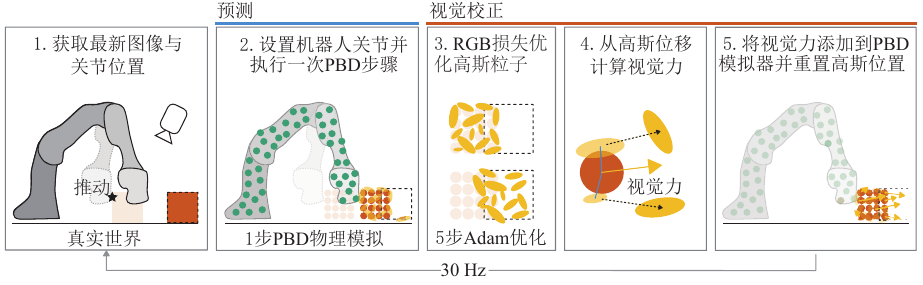}
    \bicaption{利用真实图像与虚拟渲染图像的“视觉力”对仿真状态进行迭代修正\ucite{abouphysically}}{Iterative correction of simulation states via ``visuo-force"  between real and rendered images}
    \label{fig:Embodied_Gaussian}
\end{figure*}

\paragraph{状态感知与虚实同步}在完成虚拟模型构建后，数字孪生需依托状态感知技术实现对物理实体运行状态的实时追踪和同步映射，使虚拟模型具备持续演化与闭环反馈能力。这一过程能够为预测性维护、动态优化与智能控制提供基础。

在柔性制造中，车间状态高频波动、多源异构且结构复杂，给状态感知与虚实同步带来多重挑战：
设备状态通常由视觉、力觉、振动、温度等多种类型传感器共同感知，数据维度与更新频率不一致；
传感器原始数据存在噪声、延迟与时序不一致，需在语义层面对不同模态进行融合与对齐；
同步不仅是数据映射，更要求在虚拟模型中保持拓扑结构、动态行为与语义状态的一致性，并可支持前向预测与后向校正。
为解决上述问题，研究者提出了多种高层次建模与感知策略。
Jaoua 等人\ucite{jaoua2024novel}则从生产仿真角度出发，分析了基于实时数据流驱动的孪生建模方法，强调状态感知对动态产线建模的支撑作用。
Ma等人\ucite{ma2025research}基于传感数据与装配行为序列，构建自动化生产场景下的同步孪生系统，实现对实际装配状态的动态映射。
Macías 等人\ucite{macias2024data}构建了一个覆盖数据采集、融合、质量控制与演化反馈的全生命周期状态感知框架，为复杂工业系统中的数据驱动状态建模提供了标准流程。
Abou-Chakra 等人\ucite{abouphysically}设计了一种结合可微渲染与可微物理的高斯粒子模型，利用真实图像与虚拟渲染图像的“视觉力”对仿真状态进行迭代修正，实现仿真模型对真实世界状态的同步与前向预测能力（图~\ref{fig:Embodied_Gaussian}）。

在工业场景中，遮挡和观测空间受限常导致状态观测不完整，影响虚拟模型的同步精度。
Lee等人\ucite{lee2024occlusion}构建了适用于杂乱环境的端到端双关联点自编码器，用于不完全点云的恢复。
Wang等人\ucite{WANG2024110576}引入遮挡感知几何对齐模块，学习被遮挡物体的几何特征并辅助状态估计。
Qin等人\ucite{qin2023ippe}提出一种基于点云修复的工业零件状态估计方法，首先使用语义分割网络定位被遮挡目标零件的像素区域，然后借助点云修复网络对点云进行修复，最后使用模版匹配技术得到当前零件状态。
Zhuang等人\ucite{Zhuang2024Sparse}提出一种两阶段点云神经网络框架，第一阶段设计了适用于低纹理环境的场景网络，对点云进行实例分割并给出初步状态估计；第二阶段设计了状态精化网络，基于上一步输出进一步提升状态估计精度。

\paragraph{数字孪生驱动的智能制造}通过状态感知技术获取实时数据后，
数字孪生技术能够借助虚拟空间对生产过程进行动态分析、风险预测与策略优化，驱动流程灵活调整与资源高效配置。本节聚焦其在生产优化、扰动响应与预测性维护中的应用成效。

在生产流程动态映射与优化方面，Ding 等人\ucite{ding2023implementation}提提出基于数字孪生的非标设备智能装配方法，构建装配结构与过程的双层映射模型，结合几何特征更新与快速评估提升装配精度与效率。
Zhu 等人\ucite{zhu2021intelligent}将数字孪生引入上下料产线，融合虚实数据与知识图谱，实现上下料顺序与工时的动态优化。
Yang等人\ucite{YangHSZSSCQWS23}针对PCB装配中机器人遮挡问题，提出基于单应性校准的数字孪生测量方法，消除视角误差，优化插装策略。
Liu等人\ucite{liu2014cloud}构建面向产业集群的移动云制造系统，通过资源虚拟化和复合访问控制，实现跨企业协同设计与任务调度。
西门子利用 Siemens Xcelerator\ucite{siemens2025xcelerator}与 NVIDIA Omniverse\ucite{nvidia2025omniverse}提供的双向数据接口，打通虚拟工厂与物理车间，实现排产仿真、冲突检测与调度优化的实时闭环。

在扰动弹性响应方面，Yue等人\ucite{yue2024disturbance}提出一种基于数字孪生的调度扰动评估与动态响应方法，利用因果图识别扰动源，并通过卷积网络量化其影响，触发差异化调度策略，实现快速稳定响应。
Wang等人\ucite{wangyuefei}针对虚实交互中的响应延迟与偏差问题，提出端-边-云协同的混合调度方案，结合自适应多因子遗传算法优化大规模并行任务调度。
Leng 等人\ucite{leng2019digital}则针对资源动态组织困难，提出并行控制方法，利用分布式数字孪生提升物理与数字系统的协同能力。

在生产线监控与维护方面，Jia等人\ucite{jia2024digital}提出动态演化的数字孪生系统，通过多模态建模与生成式AI增强，实现设备状态的持续感知与自适应维护。
Jin等人\ucite{jin2024research}基于三维虚拟工厂模型与标准通信协议，开发离散工厂实时控制系统，提升监控与管理效率。
Liu等人\ucite{liu2019digital}针对个性化制造中的设备集成与不确定性挑战，提出“静态配置–动态执行”的双层优化策略，融合分布式仿真与多目标优化算法，有效提升系统性能并缩短设计周期。

\paragraph{案例研究1：异构任务云-边-端协同调度}
多任务、多资源协同的智能制造场景下，调度系统往往面对任务异构、资源分布动态、通信干扰复杂等多重挑战。然而，传统的集中式调度系统难以适应动态环境，服务器资源利用率失衡导致生产效率损失。数字孪生能够动态反映系统中的资源使用情况，为复杂工业场景提供有效的智能协同决策。

\begin{figure}[h!t]
    \centering
    \includegraphics[width=1.0\linewidth]{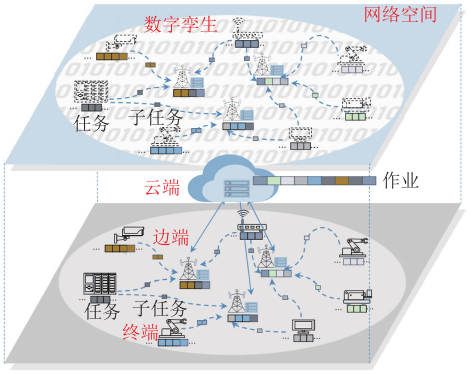}
    \bicaption{基于数字孪生的多智能体协同调度系统\ucite{xu2023digital}}{Digital twin-driven collaborative scheduling system
}
    \label{fig:digtial_twin_scheduling}
\end{figure}

中国科学院研究团队\ucite{xu2023digital}构建了一个基于数字孪生的多智能体协同调度框架。如图~\ref{fig:digtial_twin_scheduling} 所示。系统采用“单云-多边-多端”架构，对终端设备与边缘服务器进行建模。在数字孪生环境中通过多智能体深度强化学习联合训练，并将策略部署至各终端分布式执行，实现任务划分、资源匹配、功率控制与并行执行的联合优化，有效提升任务完成率与系统资源利用率，支持异构任务高效调度。基于数字孪生的智能调度决策实现了对计算资源、通信状态和任务分布的实时感知与镜像，相较于传统算法展现出更强的稳定性与泛化能力，为工业具身智能在调度领域的落地提供了可行路径。

\paragraph{案例研究2：工业软件低代码开发}
在柔性制造快速发展的背景下，产线配置、工艺流程与设备布局需频繁重构以适应多品种、小批量的生产需求，对传统工业软件提出更高要求。
借助数字孪生技术，可以构建工业智能软件低代码（Low-Code）平台，打造一站式工业流程智能重构方法。
通过快速构建工厂实景孪生，实现虚拟调试和一键部署，降低实物验证成本，高效优化项目生命周期管理。
工业智能软件低代码平台可以赋能用户快速构建和迭代各类工业应用场景，实现软件的模块化封装和灵活复用，引领工业制造进入高效、敏捷的软件时代。

\begin{figure}[h!t]
    \begin{minipage}{0.46\textwidth}
    \centering
    \subfigure[低代码虚拟调试]{
        \includegraphics[width=\linewidth]{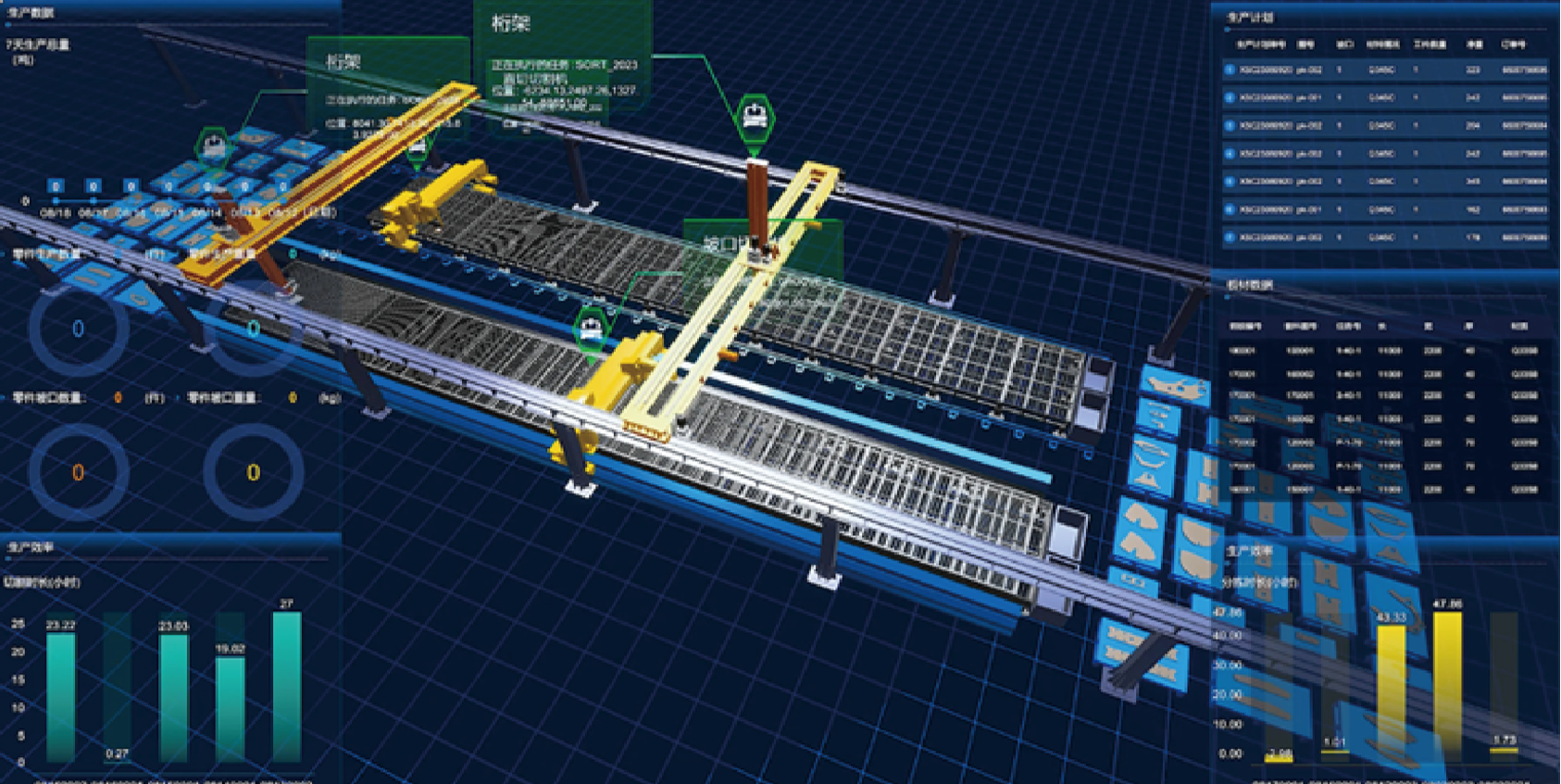  }
        \vspace{-20pt}
        }
    \end{minipage}
    \begin{minipage}{0.46\textwidth}
    \centering
    \subfigure[真实场景高效部署]{
        \includegraphics[width=\linewidth]{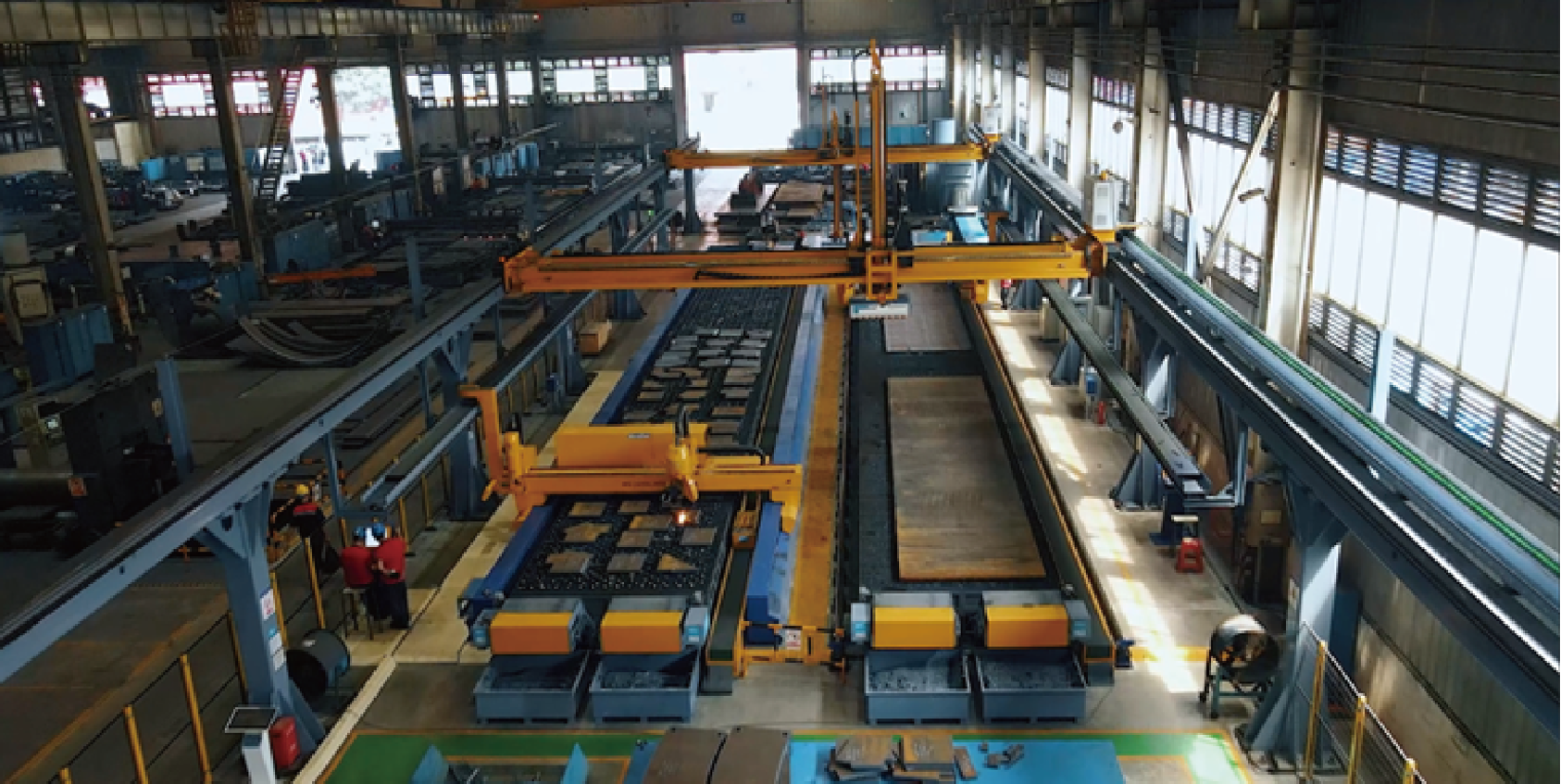  }     \vspace{-20pt}
        }
    \end{minipage}
    \bicaption{工业软件低代码平台}{Low-code platform for industrial software}
    \label{fig:kunwu}
\end{figure}

结合数字孪生与低代码平台，研究者构建了面向用户的图形化、模块化工业智能软件架构\ucite{kunwu}，以支持柔性生产下的快速调试与验证流程（图~\ref{fig:kunwu}）。该架构将数字孪生流程划分为模型构建、模型运行、连接交互、控制逻辑、可视化与人机交互五大模块，并通过图形化组件实现低代码封装。模块化设计增强了系统的灵活性与复用性，内置功能库与可视化编程环境降低了开发门槛，使用户可通过拖拽方式快速配置控制策略。目前该平台已在实际产线中验证，简化了孪生系统构建流程，提升了落地效率。

\subsection{世界模型物理感知}
除了实现对现实状态的同步映射，柔性制造环境还要求系统具备对动力学过程的精确建模与未来状态的预测能力，以支撑前瞻性决策与智能控制。
世界模型（World Model）\ucite{ha2018world}通过模拟生产过程中的关键变量，为策略训练提供可交互的虚拟环境\ucite{HafnerLB020, HansenSW22}。
然而，由于受限的观测和感知精度，传统的世界建模方法常常无法全面捕捉生产过程的动力学信息，影响决策效果。
将物理机理作为先验引入学习算法已被证明能够在训练数据有限的情况下提升泛化能力\ucite{raissi2019physics}。 表~\ref{tab:wm_mechanism_comparison} 中介绍了物理先验越丰富，模型越具泛化性，所需数据也相应减少。
本节将先介绍传统数据驱动的世界模型构建方法，进而探讨如何在观测受限条件下融合物理先验提升外推能力，并介绍基于世界模型的策略学习机制。

\begin{table*}[ht!]
  \bicaption
    {\zh{不同世界模型的特性对比}}
    {\zh{Comparative characteristics of different world‑model approaches}}
  \label{tab:wm_mechanism_comparison}
  \centering
  \footnotesize
  \setlength{\tabcolsep}{0.55em}
  \resizebox{\linewidth}{!}{
  \begin{tabular}{c|c|c|c|c|c}
      \toprule
      \textbf{类别} & \textbf{机理融合程度} & \textbf{数据需求} & \textbf{泛化能力} & \textbf{主要优势} & \textbf{主要局限}\\
      \midrule
      数据驱动世界模型 &
      低，纯依赖观测数据，无显式物理约束 &
      高，需要大量高质量样本 &
      较低，分布外表现差 &
      \makecell[l]{  能捕捉复杂非线性;\\  易与RL/规划集成} &
      \makecell[l]{  对数据需求量大;\\  缺乏物理一致性解释}\\
      \midrule
      物理信息神经网络 &
      中，数据拟合 + PDE 残差双约束 &
      中，少量数据即可训练 &
      中—高，未见条件仍守物理 &
      \makecell[l]{  可解释且外推好;\\  能反演未知参数} &
      \makecell[l]{  多物理耦合与时序预测仍难;\\  训练收敛慢}\\
      \midrule
      可微物理世界模型 &
      高，可微物理/渲染与学习深度耦合 &
      低，只需少量交互数据 &
      高，对新任务新场景稳健 &
      \makecell[l]{  少样本快速收敛;\\  端到端可解释} &
      \makecell[l]{  计算量大;\\  引擎构建复杂}\\
      \bottomrule
  \end{tabular}
  }
\end{table*}

\paragraph{数据驱动的世界模型}

在早期的世界模型研究中，研究者主要依赖数据驱动的方法来构建环境动力学模型。
Hafner等人提出了 Dreamer\ucite{HafnerLB020}，一种学习环境动态紧凑潜在表示的世界模型。  
后续工作\ucite{WuEHAG22}将 Dreamer 应用于机器人操作任务，展示了在物理机器人上快速策略学习的能力。  
DINO-WM\ucite{zhou2024dino}利用通过 DINOv2 预训练的空间特征来学习世界模型，并通过将目标特征作为预测目标，实现了与任务无关的行为规划。  
TD-MPC\ucite{HansenSW22,00010024}使用面向任务的潜在动态模型进行局部轨迹优化，并采用所学习的终值函数进行长期回报估计，在基于图像的控制任务中表现出优良性能。  
在从大规模数据集学习取得成功的基础上\ucite{brown2020language,radford2021learning}，研究\ucite{MendoncaBP23}利用互联网规模的视频数据来学习一个以人类动作空间为基础的世界模型。  
然而，纯数据驱动的世界模型严重依赖训练数据的数量和质量，难以泛化到分布外场景\ucite{yu2020mopo,rafailov2021offline}，降低了所学策略在真实环境中的鲁棒性。

\paragraph{物理信息神经网络}

物理信息神经网络（Physics-Informed Neural Network, PINN）\ucite{raissi2019physics}是一类将物理规律（通常以偏微分方程形式）融入神经网络训练过程的框架。其核心思想是：在网络的损失函数中，不仅包含与观测数据的拟合误差，还添加偏微分方程（PDE）残差项，使网络输出同时满足数据驱动与物理方程约束。
由于物理规律的引入，PINN 在未见参数配置或边界条件下往往比纯数据驱动模型表现出更好的外推性能。
本文将以焊接工艺为切入点，详细介绍在这一典型高维非线性热—流—相变复杂物理过程中，PINN 如何发挥建模优势。

焊接过程涉及热传导、材料相变与熔池行为等复杂物理机制，常由 PDE 刻画，传统数值解法在高维非线性场景下计算开销大，难以实时应用。PINN 通过将热传导等物理约束嵌入损失函数，以少量数据拟合温度场，同时确保物理一致性，实现高效、精确的过程建模。
Liao 等人\ucite{liao2023hybrid} 提出基于 PINN 的热建模方法，利用红外测温与 PDE 联合训练反演未知参数并预测全场温度，兼顾数据驱动与物理可信性。
Zhu等人\ucite{zhu2024reality}面向激光增材制造中多物理耦合场景，构建融合增强数据与自适应约束的 RAA-PIML 框架，在热传导 PDE 和边界条件约束下联合训练红外测温数据与高精仿真数据，提升预测精度并加速收敛。
Sharma等人\ucite{sharma2024physics}针对 Navier-Stokes 方程求解开销大问题，提出无速度数据的 PIML 方法，仅凭 PDE 残差预测温度、速度与压力场，并反演湍流黏度，实现高能束下热-流耦合的实时建模。
Zhu等人\ucite{zhu2025transfer}为应对 PINN 在多物理耦合下的性能瓶颈，引入迁移增强的 TLE-PINN，通过先在高保真仿真上预训练再微调输出，实现熔池形貌与温度场的准确预测，验证了其在工业过程控制中的应用潜力。

需要注意的是，现有焊接领域的 PINN 方法多聚焦于当前时刻物理场的重建与参数反演，本质上更接近静态场映射器，而非具备时序建模能力的动态“世界模型”。然而，在焊接等高速动态制造过程中，具备对未来状态的预测能力对于实现在线控制、路径优化与闭环反馈至关重要。提升 PINN 对动态演化过程的建模与预测能力，是其迈向高可控性与实用性工业建模工具的关键方向。

\paragraph{基于可微物理的世界模型}

近年来，可微物理与可微渲染技术的进展为将物理知识融入世界模型提供了新的可能性。  
Lutter等人\ucite{LutterRP19}引入了一种基于拉格朗日力学的深度网络框架，能够高效学习运动方程并保证物理合理性。  
Heiden 等人\ucite{heiden2021neuralsim}在可微刚体物理引擎中引入神经网络，以捕捉动态量之间的非线性关系。  
$\nabla$Sim\ucite{MurthyMGVPWCPXE21}结合可微物理\ucite{2DLCP,3DLCP}与渲染技术\ucite{chen2024neural,jing2024frnerf, yang2023jnerf}，联合建模场景动力学与图像生成，实现了从视频像素到物理属性的反向传播。  
随后，研究者又通过先进的渲染技术\ucite{LiQCJLJG23,caoneuma}或增强的物理引擎\ucite{kandukuri2024physics}对该方法进行了改进。

尽管在物理属性估计方面取得了一定进展，仅有少数研究\ucite{memmel2024asid, baumeister2024incremental, SongB20, li2025pinwm}将物理属性估计融入用于机器人操作的世界模型。
ASID \ucite{memmel2024asid}通过无梯度方法进行系统辨识，依赖高质量轨迹，如果缺乏此类数据，该方法易陷入局部最优。  
Song等人采用的简化的二维物理引擎\ucite{SongB20}，尽管该仿真器建模了可微通道，但是二维物理引擎的局限性使其难以捕捉复杂的三维交互，从而导致系统辨识不准确。  
相比之下，PIN-WM 利用少量任务无关（Task-Agnostic）的交互数据，通过可微渲染通道\ucite{2DGS}从视觉观测中端到端地辨识三维刚体动力学\ucite{3DLCP}，促进基于视觉的操作策略的强化学习训练。

\paragraph{基于模型的控制方法}
在世界模型的基础上，基于模型的控制方法（Model-Based Control）通过预测系统未来状态，实现对复杂任务中的决策效率与策略泛化能力的提升。
该方向主要包括两类路径：一类是基于模型预测控制（Model Predictive Control, MPC）的方法，利用世界模型对短期未来轨迹进行优化；另一类是基于策略学习的方法，通过在学习到的世界模型中模拟交互，以训练出可部署的策略网络。两种方法的特性对比总结在表
~\ref{tab:wm_control_features} 中。

\begin{table}[htbp]
  \centering
  \footnotesize
  \bicaption{\zh{基于模型的控制方法对比}}%
            {\zh{Comparisons of model-based control methods}}
  \label{tab:wm_control_features}
  \setlength{\tabcolsep}{0.6em}
  \resizebox{\linewidth}{!}{
  \begin{tabular}{>{\raggedright\arraybackslash}m{2.4cm}|
                  m{0.40\linewidth}|m{0.40\linewidth}}
    \toprule
    \makecell[c]{\textbf{特性}} &
    \makecell[c]{\textbf{模型预测控制}} &
    \makecell[c]{\textbf{策略学习}} \\
    \midrule
    \makecell[c]{决策方式}  &
    在线滚动规划，在潜在或显空间搜索多步最优控制序列 &
    在世界模型中生成虚拟轨迹，离／在线更新并部署策略 \\
    \midrule
    \makecell[c]{运行计算量}  &
    运行时成本高，随预测视窗和采样数增加，需要并行算力 &
    训练阶段大量模拟；部署阶段仅一次前向推理，计算轻量 \\
    \midrule
    \makecell[c]{模型误差敏感度} &
    高，预测误差在滚动优化中累积，需要短视窗或不确定性惩罚缓解 &
    中，可用真实数据混合或短模拟片段缓解偏差 \\
    \midrule
    \makecell[c]{主要优势} &
    能即时重规划，应对突发扰动，
    规划过程可编辑可解释 &
    部署实时性好，适合高频控制
    \\
    \midrule
    \makecell[c]{主要局限} &
    在线计算成本高
    &
    策略训练不稳定
    \\
    \bottomrule
  \end{tabular}}
\end{table}

模型预测控制最初源于工业过程控制领域，其核心思想是将预测建模与滚动优化相结合，在考虑系统约束的前提下生成最优控制序列。
Mayne 等人\ucite{mayne2000constrained}等人的经典研究为 MPC 的稳定性与最优性分析奠定了理论基础，标志着该方法从工程启发走向控制理论系统化的发展。
随着算法发展，MPC 被不断扩展至高维非线性系统控制。
MPPI\ucite{williams2017information}作为一种采样式 MPC 方法，通过路径积分框架对随机轨迹加权优化，适用于高维连续控制问题。
Lucia 等人\ucite{lucia2018deep}引入神经网络对复杂系统建模，有效提升了 MPC 在非线性系统中的表现能力。
Ramp-Net\ucite{sanyal2023ramp}将 PINN 嵌入 MPC 框架，以对未来一段时域的状态演化进行预测，以实现对四旋翼在不确定动力学下的鲁棒自适应控制。
在视觉控制领域，PlaNet\ucite{hafner2019learning}首次将隐空间动力学建模与 MPC 相结合。该方法从图像序列中学习潜在状态空间，并在该空间中进行预测与控制，展示了从像素输入到动作输出的端到端控制能力。
MuZero\ucite{schrittwieser2020mastering}则通过端到端学习预测奖励、策略与状态值，结合蒙特卡洛树搜索（MCTS）实现高效规划，在未知环境中展现出极强的泛化能力。
TD-MPC 系列方法\ucite{HansenSW22, 00010024}进一步引入任务导向的终值函数估计机制。通过在隐空间中优化轨迹，并结合时序差分学习提升长期规划性能，已在多项视觉控制与真实机器人任务中验证其高效性与鲁棒性。

另一类方法则聚焦于利用世界模型训练可部署策略，通过在“脑中演练”进行虚拟交互、策略更新与行为规划。
Sutton 提出的 Dyna 框架\ucite{sutton1990integrated}是这一方向的奠基之作，首次将环境模型引入强化学习过程，统一了学习与规划。
MBPO\ucite{yu2020mopo}强调将模型预测限制在短时间范围内以减缓误差传播问题。使用真实环境数据频繁更新世界模型，并生成短期虚拟交互轨迹以支持离策略策略更新，有效提高了数据利用效率与训练稳定性。
SimPLe\ucite{kaiser2019model}针对高维图像输入环境（Atari 游戏）构建世界模型并在模拟中训练策略，是最早将世界模型应用于图像游戏控制任务的成功案例之一。
Dreamer 系列\ucite{HafnerLB020, hafner2020mastering, hafner2023mastering}通过在潜在动态模型中采用策略梯度训练，在学习潜在世界模型的同时优化策略网络，在多个控制任务中实现了极高的样本效率与泛化能力，已成为当前最具代表性的模型学习强化学习框架之一。
DINO-WM\ucite{zhou2024dino}进一步拓展世界模型的构建方式，利用 DINOv2 等自监督视觉特征建立通用世界模型，不依赖任务奖励即可训练策略，体现了从“任务驱动建模”向“任务无关表征学习”的演进趋势。

\paragraph{案例研究1：结合机理的焊接内部温度场捕捉}

\begin{figure}[h!t]
    \centering
    \includegraphics[width=1.0\linewidth]{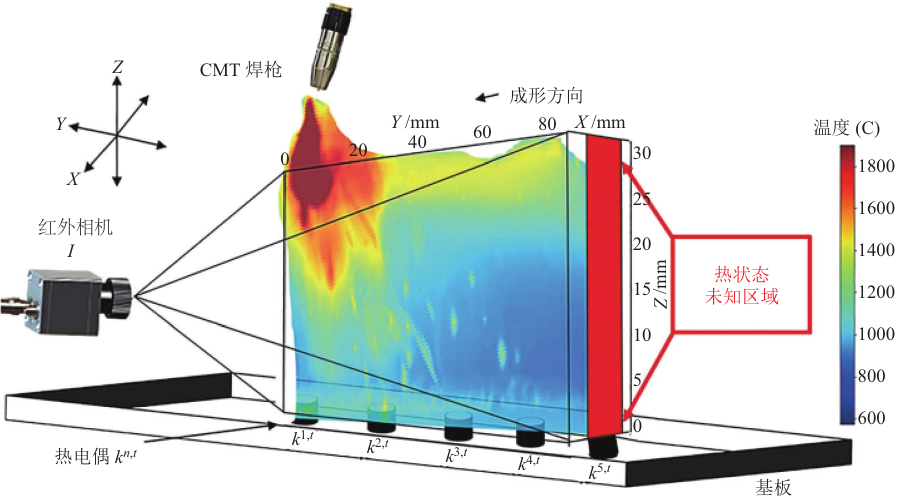  }
    \bicaption{常用传感器仅能观测材料表面温度分布}{Common sensors can only observe  temperature distribution on material surface}
    \label{partial_welding}
\end{figure}

在弧焊过程中，表面变形是一个常见的问题，其主要是由于材料内部快速变化的热梯度引起的。在增材制造过程中，材料在高温下被逐层堆积，每一层的加热和冷却都会导致材料的膨胀和收缩，从而产生变形。这种变形会影响部件的几何精度和功能性，尤其是在制造大型或复杂结构的部件时。
传统的非破坏性检测方法，如红外热成像和热电偶传感器，
通常只能提供表面的温度信息，而无法准确地捕捉到材料内部的温度分布，如图~\ref{partial_welding} 所示。
\zh{在制造过程中，典型钛合金零件的焊接线能量输入可导致局部温度达到1500 $^\circ\mathrm{C}$，内部热梯度往往在800 - 1200 $^\circ\mathrm{C}$/cm之间快速变化。
}
这一高热梯度会在冷却阶段引发0.1–1.0mm级别的热变形。
通过表面温度数据来预测内部变形非常重要且具有挑战性。

Zamiela 等人\ucite{zamiela2024physics}基于热通量方程与表面红外热图数据，提出了一种融入物理先验的表面变形预测模型，填补了增材制造过程中内部热场与表面形变关联研究的空白。
该方法采用回归型卷积神经网络自动学习三维热梯度与表面变形之间的复杂映射，既能高效处理高维热图数据，又能提升预测精度。
与此同时，研究团队利用有限差分法模拟热历史，并结合热物性知识对数据驱动近似进行校正，从而进一步优化模型性能。
实验结果显示，该模型在表面变形预测上获得了更低的误差，验证了其在焊接制造质量控制中的应用潜力。

\paragraph{案例研究2：螺栓拧紧建模与策略学习
}

在工业装配中，螺栓拧紧对机器人极具挑战：摩擦系数随表面状态与载荷剧烈变化，加之零件公差、装配误差与结构弹性变形共同作用，使得扭矩-预紧力关系高度不可预测；高精度力/扭矩传感器成本高且难以布设，实时反馈受限；且过紧易损螺纹、过松又导致松动和疲劳失效，需在速度与精度间精准平衡。
依靠现场试验获取大规模高精度数据既昂贵又耗时，而构建高保真的拧紧“世界模型”，通过仿真重现螺母-
螺栓接触与摩擦行为，成为优化拧紧策略的可行方案。
机器人可在虚拟的世界模型中试错探索、自动调整，实现从策略学习到精度验证的闭环。

\begin{figure}[h!t]
    \begin{minipage}{0.46\textwidth}
    \centering
    \subfigure[螺母与螺栓的高保真网格与 SDF 表示
    ]{
        \includegraphics[width=\linewidth]{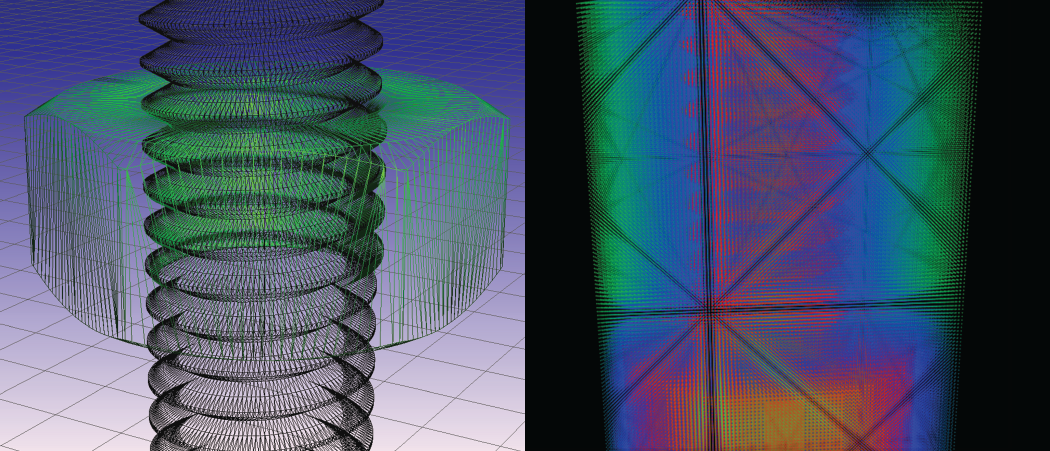  }}
    \end{minipage}
    \begin{minipage}{0.46\textwidth}
    \centering
    \subfigure[\centering 螺母拧紧至完全就位的仿真过程]{
        \includegraphics[width=\linewidth]{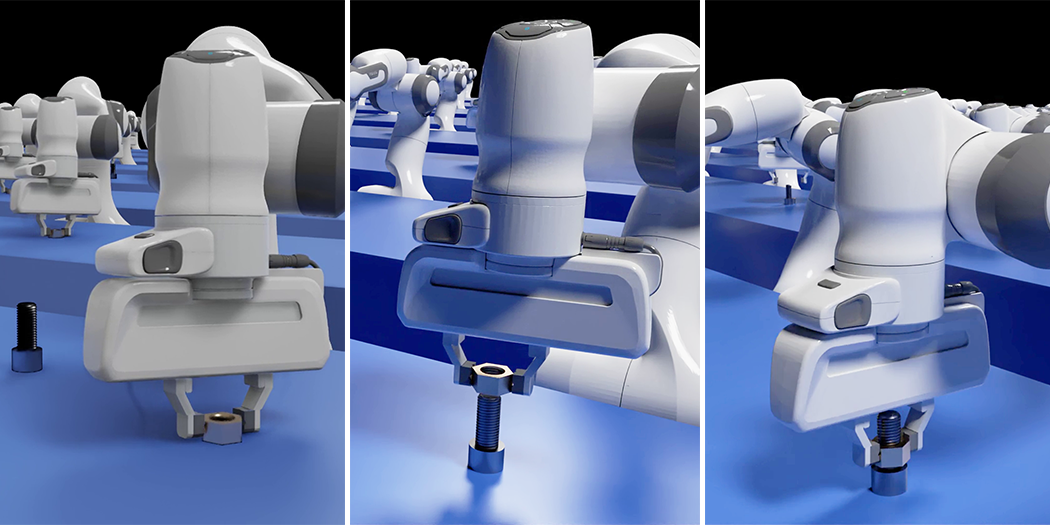  }}
    \end{minipage}
    \bicaption{拧紧过程的数字化模型和仿真过程\ucite{NarangSAMRWGMSL22}
    }{Digital modeling and simulation of bolt tightening 
    }
    \label{fig:tightening_simulation}
\end{figure}

Nvidia提出了一套可用于机器人拧紧过程的模拟方法 Factory\ucite{NarangSAMRWGMSL22}。通过结合有符号距离函数（Signed Distance Field, SDF）碰撞检测、接触减少和高斯-赛德尔求解器，Factory 实现了对复杂场景的高效、准确模拟，如图~\ref{fig:tightening_simulation} 所示。通过接触减少技术，Factory能够大幅降低接触数量，在单个GPU上实时模拟1000个螺母-螺栓交互（约为现有方法单对交互模拟速度的 20 倍）。
基于此，NVIDIA 又推出了 IndustReal 框架\ucite{tang2023industreal}，通过模拟感知策略更新、基于 SDF 的密集奖励和采样基课程（SBC），在 Factory 环境中训练机器人执行精密装配任务。该框架不仅能准确建模拧紧动力学，还成功实现了从仿真到现实的策略迁移，大幅提升了装配精度与成功率，并验证了其在高精度机器人装配场景中的有效性。

\begin{figure*}[t!]
    \centering
    \includegraphics[width=\linewidth]{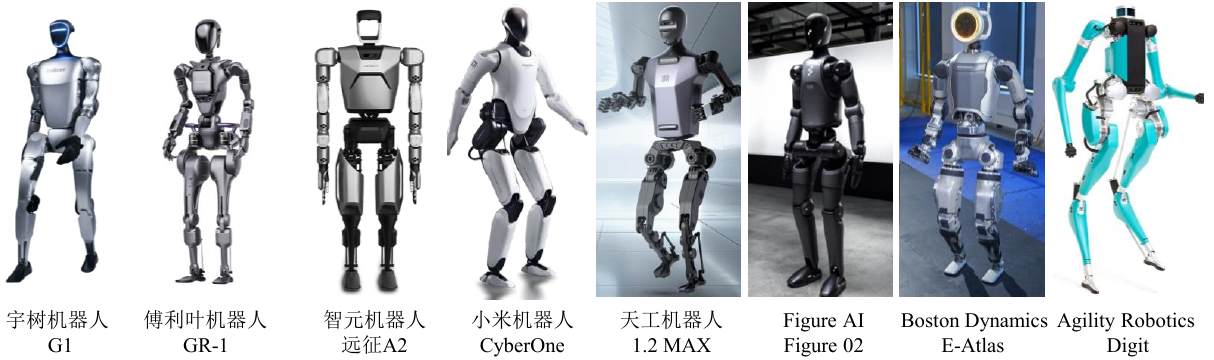}
    \bicaption{\zh{主要人形机器人产品概览
    } }{\zh{Overview of major humanoid robot products}
    }
    \label{technical_pipeline}
\end{figure*}

\zh{
\section{人形机器人与柔性制造 （Humanoid Robots and Flexible Manufacturing）}

人类的生产场所，从车间的装配线、检修平台到仓库的货架系统，往往都是按照人体尺寸、可达范围与操作习惯设计的。例如，装配工位的工装夹具、工具挂架和工控面板高度都与成年工人手臂伸展相匹配；仓储货架的层高、过道宽度以及拣选台的台面高度基于人身尺寸来规划；工业场景中复杂的地形、布线等也需要人形结构进行跨越。因此，人形机器人在工业环境中具备天然的“形态—环境对齐”优势。
借助这类优势，人形机器人可以 零改造 地融入现有工厂环境，并在产品与工序频繁变化的柔性制造场景中快速适配迁移任务。
因人形机器人同时涵盖工业之眼、工业之手与工业之脑三个维度，本文将其单独作为一个章节，介绍人形机器人研究方面的进展，并探讨其在工业领域的应用前景。当前主要的人形机器人产品如图~\ref{technical_pipeline} 所示。

早期的人形机器人研究源于对运动控制（Locomotion）的探索。这些研究首先始于四足\ucite{hoeller2024anymal, kim2025high}与轮-足\ucite{lee2024learning}领域，验证了四足机器人在碎石、金属栈板等工业地面上行走的鲁棒性\ucite{kim2024not}。后面的研究逐渐将注意力转向双足步态控制，尤其是对倒立摆模型的控制。
“倒立摆”是二足机器人在单支撑期的最低阶质心-地面动力学近似。它为 LQR、MPC 等经典控制器提供了可实时求解的线性模型，使高维双足控制问题可在上层被分解为简单、可解析的质心轨迹规划，而下层再负责满足全身多关节、接触与约束。
经典控制方法解决了简单倒立摆模型，但真正推动性能跃迁的是强化学习，最新工作已在仿真-现实并行框架中，将 2 m/s 奔跑、0.4 m 跳跃与斜坡恢复等高动态步态可靠迁移至实机\ucite{radosavovic2024real, li2025reinforcement}。
随后，研究者开始探究极端地形与高难度机动。BeamDoJo\ucite{wang2025beamdojo} 让机器人在宽度不足 10 cm 的浮动踏板上连续跨越，Humanoid Parkour\ucite{zhuang2024humanoid} 则通过分阶段对抗训练实现翻滚、墙跑等链式动作。
全身控制（Whole Body Control, WBC）成为下一阶段焦点。统一的二次规划层次控制器能在 100 Hz 频率下无缝切换走-跑-台阶等细粒度运动\ucite{xue2025unified}，GPU-级二次规划器 AMO\ucite{li2025amo} 把 30+ DoF 在线优化推至 2 kHz 并降低能耗。Homie\ucite{ben2025homie} 提出了一种“同构外骨骼驾驶舱”方案，用于让人类操作者通过穿戴式外骨骼与人形机器人一对一映射，实现行走与操作的复合任务。LangWBC\ucite{shao2025langwbc} 及 OKAMI\ucite{li2024okami} 则展示了端到端视觉-语言-动作映射和单视频模仿，仅凭自然语言或一次示范即可完成抓取、分拣等细粒度操作。}

\zh{工业部署的连续作业中，人型机器人需要具备高可靠性与容错恢复能力。
针对现实场景中类人机器人在工作过程可能遭遇跌倒的情况，He 等人\ucite{he2025learning}采取层次化强化学习结构。高层策略 负责根据当前姿态类别（倒地模式）选择合适的恢复子策略，低层策略 基于反向动力学与阻抗控制，执行具体的起身动作。该方案能够在真实硬件上学习的起身策略在 多 种跌倒模式下成功率达 78.3\%-98.3\%，较初始控制器提升超过36\%，提升了连续作业可靠性。
Huang 等人\ucite{huang2025learning}在仿真环境中收集从不同倒地姿态到站立姿态的多阶段数据，基于强化学习与优先经验回放学习站起策略。让人形机器人能够从多种非标准站立姿态（例如跪姿、侧卧、趴卧等）迅速且安全地恢复到站立姿态。
为缩短部署周期，研究者提出了高效训练与仿真-现实对齐机制。ASAP\ucite{he2025asap} 通过物理标定、双域对抗和在线微调，将 Sim2Real 调参时间削减 60\%。FastTD3\ucite{seo2025fasttd3} 通过网络裁剪与自适应噪声，将 Humanoid-v2 收敛时间缩短 30\%，样本量减半。为了标准化评测与降低硬件门槛，HumanoidBench\ucite{sferrazza2024humanoidbench} 提供了涵盖 15 项行走-操控任务的开源仿真基准。Berkeley Humanoid Lite\ucite{chi2025demonstrating} 用 3D 打印和 ROS 软件栈构建了 22 DoF 教学级平台，方便快速验证算法。}

\zh{人形机器人在智能制造领域的价值日益凸显，尤其在柔性制造场景中表现尤为突出。
2024年，特斯拉发布了其人形机器人Optimus\ucite{tesla2024optimus}，展示了其在工厂环境中自主搬运和分拣物品的能力。
同期，Figure AI 与宝马合作\ucite{reuters2024figure}，在宝马美国制造工厂部署其Figure 01人形机器人，以应对生产线上的重复性劳动。
Sanctuary AI 推出的 Phoenix 机器人则面向通用作业\ucite{venturebeat2024sanctuary}，可执行轻工业环境中的搬运、上货及质量检查等任务，进一步拓展了人形机器人在各类制造场景中的适用性。
在国内市场方面，优必选作为首家上市的人形机器人企业，其工业版机器人Walker S已在蔚来新能源汽车工厂开展实地“实训”\ucite{UBTECHWalkerS2024}。该机器人在车门锁质量检测、安全带检测及车灯盖板检验等关键工序中展现了稳定可靠的性能。智元机器人推出的通用型具身智能机器人“远征 A1”同样令人瞩目\ucite{IyiouYuanZhengA12024}。其已在3C装配线上完成齿轮点油任务，在汽车底盘线上执行底盘装配，并于OK线实现了外观检测，充分验证了通用操作能力。华为云与乐聚机器人联合研发的首款鸿蒙系统人形机器人“夸父”已进入蔚来与江苏亨通集团工厂进行测试\ucite{EETChinaKuafu2024}，能够胜任扫码包装、物流搬运、焊锡等非标作业。
从国际到国内，人形机器人在各类生产场景中的落地实践不断拓展，已逐步成为提升生产线柔性与效率的潜在力量。
}

\section{现有研究的联系、挑战及展望（Connections, Challenges, and Outlook）}

\begin{figure*}[t!]
    \centering
    \includegraphics[width=\linewidth]{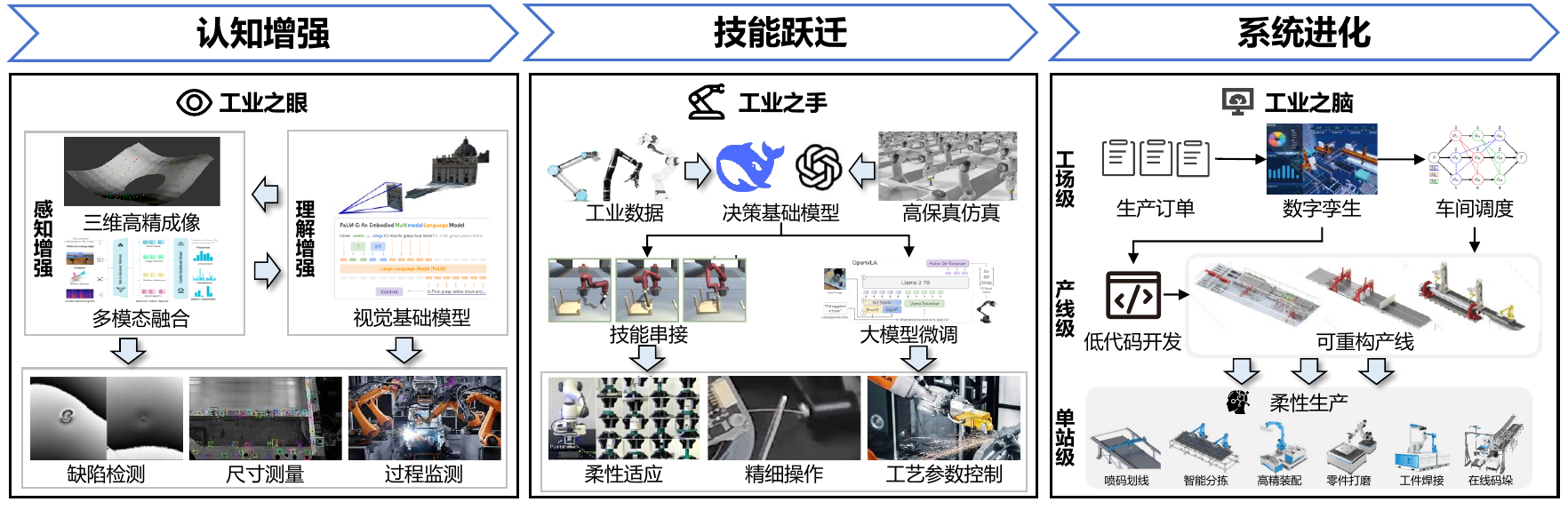}
    \bicaption{\zh{面向柔性制造的具身智能“认知增强-技能跃迁-系统进化”技术路线图} }{\zh{A technology roadmap for embodied intelligence in flexible manufacturing: from cognitive augmentation to skill transition and system evolution}
    }
    \label{technical_pipeline}
\end{figure*}

本文系统梳理了面向柔性制造的工业具身智能研究进展，对感知层（工业之眼）复杂动态环境下的多模态数据融合与实时建模、控制层（工业之手）中复杂制造工艺的柔性自适应精准操控、以及决策层（工业之脑）工艺规划与产线调度的智能优化等内容进行了重点分析。
需要指出的是，这些技术之间并不是相互独立的，而是相互耦合，形成了一个完整的技术闭环，共同促进了工业具身智能在柔性制造中核心挑战的解决与应用的推进。本文将进一步探讨现有研究的联系、仍然存在的挑战以及未来的研究展望。

\subsection{现有研究的联系}

现有研究沿着“工业之眼-工业之手-工业之脑”呈阶梯式耦合，可抽象为柔性制造场景下具身智能的 “认知增强 - 技能跃迁 - 系统进化” 三阶段演进路径。首先，\emph{认知增强}阶段聚焦受限感知下的工艺精准建模，工业之眼借助三维视觉、多模态融合与视觉基础模型，在遮挡、反光与稀疏观测条件下补全环境几何并辨识动力学，为制造系统提供精确且可泛化的环境表征。继而进入 \emph{技能跃迁}阶段，工业之手依托所获物理认知，通过残差控制、元强化学习及 Real2Sim2Real 管线，驱动工业之手在真实世界的低精度产线上实现高精度操作，使柔性适配与工艺精度达到动态平衡。最终在 \emph{系统进化}阶段，数字孪生用端-边-云协同实现虚实同步，工业之脑学习 JSSP/VRP/BPP 与多智能体调度，联动可重构产线等执行终端，完成跨工段、跨工位、跨设备的实时全局优化，推动制造系统由局部自适应迈向全局协同。图~\ref{technical_pipeline} 展示了这一技术路线图。

\subsection{面向柔性制造的工业具身智能挑战与展望}
在人工智能与机器人技术持续突破、制造业加速向柔性化、客制化转型的背景下，工业具身智能研究正面临全新的机遇与挑战。本文系统梳理了当前工业具身智能领域的仍然存在的技术瓶颈，并展望了
未来的发展方向。
（1）\textbf{工业级数据平台建设：} 相较于通用具身智能任务，工业场景中的数据获取具有更多壁垒：一方面，生产环境高度专业化、定制化，数据采集成本高；另一方面，生产过程中的数据往往涉及企业核心机密，企业对数据开放共享存在天然顾虑。因此，未来亟需探索可信数据协同机制，推动构建跨企业、跨系统的数据共享平台，发展支持访问控制、隐私保护与审计追溯的工业数据基础设施，为工业具身智能模型的训练与评估提供保障。
（2）\textbf{高性能高保真物理仿真器：}
通用具身智能中的仿真平台多以刚体动力学为主，难以满足工业任务中对柔性接触、热场、电场等多物理场过程的高保真建模需求。未来需要构建支持多模态、多物理场、高精度过程控制以及任务级反馈的工业仿真平台，兼顾仿真精度与仿真速度，为模型训练与策略测试提供可信赖的试验环境。
（3）\textbf{通用工业控制基础模型：}
目前基础模型在工业控制领域的应用仍处于起步阶段，主要集中在使用预训练模型面向特定任务进行微调，缺乏针对工业控制任务的通用基础模型。而面向应用的工业控制策略往往需要从头开始训练，导致开发成本高、开发周期长。构建支持多工艺、多装备、多任务迁移的通用控制大模型，提升工业控制策略的泛化能力与适应性，降低模型开发与维护成本，将是有价值的研究方向。  
（4）\textbf{轻量化工业控制策略：} 工业场景对生产节拍有严格要求，常运行在资源受限的工控设备上，这对策略模型的推理速度与资源消耗提出挑战。未来的工业具身智能研究需关注轻量化策略建模方法，探索模型压缩、结构重构、边缘计算友好的控制策略设计，确保在小型计算平台上实现稳定、实时、高效的工业控制。
\zh{
（5）\textbf{工业具身智能的标准化：}
随着工业具身智能技术的快速发展，标准化成为确保其广泛应用和互操作性的关键因素。当前，工业制造环境中存在多种系统和设备，它们来自不同的制造商，使用不同的标准和协议。这种多样性导致了跨系统整合的复杂性，增加了技术实现的难度。为了克服这一挑战，需要建立统一的工业具身智能标准，涵盖硬件接口、通信协议、数据格式和控制策略等方面，以促进工业具身智能技术的广泛应用，降低系统集成的成本和复杂性。
（6）\textbf{工业具身智能的安全挑战：} 
工业具身智能的部署将传统工业系统的安全边界从"机械-电气"层面扩展至"算法-网络-物理"融合领域，带来多维度的安全风险：在网络安全层面，高度互联的具身智能系统易遭受数据篡改、中间人攻击等威胁，可能导致产线瘫痪或产品质量失控；在算法安全层面，模型"幻觉"或对抗样本攻击可能引发错误决策；在物理安全层面，人机协作场景中的碰撞检测失效或力控异常可能造成人员伤害。需要构建集网络安全防护、算法安全保障、物理安全监控三位一体的防护体系，建立全生命周期安全管理机制，包括安全审计、故障追溯、应急响应等，形成闭环防护。
}

\section{结论（Conclusion）}

本文系统梳理了面向柔性制造的具身智能这一新兴交叉领域的发展脉络，重点探讨了感知、建模与决策在柔性制造场景中的集成路径。文章从“工业之眼”“工业之手”“工业之脑”三个维度出发，围绕受限感知条件下的工艺精准建模监测、柔性适配与高精操控的动态平衡、通用技能与专用工艺的协同融合三大核心难题，
归纳总结了当前的代表性研究进展，并结合典型案例分析了相关技术在实际工业场景中的应用，期望
为柔性制造趋势下的工业具身智能跨学科融合发展提供理论框架和实践参考。


\balance
\bibliographystyle{jqr-num}

\bibliography{test}

\begin{AuthorIntroduction}

\Intro 徐凯（1982~--），男，博士，教授，博士生导师。研究领域：计算机图形学、三维视觉、具身智能、数字孪生等。
\end{AuthorIntroduction}

\end{document}